\documentclass[lettersize,journal]{IEEEtran}
\usepackage{amsmath,amsfonts}
\usepackage{algorithmic}
\usepackage{algorithm}
\usepackage{array}
\usepackage[caption=false,font=normalsize,labelfont=sf,textfont=sf]{subfig}
\usepackage{textcomp}
\usepackage{stfloats}
\usepackage{url}
\usepackage{verbatim}
\usepackage{graphicx}
\usepackage{cite}
\usepackage{booktabs} 
\usepackage{wrapfig}
\usepackage{hyperref}
\usepackage{url}
\usepackage{color}
\usepackage{amsmath}
\usepackage{amssymb}
\usepackage{mathtools}
\usepackage{amsthm}
\usepackage{amsfonts,extarrows}
\usepackage{bm}
\usepackage{multirow}
\usepackage{fancyhdr}
\theoremstyle{plain}
\newtheorem{theorem}{Theorem} 
\newtheorem{observation}{Observation}
\newtheorem{lemma}{Lemma}

\newtheorem{assumption}{Assumption}

\begin{document}
\title{FedFA: Federated Learning with Feature Anchors to Align Features and Classifiers for Heterogeneous Data}

\author{Tailin~Zhou, \IEEEmembership{Graduate Student Member,~IEEE,}
        Jun~Zhang,~\IEEEmembership{Fellow,~IEEE,}
        and~Danny~H.K.~Tsang,~\IEEEmembership{Life~Fellow,~IEEE}
\thanks{
This work was supported in part by the Hong Kong Research Grants Council under the Areas of Excellence scheme grant AoE$/$E-601$/$22-R,  in part by NSFC$/$RGC Collaborative Research Scheme grant CRS$\_$HKUST603$/$22, in part by Guangzhou Municipal Science and Technology Project under Grant 2023A03J0011, Guangdong Provincial Key Laboratory of Integrated Communications, Sensing and Computation for Ubiquitous Internet of Things, and  National Foreign Expert Project, Project Number G2022030026L.

 T. Zhou is with IPO,  Academy of Interdisciplinary Studies, The Hong Kong University of Science and Technology, Clear Water Bay, Hong Kong SAR, China (Email: tzhouaq@connect.ust.hk).
J. Zhang is with the Department of Electronic and Computer Engineering, The Hong Kong University of Science and Technology, Clear Water Bay, Hong Kong SAR, China  (E-mail: eejzhang@ust.hk).
D. H.K. Tsang is with the Internet of Things Thrust, The Hong Kong University of Science and Technology (Guangzhou), Guangzhou, Guangdong, China, and also with the Department of Electronic and Computer Engineering, The Hong Kong University of Science and Technology, Clear Water Bay, Hong Kong SAR, China (Email: eetsang@ust.hk). (The corresponding author is J. Zhang.)
}
}

 \markboth{Accepted by IEEE Transactions on Mobile Computing. 
  Copyright was transferred to IEEE.  DOI: \href{https://doi.org/10.1109/TMC.2023.3325366}{10.1109/TMC.2023.3325366}.}%
{}

\maketitle

\begin{abstract}
Federated learning allows multiple clients to collaboratively train a model without exchanging their data,  thus preserving data privacy. 
Unfortunately, it suffers significant performance degradation due to heterogeneous data at clients. 
Common solutions involve designing an auxiliary loss to regularize weight divergence or feature inconsistency during local training. 
However, we discover that these approaches fall short of the expected performance because they ignore the existence of a \textit{vicious cycle} between feature inconsistency and classifier divergence across clients.
This \textit{vicious cycle} causes client models to be updated in inconsistent feature spaces with more diverged classifiers.
To break the \textit{vicious cycle}, we propose a novel framework named  \textit{\textbf{Fed}erated learning with \textbf{F}eature \textbf{A}nchors} (FedFA).
FedFA utilizes feature anchors to align features and calibrate classifiers across clients simultaneously. 
This enables client models to be updated in a shared feature space with consistent classifiers during local training.
Theoretically, we analyze the non-convex convergence rate of FedFA.
We also demonstrate that the integration of feature alignment and classifier calibration in FedFA brings a \textit{virtuous cycle} between feature and classifier updates, which breaks the \textit{vicious cycle} existing in current approaches.
Extensive experiments show that FedFA significantly outperforms existing approaches on various classification datasets under label distribution skew and feature distribution skew. 
\end{abstract}


\section{Introduction}\label{Introduction}

With massive data located at mobile clients of large-scale networks such as the Internet of Things (IoT) networks, mobile networks and vehicular networks, federated learning \cite{mcmahan2017communication} enables clients to jointly train a machine learning model without collecting client data into a centralized server, thus preserving data privacy.
However, the private data are typically heterogeneous across clients, resulting in slower convergence \cite{li2020federated,karimireddy2020scaffold,wang2022accelerating,sun2023accelerating} and degraded generalization performance \cite{zhao2018federated,li2021federated,wei2021user}.
This is because data heterogeneity makes the local objectives inconsistent with the global objective and causes drifts in client updates to slow down convergence.
The drifts then deviate the converged model from the expected optima and degrade its generalization performance according to \cite{kairouz2021advances,wang2020tackling}.

Existing works have observed that data heterogeneity induces weight divergence (from the view of parameter space) and feature inconsistency (from the view of feature space) when clients train their models.
Furthermore, the implementation of federated learning in wireless mobile networks may exacerbate the negative impact of data heterogeneity due to limited wireless resources \cite{chen2020joint, zhang2021optimizing}.
Common solutions add a regularizer to control weight divergences such as \cite{li2020federated,xu2022adaptive} or feature inconsistency across clients such as  \cite{li2021model,mu2021fedproc}. 
See more discussion in Section \ref{Related Work}.
Nevertheless, recent works like \cite{li2021federated} found that these methods did not show clear advantages over the canonical  FedAvg \cite{mcmahan2017communication} on various classification tasks.

\begin{table}[t]
\centering  NOMENCLATURE LIST  \\
\vspace{5pt}
\label{tab:Nomenclature}
\begin{tabular}{@{}ll|ll@{}}
\toprule
$\textbf{x}$                     & Data sample                       & $y$                    & Label                          \\
$\mathcal{D}_i$                  & Client $i$ dataset                & $\mathcal{D}_{i,c}$    & $\mathcal{D}_i$ w/ label $c$   \\
$\mathcal{D}$                    & Global dataset                    & $C$                    & Total class number             \\
$n_i$                            & Client $i$ sample number          & $n$                    & Total sample number            \\
$N$                              & Total client number               & $\mathcal{L}(\cdot)$   & Global loss                    \\
$\mathcal{L}_{{\rm sup}}(\cdot)$ & Supervised loss                   & $\mathcal{L}_i(\cdot)$ & Client $i$ loss                \\
$\mathbf{w}$                     & Global model                      & $\mathbf{w}_i$         & Client $i$ model               \\
$\bm{\theta}$                    & Feature extractor                 & $\mathbf{h}$           & Feature                        \\
$f_{\bm{\theta}}(\cdot)$         & Forward function of $\bm{\theta}$ & $\bm{\phi}$            & Linear classifier              \\
$f_{\bm{\phi}}(\cdot)$           & Forward function of $\bm{\phi}$   & $p_{\cdot,c}$          & classifier output on class $c$ \\
$\Delta_{\bm{\phi}}$             & Classifier update                 & $\Delta_{\mathbf{h}}$ & Feature update                 \\
$\mathbf{a}_{c}$                 &  Class $c$ feature anchor      & $\mathbf{m}_{c,\cdot}$ & Class $c$ anchor momentum   \\ \bottomrule
\end{tabular}
\end{table}

\begin{figure}[t]
    \centering
    \includegraphics[width=0.48\textwidth]{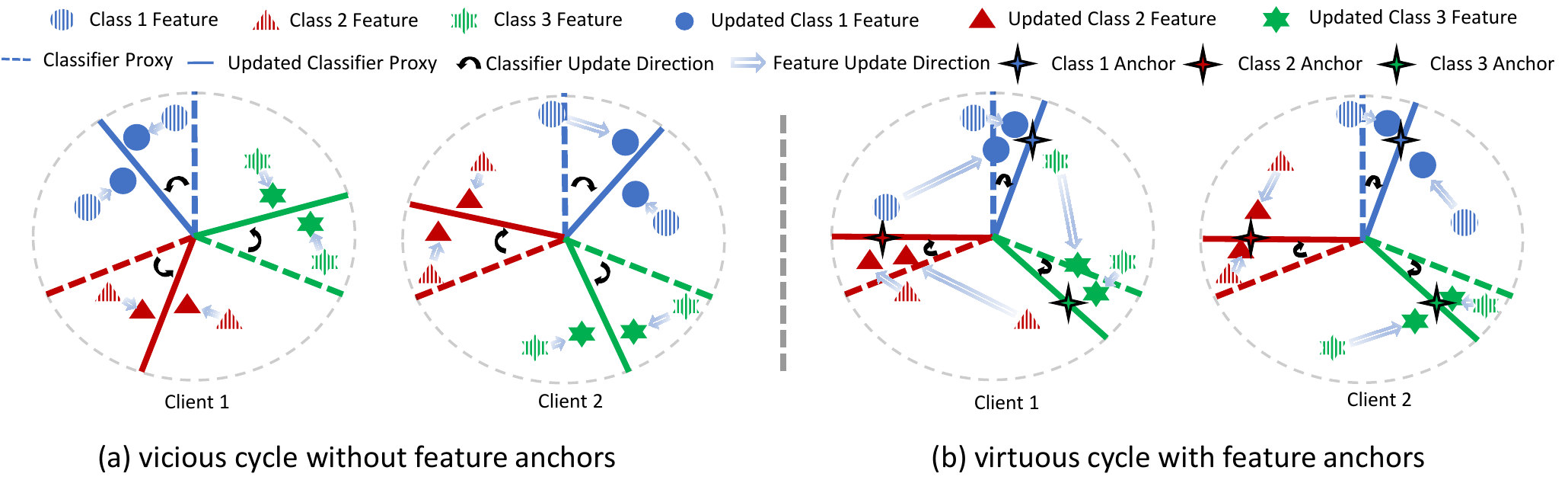}
    \caption{A toy example of two clients with three class features to show the rationale of FedFA.
Figures \ref{toy_example}(a) and  \ref{toy_example}(b) illustrate the relationship between feature and classifier updates without and with feature anchors, respectively.
The \textit{vicious cycle} in Figure \ref{toy_example}(a) describes that the diverged classifier updates between client 1 and client 2 due to the inconsistent features (to be verified in Figure \ref{Experimental Validation}) make their extractors to map more inconsistent features to reduce the classification error (i.e., the angle between features and their corresponding classifier proxies) in local training.
    The \textit{virtuous cycle} in Figure \ref{toy_example}(b)  means that features and classifiers aligned by feature anchors between clients promote the consistency of client features and classifiers (to be verified in Figure \ref{Experimental Validation}).}
    \label{toy_example}
\end{figure}

To unravel the underlying reasons for the ineffectiveness of existing methods, we first observe that data heterogeneity (including heterogeneous label and feature distributions across clients) induces feature inconsistency and classifier divergence concurrently across clients.
We then theoretically and empirically identify the existence of a \textit{vicious cycle} between feature inconsistency and classifier divergence across clients, as shown in Figure \ref{toy_example}(a). 
Specifically, inconsistent features diverge the classifier updates, and then the diverged classifiers force feature extractors to map to more inconsistent feature spaces, thus diverging client updates.
Therefore, the \textit{vicious cycle} between feature inconsistency and classifier divergence causes client models to be updated in inconsistent feature spaces with more diverged classifiers.

To overcome the \textit{vicious cycle},   we propose a novel and effective framework called \textit{\textbf{Fed}erated learning with \textbf{F}eature \textbf{A}nchors} (FedFA) for classification tasks to address the skewed label and feature distributions across clients.
FedFA introduces the feature anchors to unify the extraction of features by clients from a shared feature space and to calibrate classifiers into this space during local training.
We show theoretically and empirically that FedFA enables smoother classifier updates and polymerized features, which brings a \textit{virtuous cycle} between classifier similarity and feature consistency, as shown in Figure \ref{toy_example}(b),  contrary to the above \textit{vicious cycle}. 
Meanwhile, we analyze the non-convex convergence rate of FedFA.
Finally, our experiments show that FedFA significantly outperforms the existing methods under label distribution skew, feature distribution skew, and their combined skew.
To the best of our knowledge, we are the first to study the combined label and feature distribution skews.

With insight into the relationship between feature and classifier updates on heterogeneous data, the proposed FedFA, assisted by feature anchors, trains client models in a consistent feature space with the classifiers corresponding to this space.
Our main contributions are summarized as follows:
\begin{itemize}
    \item We demonstrate that data heterogeneity across clients (i.e., skewed label and feature distributions) leads to a \textit{vicious cycle} between classifier divergence and feature inconsistency across client models, which degrades the training performance.
    \item To break the \textit{vicious cycle}, we introduce a novel framework, FedFA, which leverages feature anchors to align features and classifiers across clients such that all client models are updated in a uniform feature space with corresponding classifiers.
    \item We prove that FedFA improves the Lipschitzness of the loss on the classifier-weight space, which brings a \textit{virtuous cycle} between feature consistency and classifier harmony. Meanwhile, we analyze the non-convex convergence rate of FedFA.
    \item   Our experiments demonstrate the significant advantage of FedFA over the baseline algorithms under various data heterogeneity settings.
\end{itemize}

The remainder of this paper is organized as follows.
Section \ref{Related Work} reviews related works, and Section \ref{Problem} introduces preliminaries and problems. 
Our motivation and our method are presented in Section \ref{Motivation} and Section \ref{Proposed Framework}, respectively. 
Simulation results are given in Section \ref{Experiments}, 
and the concluding remarks and future works are provided in Section \ref{Conclusion}.

\section{Related Works}\label{Related Work}

To alleviate the model divergence \cite{zhao2018federated}, common methods add an auxiliary loss or improve the model aggregation scheme to tackle data heterogeneity across clients in federated learning.  
Here, we mainly introduce the client-side-based methods closely related to ours and briefly introduce other methods. Comprehensive field studies have appeared in \cite{kairouz2021advances,tan2022towards,shao2023survey}.

\subsubsection{Tackle data heterogeneity by controlling weight divergence}
To prevent local models from converging to their local minima instead of global minima, many works introduce a  regularizer to control local updates from the perspective of weight space.
For example, FedProx \cite{li2020federated} uses the Euclidean distance between the local and global models as a regularizer.
FedDyn \cite{acar2021federated}  modifies the local objective with a dynamic regularizer based on the first-order condition to make clients' local minima consistent with the global minima.
SCAFFOLD \cite{karimireddy2020scaffold} uses the variance-reduction technique found in standard convex optimization to create a control variate that adjusts client updates to be more similar to the global update.
According to \cite{luo2021no}, instead of adjusting the weights of the entire model, it is more effective to focus on the classifier layer (which is the final layer of the model) as it is most affected by label distribution skew. The solution proposed is to calibrate the classifiers using virtual features after training.
Additionally, in \cite{zhang2022federated}, a fine-grained calibrated classifier loss is incorporated to address the issue of over-fitting of underrepresented classes in clients' datasets that are affected by the long-tail effect.
Some methods enable clients to share their data with privacy guarantees, such as sharing a synthesized dataset in \cite{luo2021no,li2022federated,tang2022virtual} and sharing coded data in \cite{sun2022stochastic,shao2022DReS}.


\subsubsection{Tackle data heterogeneity by controlling feature inconsistency}
Recent studies have discovered feature inconsistency among clients from the perspective of feature space.
To control model divergence, certain contrastive learning techniques like feature alignment and logit distillation are employed.
For instance, MOON \cite{li2021model} introduces a model-contrastive regularizer to maximize (or minimize) the agreement of the features extracted by the local model and that by the global model (or the local model of the previous round).
In place of the model-contrastive term, FedProc \cite{mu2021fedproc} and FedProto  \cite{tan2022fedproto} add a prototype-contrastive term to regularize the features within each class with class prototypes  \cite{snell2017prototypical}.
In \cite{itahara2021distillation,zhang2021federated},   clients share their own models with other clients and take logit distillation to align the logit outputs of all client models.
 

\subsubsection{Tackle data heterogeneity by improving aggregation schemes}
Some works have developed alternative aggregation schemes at the server to tackle data heterogeneity in federated learning.
For instance,    unbalanced data induces a different number of local updates and causes an objective inconsistency problem found in \cite{wang2020tackling}, which propose FedNova to eliminate the inconsistency by normalizing the local updates before averaging. 
Besides,   adaptive momentum updates on the server side was adopted in \cite{reddi2021adaptive}    to mitigate oscillation of global model updates when the server activates the clients with a limited subset of labels.
Beyond layer-weighted averaging,  some works like  FedMA \cite{Wang2020Federated} 
introduce neuron-wise averaging because there may exist neuron mismatching from permutation invariance of neural networks in federated learning. 
These ideas complement our work and can be integrated into our method because our method only adds an auxiliary loss at the client side.

\subsubsection{Tackle data heterogeneity in wireless networks}
When implemented in a realistic wireless network, the performance of federated learning is further affected by wireless factors and client availability  \cite{chen2020joint}.
To enhance the effectiveness of federated learning, wireless resource allocation and client selection can be optimized together using a closed-form expression for the expected convergence rate on FedAvg, as described in \cite{chen2020joint}, or a hierarchical training architecture, as discussed in \cite{zhang2021optimizing}.
A recent study \cite{wang2022performance} delved into the concept of model quantization as a means to enhance wireless communication.
This study suggests a reinforcement learning approach using a model-based method to determine the participating clients and the bitwidths used for model quantization. 
To further preserve privacy,   the truth-discovery technique and homomorphic cryptosystem are introduced by \cite{zhang2023reliable} to identify the client reliability and thereby decrease the impact of anomalous clients.
Besides, DetFed is a recent work \cite{Yang2023DetFed} that presents a deterministic federated learning framework for industrial IoT, which integrates 6G-oriented time-sensitive networks to improve the reliability and latency of the training process.

According to \cite{he2021fedcv},   existing works may not provide stable better performance gains over FedAvg\cite{mcmahan2017communication} in classification tasks, which motivates us to analyze the relationship between classifier updates and features in local training. 
We find that existing methods ignore the inherent relationship (i.e., a \textit{vicious cycle}) between these two updates and then still suffer from either feature inconsistency or classifier divergence. 
Different from these methods, our method breaks the \textit{vicious cycle} by taking feature anchors to align both feature and classifier updates across clients. 
Moreover, our method addresses both label and feature distribution skew, unlike others that only tackle label distribution skew and improve the performance of federated learning under both skews.

\section{Preliminaries and Problem Formulation}\label{Problem}

\subsection{Terminology} Suppose that the global dataset $\mathcal{D}$ consists of $C$ classes indexed by $[C]$ for classification tasks. 
Let $  (\textbf{x}, y) \in \mathcal{D} $ and $   \mathcal{D} \subseteq \mathcal{X} \times [C] $  where $  (\textbf{x}, y)$ denotes a sample $\textbf{x}$ in the input-feature space $\mathcal{X} $, with the corresponding label $y$ in the label space $[C]$.
We represent $[C_i]$  as  a subset of $[C]$ (i.e., $\cup_{i=1}^N [C_i] = [C]$) and $   \mathcal{D}_{i,c} = \{ (\textbf{x}, c) \in \mathcal{D}_{i}; c \in [C_i]\} $ as the subset of $   \mathcal{D}_{i}$ with the  label $c$ at the dataset $\mathcal{D}_{i}$.

Furthermore, we decompose the classification model parameterized by $\mathbf{w}=\{\bm{\theta},\bm{\phi}\}$ into a \textit{feature extractor} (i.e., other layers except for the last layer of the model denoted by $f_{\bm{\theta}}: \mathcal{X} \rightarrow \mathcal{H} $)  and a \textit{linear classifier} (i.e., the last layer of the model denoted by $f_{\bm{\phi}}: \mathcal{H}\rightarrow \mathbb{R}^{[C]} $). 
Specifically, the feature extractor maps a sample   $\textbf{x}$ into a feature vector $\mathbf{h}= f_{\bm{\theta}}(\textbf{x})$  in the feature space $ \mathcal{H} $, and then the classifier  generates a probability distribution $f_{\bm{\phi}}(\mathbf{h})$ as the prediction for $\textbf{x}$.

\subsection{Federated learning}
We consider a federated learning framework with $N$ clients,   each with its own dataset $\mathcal{D}_i  \sim \mathbb{P}_i:\mathbb{R}^d \times  \mathbb{R}$ with $n_i$ data samples.
The global dataset is the union of all client datasets and denoted by $\mathcal{D} = \cup_{k=1}^K \mathcal{D}_k \sim \mathbb{P}$ on $ \mathbb{R}^d \times  \mathbb{R}$ with  $n = \sum_{i=1}^N n_i$ data samples.
The  objective of  federated learning is to minimize the expected global loss $  \mathcal{L}(\mathbf{w}) $   on  $\mathcal{D}$, which is formulated as: 
\vspace{-2mm}
\begin{equation}
 \begin{aligned}
   \min_{\mathbf{w} \in \mathbb{R}^d}  \mathcal{L}(\mathbf{w} ) & 
 :=\mathbb{E}_i[\mathcal{L}_i(\mathbf{w} ) ]    =  \sum_i^N \frac{n_i}{n} \mathcal{L}_i(\mathbf{w} ) \\ &   =  \sum_i^N \frac{n_i}{n} \mathbb{E}_{(\textbf{x}, y) \in \mathcal{D}_i}[  l_i (\mathbf{w};(\textbf{x}, y))],
    \end{aligned}
   \label{fl}
   \vspace{-1mm}
\end{equation}
where  $\mathcal{L}_i(\mathbf{w})$ is the expected local objective function on the local dataset $\mathcal{D}_i$ of the $i$-th client. 
When heterogeneous data exist at clients i.e., $\mathbb{P}_i\neq \mathbb{P}$, FedAvg   \cite{mcmahan2017communication}, a   canonical method,   takes client models $\{\mathbf{w}_i\}_{i=1}^N$ to solve (\ref{fl}) by minimizing the client loss function $\mathcal{L}_i(\mathbf{w})$ locally and obtaining the global model $\boldsymbol{w} =\sum_{i=1}^{N} \frac{n_k}{n}\boldsymbol{w}_i$  round by round until  $\boldsymbol{w}$ converges.

\subsection{Data heterogeneity in federated learning}
According to \cite{kairouz2021advances}, there are two types of data heterogeneity: feature distribution skew and label distribution skew. 
In this work, we may refer to them as feature skew and label skew for saving pages, respectively.
Suppose that the $i$-th client data distribution follows $P_i(\textbf{x},y)=P_i(\textbf{x}|y)P_i(y) =P_i(y|\textbf{x})P_i(\textbf{x}) $, where $x$ and $y$ denote the feature  and label, respectively.
Here, $P_i(\textbf{x})$ and $P_i(y)$ denote the input feature marginal distribution and label marginal distribution of the $i$-th client distribution, respectively.
Our research delves into three distinct forms of data heterogeneity, in contrast to prior studies which only investigated one of these forms. These include:
\begin{itemize}
    \item   Label distribution skew:   The label marginal distribution $P_i(y)$  varies across clients while $P_i(\textbf{x}|y) = P_j(\textbf{x}|y) $ for all clients $i$ and $j$. 
    For example, the $i$-th client holds different labels from that of the $j$-th client when $i\neq j$.
    \item Feature distribution skew:   The input feature marginal distribution $P_i(\textbf{x})$  varies across clients while $P_i(y|\textbf{x}) = P_j(y|\textbf{x}) $ for all clients $i$ and $j$. 
    For instance,   the samples held by $i$-th client have different styles from that of the $j$-th client when considering the same label and $i\neq j$.
    \item   Label and feature  distribution skews:   The $i$-th client and   $j$-th client hold  $P_i(y) \neq P_j(y)$ when sharing the same $P(\textbf{x})$, and hold $P_i(\textbf{x}) \neq P_j(\textbf{x}) $ when sharing the same $P(y)$. 
    Specifically, there is an occurrence of label distribution skew and feature distribution skew.
\end{itemize}

When these distribution skews exist across clients (i.e., $\mathcal{D}_i \neq \mathcal{D}_j$ when $i \neq j$) as per \cite{zhao2018federated},  federated learning by optimizing (\ref{fl}) is incomparable to centralized training with the global dataset $\mathcal{D} = \cup_i^N\mathcal{D}_i$.
Specifically,  these skews cause client updates to diverge and degrade the performance of federated learning, such as decreasing convergence speed \cite{wang2020tackling} and degrading the generalization of trained models \cite{li2020federated}. 
However, there is still a lack of work to observe the impact of these skews on different architectures of the trained model.

\begin{figure*}[th]
    \centering
    \includegraphics[width=\textwidth]{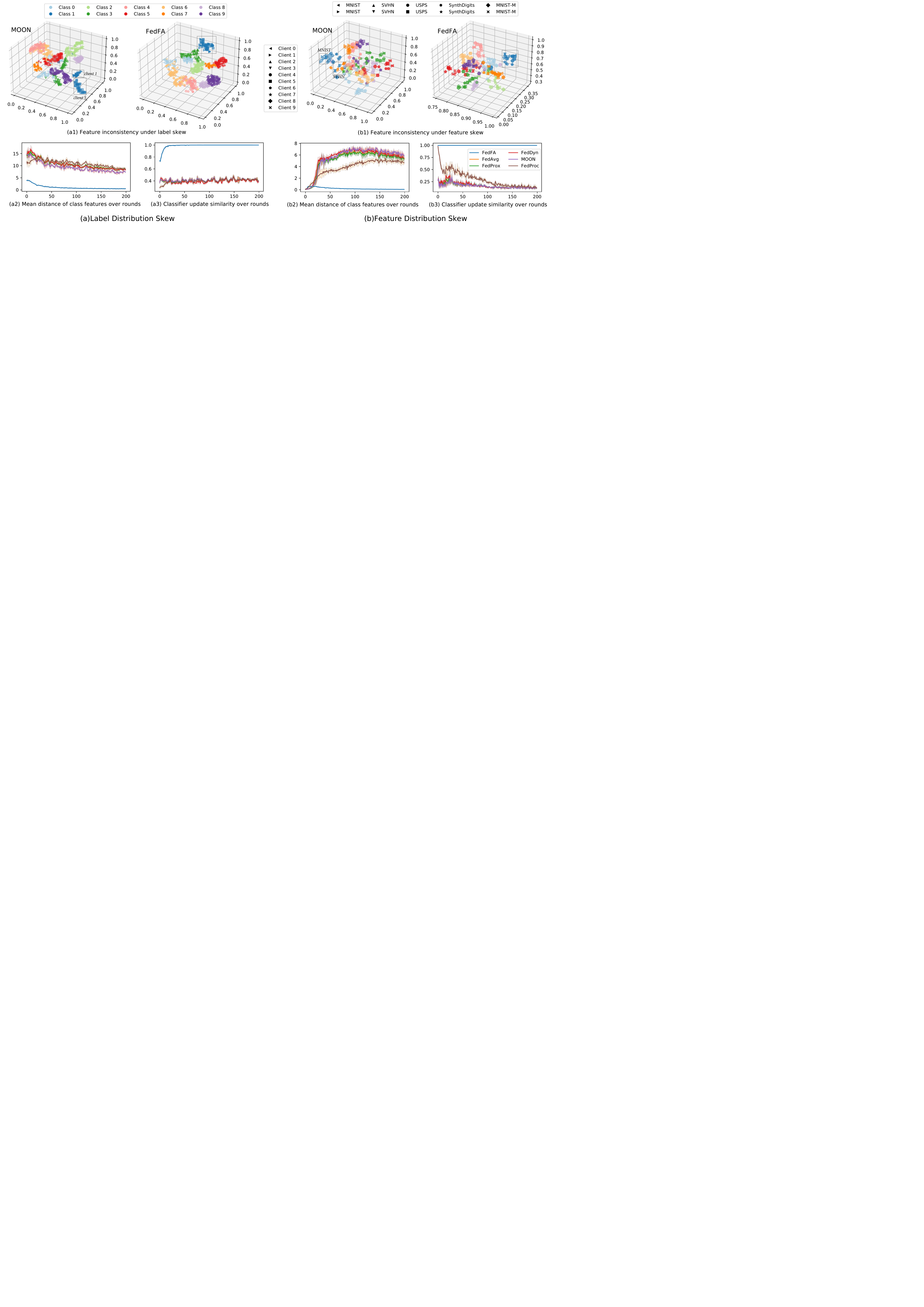}
        \vspace{-4mm}
    \caption{The t-SNE visualization of feature maps, mean distance of class features, and classifier update similarity with MOON and FedFA (our proposed scheme) under label  skew ($\#C=2$)  in FMNIST  and feature skew in Mixed Digit. 
It demonstrates the simultaneous occurrence of feature inconsistency and classifier divergence, which persists even as the global model converges in the baselines. 
    Meanwhile, both label and feature skews cause these two issues. 
    For example, for class 1 of feature inconsistency (i.e., dark blue), Figure \ref{Experimental Validation}(a1) shows that client 1 (i.e., \textit{right triangle}) and client 5 (i.e., \textit{square}) extract inconsistent features (in the black box) under label skew, and Figure \ref{Experimental Validation}(b1) presents that the features of MNIST (i.e., \textit{left and right triangles}) deviate from that of SVHN (i.e., \textit{up and down triangles}) under feature skew. 
    In contrast, FedFA alleviates feature inconsistency and classifier divergence significantly.
    }
    \label{Experimental Validation}
\end{figure*}

\section{Motivation: Inconsistent Features and Diverged Classifiers Across Clients}
\label{Motivation}


  In this section,  we explore the effect of data heterogeneity in federated learning in view of the relationship between feature and classifier updates.
We empirically and theoretically demonstrate the simultaneous occurrence of feature inconsistency and classifier divergence across clients during training.

\subsection{Experimental demonstration}
We consider the FMNIST \cite{xiao2017fashion} task with label skews and the Mixed Digits task \cite{li2021fedbn} with feature skews at ten clients in federated learning. 
For feature visualization, as shown in the first row of Figure  \ref{Experimental Validation}, we visualize feature maps of different methods of federated learning using t-SNE visualization \cite{van2008visualizing}.
For classifier-update visualization, we input the same samples into all client models to compute the mean distance of class feature and classifier update similarity at the end of each round during training, as shown in the second row of Figure \ref{Experimental Validation}.
Note that features are visualized as per the classes (or digit dataset) owned by a client under label (or feature) distribution skew.

Figure  \ref{Experimental Validation} shows that  MOON aiming at feature alignment still suffers from feature inconsistency under both label and feature skews.
Specifically, there exists significant feature inconsistency of class 1 (i.e., dark blue), class 5 (i.e., dark red) and class 9 (i.e., dark purple) samples, but our method FedFA introduced in Section \ref{Proposed Framework} overcomes the inconsistency. 
Moreover, we find that feature inconsistency also exists in other existing methods, where the visualizations are provided in our supplementary materials. 
This indicates that the existing methods cannot fully generate a  consistent feature space for client models even though they focus on aligning features or controlling classifier divergence across clients.

Furthermore, Figure \ref{Experimental Validation} reveals the simultaneous occurrence of feature inconsistency and classifier divergence.
Specifically, the lower the similarity of classifier updates, the more inconsistent the feature mapping between clients in Figure \ref{Experimental Validation}(b2) and \ref{Experimental Validation}(b3).
Meanwhile, as the global model converges, these two issues are only slightly alleviated under label skew, while those under feature skew even get worse.
These findings indicate an interactive relationship between feature and classifier updates, making it impossible to solve the heterogeneous data problem of federated learning by only controlling any feature extractors or classifiers.

\subsection{Theoretical demonstration} 
We follow \cite{movshovitz2017no} to represent the classifier parameters $\bm{\phi}_i$ of the $i$-th client as $C$ weight vectors  $\{\bm{\phi}_{i,c} \}_{c=1}^C$, where $\bm{\phi}_{\cdot,c} $ refers to the proxy for the $c$-th class samples. 
For simplicity, we set all the bias vectors of the classifier as zero vectors and use the cross-entropy loss as the supervised loss.
The supervised loss of the $i$-th client  on its classifier is represented as:
\begin{equation}
\begin{aligned}
 \mathcal{L}_{{\rm sup}_i} (\bm{\phi}_i):= 
 \mathbb{E}_{(\textbf{x},y)  \in \mathcal{D}_i} [l_{{\rm sup}_i}(\bm{\phi}_i; (\textbf{x}_j,y_j) )]
\\=-\frac{1}{n_i}\sum_{j=1,y_{j}=c}^{n_i}  \log
   \frac{\exp \left(\bm{\phi}_{i,c}^\top \mathbf{h}_{i,y_j}\right)}{\sum_{z=1}^{C} \exp \left(\bm{\phi}_{i,z}^\top \mathbf{h}_{i,y_j}\right)},
\end{aligned}
\label{sup loss}
\end{equation}
 where $\mathbf{h}_{i,y_j} = f_{\mathbf{\theta}_i}(\mathbf{x}_j)$ is the feature mapping of a sample  $(\mathbf{x}_j,y_j)$. 
We will demonstrate the relationship between classifier and feature updates across clients as follows.
 
Firstly, the classifier updates diverge across clients. 
For the $c$-th local proxy $\bm{\phi}_{\cdot,c}$, the
\textit{positive features} and \textit{negative features} denote the features from the $c$-th class samples and other classes, respectively. Let  $p_{\cdot, c}^{(j)} = \exp \left(\bm{\phi}_{\cdot,c}^\top \mathbf{h}_{\cdot,y_j}\right)/{\sum_{z=1}^{C} \exp \left(\bm{\phi}_{\cdot,z}^\top \mathbf{h}_{\cdot,y_j}\right)} $ and we follow a mild assumption in \cite{wang2021addressing,zhang2022federated} that the extracted feature $\mathbf{h}_{\cdot,c}$ of samples and their corresponding prediction output $p_{\cdot, c}^{(j)}$   will be similar for the $c$-th class within one client (i.e., $\overline{p_{\cdot,{c}}^{({c})}} \overline{\mathbf{h}}_{\cdot,{c}} = \frac{1}{n_{\cdot,c}}\sum_{y_{j}=c}^{n_{\cdot,c}} p_{\cdot, c}^{(j)}  \mathbf{h}_{\cdot,y_j}$  where $\overline{p_{\cdot,{c}}^{({c})}}=\frac{1}{n_{\cdot,c}}\sum_{y_{j}=c}^{n_{\cdot,c}} p_{\cdot, c}^{(j)}$ and $ \overline{\mathbf{h}}_{\cdot,{c}} = \frac{1}{n_{\cdot,c}}\sum_{y_{j}=c}^{n_{\cdot,c}} \mathbf{h}_{\cdot,y_j}$).
Without losing the generality, we characterize classifier update deviation across clients as follows.
\begin{lemma}
(Classifier update deviation. See proof in Appendix \ref{proof of classifier deviation}).
For client $a$ and client $b$ with the same sample number $n_a=n_b=n$, the deviation of classifier update  $\Delta_{\bm{\phi}_c} = \Delta  \bm{\phi}_{a,c} -  \Delta \bm{\phi}_{b,c} $ is:
 \begin{equation*}
 \begin{aligned}
\Delta_{\bm{\phi}_c}= &\frac{\eta}{n}   [ \underbrace{\left( n_{a,c}\left(1-\overline{p_{a, c}^{(c)}}\right) \overline{\mathbf{h}}_{a,c} - n_{b,c}\left(1-\overline{p_{b, c}^{(c)}}\right) \overline{\mathbf{h}}_{b,c}\right)}_{\text{deviation by \textit{mean positive features}}: \Delta_{\bm{\phi}_c}^{(+)}}\\
  \end{aligned}
\end{equation*}
 \begin{equation*}
 \begin{aligned}
 & - \underbrace{ \left( \sum_{\bar{c}_a\neq c} n_{a,\bar{c}_a} \overline{p_{a,c}^{(\bar{c}_a)}} \overline{\mathbf{h}}_{a,\bar{c}_a}-\sum_{\bar{c}_b\neq c}   n_{b,\bar{c}_b} \overline{p_{b,c}^{(\bar{c}_b)}} \overline{\mathbf{h}}_{b,\bar{c}_b}  \right)}_{\text{deviation by \textit{mean negative features}}: \Delta_{\bm{\phi}_c}^{(-)}}]
  \end{aligned}
\end{equation*}
where $\eta$ is the learning rate, $n_{\cdot,c}$ is  the $c$-th class sample number, and  $\bar{c}_{(\cdot)}$ is one of the negative classes of the $c$-th class of one client (e.g.,  $\bar{c}_a \in \{[C_a] \setminus c\}$ for client $a$).
\label{gradient of classifier}
\end{lemma}
Lemma \ref{gradient of classifier} formulates the deviation of classifier updates between any two clients. Then, let clients $a$ and  $b$   hold the same samples if their datasets have the same class under label distribution skew, and let clients $a$ and $b$ have all classes under feature distribution skew. 
With Lemma \ref{gradient of classifier},   we demonstrate how data
heterogeneity diverges classifier updates.
\begin{theorem}
(Classifier update divergence under data heterogeneity. See proof in Appendix  \ref{proof:classifier}).
 For label distribution skew (different label sets $[C_a] \neq [C_b]$), 
 when $c \in [C_a] \cap [C_b]$,   $\mathbf{\overline{h}}_{a,c}=\mathbf{\overline{h}}_{b,c}$, $\overline{p_{a, c}^{(c)}} = \overline{p_{b, c}^{(c)}}$ and $\Delta_{\bm{\phi}_c}^{(+)}=0$, and then $\|\Delta_{\bm{\phi}_c}\|^2= \frac{\eta^2}{n^2}\| \Delta_{\bm{\phi}_c}^{(-)} \|^2 > 0$; 
  when $c \in \{[C]\setminus \{[C_a]\cup [C_b]\} \}$, $n_{a,c}=n_{b,c}=0$ and $\Delta_{\bm{\phi}_c}^{(+)}=0$, and then $\|\Delta_{\bm{\phi}_c}\|^2=\frac{\eta^2}{n^2}\| \Delta_{\bm{\phi}_c}^{(-)} \|^2  > 0$;
 when  $c \in [C_a] \setminus \{[C_a] \cap [C_b]\}$ or $c \in [C_b] \setminus \{[C_a] \cap [C_b]\}$, $n_{a,c}=0$ or $n_{b,c}=0$, and then $\|\Delta_{\bm{\phi}_c}\|^2 > 0$.
 For feature distribution skew,  $\mathbf{\overline{h}}_{a,c} \neq \mathbf{\overline{h}}_{b,c}$, and then  $\|\Delta_{\bm{\phi}_c}\|^2 > 0$. .
\label{theorem classifier}
\end{theorem}
 Theorem \ref{theorem classifier} reveals that both label and feature skews diverge classifier updates across clients (i.e., $\Delta_{\bm{\phi}_c}\neq 0$).
 This explains the classifier divergence as shown in Figure \ref{Experimental Validation}.
For example, when both clients have the label $c$, i.e., $c \in [C_a] \cap [C_b]$, the classifier divergence is induced by mean negative features;  when both clients do not have the label $c$, i.e., $c \in \{[C]\setminus \{[C_a]\cup [C_b]\} \}$, the classifier divergence is induced by mean positive features.

Secondly, the diverged classifiers would induce feature inconsistency across clients.
We characterize the feature deviation of the same samples across clients as follows.
\begin{lemma}
(Feature update deviation. See proof in Appendix  \ref{proof:Def 2}).
For client $a$ and client $b$ with $n$ samples of one class, the   deviation of  mean class features  $\Delta_{\mathbf{h}_c} = \Delta \mathbf{\overline{h}}_{a,c} -  \Delta \mathbf{\overline{h}}_{b,c} $ is:
 \begin{equation*}
 \begin{aligned}
 \Delta_{\mathbf{h}_c}= \eta  [ \underbrace{\left( (1-\overline{p_{a, c}^{(c)}}) {\bm{\phi}}_{a,c} - (1-\overline{p_{b, c}^{(c)}}) {\bm{\phi}}_{b,c}\right)}_{\text{deviation by \textit{ positive classifier proxy}}: \Delta_{{\mathbf{h}_c}}^{(+)}}
 \\- \underbrace{ \left( \sum_{\bar{c}} \overline{p_{a,\bar{c}}^{(c)}} {\bm{\phi}}_{a,\bar{c}}-\sum_{\bar{c}}   \overline{p_{b,\bar{c}}^{(c)}}  {\bm{\phi}}_{b,\bar{c}}  \right)}_{\text{deviation by \textit{negative classifier proxies}}: \Delta_{{\mathbf{h}_c}}^{(-)}}]
  \end{aligned}
\end{equation*}
where $\bar{c} \in \{[C]\setminus c\}$.
\label{gradient of feature}
\end{lemma}
 Lemma \ref{gradient of feature} formulates the deviation of feature updates between any two clients. 
According to Theorem \ref{theorem classifier}, all classifier proxies under data heterogeneity have $\bm{\phi}_{a,c}\neq\bm{\phi}_{b,c}$, inducing $\Delta_{{\mathbf{h}_c}}^{(+)} \neq \Delta_{{\mathbf{h}_c}}^{(-)}$.
Therefore,  $\|\Delta_{\mathbf{h}_c}\|^2 > 0$  and inconsistent feature updates occurs across clients under heterogeneous data.

 Finally, we conclude the relationship between classifier divergence and feature inconsistency as follows.
\begin{theorem}
(\textit{Relationship between classifier deviation and feature deviation}. See proof in Appendix  \ref{proof:Def 2})
Combining Lemma \ref{gradient of classifier} and Lemma \ref{gradient of feature}, for client $a$ and client $b$ with output   $\overline{p_{a,{c}}^{(c)}} =\overline{p_{b,{c}}^{(c)}}$ and $\overline{p_{a,\bar{c}}^{(c)}} = \overline{p_{b,\bar{c}}^{(c)}}$ on the $c$-th class and the same sample numbers $n_{a,c}=n_{b,c}$, the relationship between their classifier update deviation and feature update deviation is $   \Delta_{\mathbf{h}_c}=  -\eta \Delta_{\bm{\phi}_c} + \sum_{\hat{c}\in [C]}   \overline{p_{b,\hat{c}}^{(c)}} \Delta_{\bm{\phi}_{\hat{c}}} $. 
\label{vicious cycle}
\end{theorem}
  Theorem \ref{vicious cycle} unravels a negative effect of data heterogeneity:
\begin{observation}
(A vicious cycle)
  \textit{Data heterogeneity firstly induces classifier update divergence, which then leads to inconsistent feature maps; these inconsistent features in turn force different classifiers to diverge worse}, as illustrated in the toy example of Figure \ref{toy_example}(a).
Moreover, as shown in Figures \ref{Experimental Validation}(a2) and (a3), the \textit{vicious cycle} does not disappear even when the training of the global model converges.
\label{observation1}
\end{observation}

In summary,   feature inconsistency and classifier divergence are coupled to degrade the performance of federated learning.
To break the \textit{vicious cycle}, it is necessary to address both issues simultaneously.

\section{Federated Learning with Feature Anchors}
\label{Proposed Framework}

We propose FedFA to train client models in a consistent feature space with the classifiers corresponding to this space in order to break the \textit{vicious cycle} revealed in Theorem \ref{vicious cycle}.

\subsection{Propose Method: FedFA}
With a total of $C$ classes in the whole dataset, the server initiates $C$ feature anchors  $\{\mathbf{a}_c\}_{c=1}^C \in \mathcal{H} \times [C]$ indexed by $c \in [C]$ before training.
We introduce a feature anchor loss to align the each-class features across clients and formulate it as:
\begin{equation}
\begin{aligned}
  l_{{\rm fa}_i}({\bm{\theta}}_i;\mathbf{a}_c,(\textbf{x},c) \in \mathcal{D}_i )   = 
   \| \mathbf{h}_{i,c} - \mathbf{a}_{c}\|^2.
    \label{fa loss}
\end{aligned}
\end{equation}
 where $  \mathbf{h}_{i,c} $  denotes the feature mapped by a feature extractor $f_{\bm{\theta}_i}(\textbf{x}) $   for a  sample $(\textbf{x},c)$.
The feature anchor loss measures the average distance between features and their corresponding feature anchors. 
When minimizing (\ref{fa loss}),  the intra-class feature distance for a given client, as well as across all clients, can be reduced since the anchors are the same across clients, as shown in Figures \ref{toy_example}(b) and \ref{Experimental Validation1}(c). 
Thereafter, we show the whole training process of FedFA  as follows.

\subsubsection{Minimizing local objective with feature anchor loss} 
 In client local objectives, FedFA introduces the feature anchor loss in addition to a supervised loss (e.g., the cross-entropy loss as shown in (\ref{sup loss}) represented as ${l}_{\rm sup} $).
At the start of the $t$-th  round, the server sends the current global model $\mathbf{w}^{(t-1)}$ and feature anchors $\{\mathbf{a}^{(t-1)}_c\}_{c=1}^C$  to a set $\mathcal{S}$ of active clients.
  Each client then $i \in \mathcal{S}$ locally updates  $\mathbf{w}^{(t-1)}$
to $\mathbf{w}_i^{(t)}$ by optimizing the following local objective:
\begin{equation}
\begin{aligned}
      \min_{\mathbf{w}_i }  \mathcal{L}_i(\mathbf{w}_i ) 
      :=\mathbb{E}_{(\textbf{x},y)  \in \mathcal{D}_i}[ l_{{\rm sup}_i}(\mathbf{w}_i)  + \mu l_{{\rm fa}_i} ({\bm{\theta}}_i)]
   \label{local optimization} 
\end{aligned}
\end{equation}
where ${\bm{\theta}_i}\in \mathbf{w}_i=\{{\bm{\theta}_i}, {\bm{\phi}_i}\}$ and  $\mu$ is a hyper parameter to balance ${{l}_{\rm sup}}_i $ and $l_{{\rm fa}_i}$.
The feature inconsistency across clients found in Figure \ref{Experimental Validation1} can be alleviated by minimizing the feature anchor loss $l_{{\rm fa}_i}$ at each mini-batch update.

\subsubsection{Calibrating local classifiers with feature anchors} 
Beyond aligning features, feature anchors are also used to calibrate the updates of classifier proxies.
Specifically, at the end of each mini-batch update,   the active client $i \in \mathcal{S}$ takes feature anchors $\{\mathbf{a}^{(t-1)}_c\}_{c=1}^{C}$ as  one mini-batch input of its classifier $f_{\bm{\phi}_i}$ and  their corresponding classes as the label set ${[C_i]}$ to calibrate classifiers by the following objective:
\begin{equation}
\begin{aligned}
  \min_{\bm{\phi}_i} {l}_{{\rm cal}_i} ({\bm{\phi}_i};\mathbf{a}_{c}) 
= -\frac{1}{C} \sum_{c \in {C} } \log
\frac{\exp \left(\bm{\phi}_{i,c}^\top \mathbf{a}_{c}\right)}{\sum_{z=1}^{C} \exp \left(\bm{\phi}_{i,z}^\top \mathbf{a}_{c}\right)},
\end{aligned}
\label{calibration loss}
\end{equation}
where   $l_{{\rm cal}_i}$ is the classifier calibration loss.
The loss corrects the classifier divergence and keeps classifiers similar at the beginning of each mini-batch update by reducing the distance between the $c$-th class proxy and feature anchor.
The calibration corrects classifier divergence and then mitigates feature inconsistency across clients, as their relationship demonstrated in Theorem \ref{vicious cycle}.  
In return,   consistent features aid in aligning classifiers across clients
as illustrated in Figure \ref{toy_example}(b).

\subsubsection{Fixing feature anchors in local training but computing their momentum} 
 Feature anchors $\{\mathbf{a}^{(t-1)}_c\}_{c=1}^C$ are fixed in local training to keep the feature space consistent across clients under heterogeneous data,  instead of being updated by gradient descent.
To obtain the latest state of the feature space, clients compute the momentum of class features in local training, and the server aggregates the momentum to update feature anchors at the end of one round.
Specifically, 
although client $i$ does not update $\mathbf{a}^{(t-1)}_{c}$, it accumulates the $c$-th class  features of the $\tau$-th batch $\mathcal{B}^{(t,{k_\tau})}_{i}$    as:
\begin{equation}
    \mathbf{m}^{(t,k_\tau)}_{c,i}= \mathbf{m}^{(t,k_{\tau-1})}_{c,i} +  \frac{1}{B|\mathcal{B}^{(t,{k_\tau})}_{i,c}|}\sum_{(\mathbf{x},c)\in \mathcal{B}^{(t,{\tau})}_{i,c}}  f_{\bm{\theta}_i}(\mathbf{x})
    \label{moving average}
\end{equation}
where $B$ represents the total  mini-batch number of one epoch and $\mathbf{m}^{(t,k_0)}= 0$. 
Furthermore, we take  epoch momentum $\mathbf{m}^{(t,k)}_{c,i}$ to estimate the class features by 
\begin{equation}
    \mathbf{\bar{a}}_{c,i}^{(t,k)} = \lambda \mathbf{m}^{(t,k-1)}_{c,i} +  (1-\lambda)\mathbf{m}^{(t,k)}_{c,i}.
    \label{estimation average}
\end{equation}
The estimation reduces the computation overhead of FedFA since it does not need to compute the latest class feature with the training dataset after local training.

\subsubsection{Feature anchor and model aggregation at server} The server performs weighted averaging on all the $c$-th  class  feature $ \mathbf{\bar{a}}^{(t,K)}_{c,i}$ of active clients to generate the next-round feature anchors $\{\mathbf{a}^{(t)}_c\}_{c=1}^C$, where $K$ represents the total number of the local epoch.
The update of feature anchors is represented as  $  \mathbf{a}_{c}^{(t)} =  \sum_{i \in \mathcal{S}} \frac{n_i}{n_s} \mathbb{E}(\mathbf{{a}}_{c,i})  = \sum_{i \in \mathcal{S}} \frac{n_i}{n_s}  \mathbf{\bar{a}}_{c,i}^{(t,K)}$.
Meanwhile,   model aggregation in FedFA is the same as FedAvg, i.e., the global model is   $\boldsymbol{w} =\sum_{i=1}^{N} \frac{n_k}{n}\boldsymbol{w}_i$.  

Note that the above four procedures are a one-round process of FedFA.
FedFA performs the process along rounds until the global model converges, where Algorithm \ref{alg:example} illustrates the pseudo-code of FedFA.

\begin{algorithm}[t]
   \caption{FedFA (Proposed Framework)}
   \label{alg:example}
\begin{algorithmic}
   \STATE {\bfseries  Input:} initial model $\mathbf{w}=\{{\bm{\theta}},  {\bm{\phi}}\}$, initial feature anchors $\{\mathbf{a}_c\}_{c=1}^C$, learning rate $\eta$, local epoch \textit{K},  client number \textit{N}, class number $C$
   \FOR{ each round $t=1,\cdots, R$  }
 
\STATE Server    samples clients $\mathcal{S} \subseteq \{1,\cdots,N\}$
   \STATE Server   communicates  $\mathbf{w}^{(t-1)}$ and  $\{\mathbf{a}_c^{(t-1)}\}_{c=1}^C$ to all clients $i\in \mathcal{S}$
   \STATE \textbf{on client }$i\in \mathcal{S}$ \textbf{in parallel do}
   \STATE \quad Initialize the local model $\mathbf{w}_i \gets \mathbf{w}^{(t-1)}$,  the local feature anchor $\mathbf{a}_{c,i} \gets \mathbf{a}_c^{(t-1)}$ 
 
    \STATE \quad {\textbf{for} local epoch $k=1,\cdots, K$ \textbf{do}}
     \STATE \quad  \quad  {\textbf{for} each mini-batch  \textbf{do}}
   \STATE \quad  \quad  \quad $\%$ \textit{feature alignment with feature anchors} $\%$ 
    \STATE \quad  \quad  \quad  Calculate the local loss $l_i \gets l_{{\rm sup}_i} +\mu l_{{\rm fa}_i}  $ 
        
        \STATE \quad  \quad  \quad   Compute mini-batch gradient $g_i(\mathbf{w}_i) \gets  \nabla_{\mathbf{w}_i}  l_i$
    
    \STATE \quad  \quad  \quad Update local model $ \mathbf{w}_i - \eta g_i(\mathbf{w}_i)$
     \STATE \quad  \quad  \quad $\%$ \textit{classifier calibration with feature anchors} $\%$ 
    \STATE \quad  \quad  \quad  Calculate the calibration loss $l_i \gets l_{{\rm cal}_i} $ 
    
       \STATE \quad  \quad  \quad   Compute mini-batch gradient $g_i(\bm{\phi}_i) \gets  \nabla_{\bm{\phi}_i}  l_{{\rm cal}_i}$
    
    \STATE \quad  \quad  \quad Calibrate classifier proxies $ \bm{\phi}_i - \eta g_i(\bm{\phi}_i)$
    
     \STATE \quad  \quad  \quad $\%$ \textit{Accumulate class features} $\%$ 
    \STATE \quad  \quad  Estimate feature anchors $\mathbf{m}_{i}^{(t,k)}$
 
    

    \STATE \quad  \textbf{end for}
    \STATE \quad  Communicate $\mathbf{w}_i^{(t)} $ and $\{\mathbf{a}_{c,i}^{(t)}\}_{c=1}^C$ back to the server
    \STATE   \textbf{end on client}
    \STATE Server aggregates the global model $\mathbf{w}^{(t)} \gets \frac{1}{|\mathcal{S}|}\sum_{i\in \mathcal{S}}\mathbf{w}_i^{(t)}$, and the feature anchors $\mathbf{a}_c^{(t)} \gets \frac{1}{|\mathcal{S}|}\sum_{i\in \mathcal{S}}\mathbf{a}_{c,i}^{(t)}$ with weighted averaging
   \ENDFOR
 
\end{algorithmic}
\end{algorithm}

\subsection{Non-convex Convergence analysis of FedFA}

To show the convergence results, we first make the following commonly used assumptions as per \cite{wang2020tackling, tan2022fedproto}.

\begin{assumption}\label{ass-1}
(Lipschitz smoothness)
 Each local objective function is Lipschitz smooth, that is,   $\| \nabla \mathcal{L}_i(\mathbf{w}_1) - \nabla \mathcal{L}_i(\mathbf{w}_2) \|_2 \leq L_1 \| \mathbf{w}_1 - \mathbf{w}_2 \|_2$,  ${\forall} i \in \{1,2,\dots,N\}$.
\end{assumption}

\begin{assumption}\label{ass-2}
(Unbiased gradient with bounded  variance)
For any stochastic gradient $ \mathbf{g}_i(\mathbf{w}) $, there exists a constant $\sigma $ such that $\mathbb{E}\left[  \mathbf{g}_i(\mathbf{w} ) \right] = \nabla \mathcal{L}_i(\mathbf{w})$ and $\mathbb{E}\left[ \|  \mathbf{g}_i(\mathbf{w} ) - \right.$$ \left. \nabla \mathcal{L}_i(\mathbf{w}) \|^2 \right] \leq \sigma^2$,  ${\forall} i \in \{1,2,\dots,N\}$.
\end{assumption}

\begin{assumption}\label{ass-3}
(Bounded expectation of norm  of stochastic gradient) 
The expectation of any  stochastic gradient norm is bounded by $G$ such that 
$\mathbb{E}\left[ \| \mathbf{g}_i(\mathbf{w} ) \|_2^2 \right] \leq G^2$,  ${\forall} i \in \{1,2,\dots,N\}$.

\end{assumption}

\begin{assumption}\label{ass-4}
(Lipschitz continuity of feature extractors) 
Each local feature-extractor  function    is $L_2$-Lipschitz continuous, that is, $\|  f_i(\bm{\theta}_{1}) -  f_i(\bm{\theta}_{2}) \|_2 \leq L_2 \| \bm{\theta}_{1} - \bm{\theta}_{2} \|_2$,  ${\forall} i \in \{1,2,\dots,N\}$.
\end{assumption}

According to the update of FedFA at one round, we have:
\begin{lemma}
    (One-round loss deviation of FedFA. See proof in Appendix \ref{proof:FedFA_convergence}).
Let Assumption \ref{ass-1}-\ref{ass-4} hold, for a one-round update of FedFA, we denote as the total iteration number of one-client updates as $\tau_K$ and have: 
   $ \mathbb{E}  \left[\mathcal{L}\left(\mathbf{w}^{(t+1,0)}\right)\right]-\mathcal{L}\left(\mathbf{w}^{(t, 0)}\right)     \leq     \frac{-\tau_K \eta}{2}\left\|\nabla \mathcal{L}\left(\mathbf{w}^{(t,0)}\right)\right\|^2 +  \tau_K L_1 \eta^2 \sigma^2
   +  \frac{G^2 L_1^2 \tau_K^2 \eta^3}{4}   + \mu  L_2^2 \eta^2  \tau_K G^2 $.
\label{lemma:one round}
\end{lemma}
 
\begin{theorem}
(Non-convex convergence rate of FedFA.  See proof in Appendix \ref{proof:FedFA_convergence}).
With Lemma \ref{lemma:one round}, let $\Delta = \mathcal{L}\left(\mathbf{w}^{(0)}\right)     -  \mathcal{L}\left(\mathbf{w}^{(*)}\right) $ where $\mathbf{w}^{(*)}$ denotes the local optimum. Given any $\epsilon$, we have  $ \frac{1}{T}\sum_{t=0}^{T-1}  \left\|\nabla \mathcal{L}\left(\mathbf{w}^{(t,0)}\right)\right\|^2 < \epsilon$, when    the total communication rounds  $T$ of FedFA meet  $ T >  \frac{8\Delta  }{4\tau_K \eta  \epsilon -  \tau_K \eta  (4  \eta L_1   \sigma^2    +    \tau_K     \eta^2 L_1^2   G^2    +    4\mu  \eta    L_2^2G^2 ) }$, $\eta < \min \{\frac{4 \epsilon }{4 L_1 \sigma^2 +  \tau_K  G^2 L_1^2  +   4\mu  L_2^2  G^2}, \frac{1}{2\tau_KL_1}\}$ and $\mu < \frac{ \epsilon -   L_1 (\sigma^2 + L_1 G^2)}{ L_2^2G^2}$.
\label{theorem_convergence}
\end{theorem}

\subsection{Loss smoothness analysis of FedFA.}
  With the feature polymerization under feature anchor loss (\ref{fa loss}) \cite{wen2016discriminative}, we assume $\overline{\mathbf{h}}_{\cdot,c}=\mathbf{a}_c$ in the following analysis.
\begin{theorem}
(\textit{The effect of FedFA on the Lipschitzness of the loss} on classifier weight. See proof in Appendix \ref{proof:Lipschitzness improvement}.) Let $ \|\nabla_{\bm{{\phi}}_{c}}\hat{\mathcal{L}}\|^2$ and $ \|\nabla_{\bm{{\phi}}_{c}}{\mathcal{L}}\|^2$ be the gradient norms of FedFA and FedAvg, respectively. For $\mathbf{a}_{{c}}\cdot {\mathbf{a}}_{\bar{c}}=0$, the deviation of gradient norms of the global classifier between FedFA and FedAvg is computed as:
\begin{equation*}
\begin{aligned}
\|\nabla_{\bm{{\phi}}_{c}}\hat{\mathcal{L}}\|^2 - \|\nabla_{\bm{{\phi}}_{c}}{\mathcal{L}}\|^2 =\|\Delta \bm{\hat{\phi}}_{c}\|^2 - \|\Delta \bm{{\phi}}_{c}\|^2 \\ =\frac{\eta^2}{n^2}[ {(\hat{A}_c^2 - {A}_c^2)}\|{\mathbf{a}}_{c}\|^2 + \sum_{\bar{c}\neq c}{(\hat{B}_{\bar{c}}^2 - {B}_{\bar{c}}^2)}\|{\mathbf{a}}_{\bar{c}}\|^2 < 0
\end{aligned}
\end{equation*}
where 
${A}_c=\sum_{i}^{N} \left( n_{i,c}(1-\overline{{p}_{i, c}^{(c)}})\right)$,
$\hat{A}_c=\sum_{i}^{N} \left( n_{i,c}(1-\overline{\hat{p}_{i, c}^{(c)}})\right)$,
$B_{ \bar{c}} = \sum_{i}^{N}  n_{i,\bar{c}} \overline{{p}_{i,{c}}^{(\bar{c})}}$, 
$\hat{B}_{ \bar{c}}  = \sum_{i}^{N}   n_{i,\bar{c}} \overline{\hat{p}_{i,{c}}^{(\bar{c})}}$ and $\hat{A}_c <\hat{A}_c, \hat{B}_{\bar{c}} < \hat{B}_{\bar{c}} $.
Note that $\mathbf{a}_{{c}}\cdot {\mathbf{a}}_{\bar{c}}=0$ provides an orthogonal initialization for feature anchors. 
\label{Lipschitzness improvement}
\end{theorem}

Theorem \ref{Lipschitzness improvement} suggests that the incorporation of feature alignment and classifier calibration in FedFA leads to an improvement in the Lipschitzness of the loss on classifier weight space, as compared to FedAvg. This results in a smoother loss function and accelerates the convergence of FedFA.

\begin{observation}
(\textit{A virtuous cycle in FedFA})
Combining Theorem  \ref{vicious cycle} and Theorem \ref{Lipschitzness improvement}, 
 feature alignment and classifier calibration together smooth the loss of classifier updates to boost classifier harmony across clients, which in turn promotes feature mapping consistency across clients, as illustrated in the toy example of Figure \ref{toy_example}(b).
\label{virtuous cycle}
\end{observation}
Different from Observation \ref{observation1},  FedFA breaks the \textit{vicious cycle} to obtain a \textit{virtuous cycle} between feature and classifier updates under both label distribution skew and feature distribution skew, as shown in Figures \ref{Experimental Validation}(b2) and (b3). 

\subsection{Computational Overhead  of FedFA}
Assuming $N$ clients participate in federated learning, we aim to train a fully connected neural network for the sake of simplicity,  which can be extended to other network architectures \cite{molchanov2016pruning}.
The computational overhead   with $L$ layers is represented as $G_1 = \mathcal{O}\left(\sum_l\left(2 I_l-1\right) I_{l+1}\right)$, where $l \in[0, L]$ denotes the layer index and $I_l$ represents the neuron number of hidden layers. 
The computational overhead of feature anchor loss (\ref{fa loss}) is $G_2 = \mathcal{O}\left(I_{L-1}\right)$,  depending on the feature dimension.
The computation of classifier calibration is $G_3 = \mathcal{O}\left((2 I_{L-1}-1)I_{L}\right)$, while the computation of  model averaging is $G_4 = \mathcal{O}\left(\sum_l I_{l}\right)$.

We then analyze the computational overhead of FedFA based on the client and server sides and formulate it as follows.
For the client side, the total computation of all client models is $N \times G_1$, the computation of feature anchor loss (\ref{fa loss}) is $N \times G_2$, and the computation of classifier calibration is $N \times G_3$.
For the server side, the computation of model averaging on $N$ client models is $G_4$.
The computation of feature-anchor accumulation on the client side and feature-anchor averaging on the server side is based on the neuron number of the last layer  $I_{L}$, which is small and can be ignored.
In summary,  the total computational overhead of FedFA is $N \times (G_1+G_2+G_3) + G_4$, compared with that of FedAvg  $N \times G_1 + G_4$. 
With low feature dimensions (i.e., small  $I_{L-1}$), the computational overhead of FedFA is similar to  FedAvg.

\subsection{Advantages of FedFA}
When facing heterogeneous data,  FedFA breaks the \textit{vicious cycle}  between feature and classifier updates and brings the \textit{virtuous cycle} between the two with the help of feature anchors.
The anchors help FedFA create a shared feature space across clients and keep classifiers consistent in this space. 
Moreover,  the feature anchor loss of FedFA improves feature polymerization within the same class. 
The polymerization reduces intra-class feature distance and increases inter-class feature distance across clients. 
These advantages result in significant performance benefits, such as improved accuracy in classification, smoother loss and better convergence for a variety of data-heterogeneous tasks.

\section{Experiments}
\label{Experiments}
\subsection{Experimental Setup}

\subsubsection{Datasets and data heterogeneity setups} 
This work aims at image classification tasks under label and feature distribution skews, and it uses federated benchmark datasets  as  \cite{mcmahan2017communication,yurochkin2019bayesian,li2021federated}, including  EMNIST\cite{cohen2017emnist}, FMNIST, CIFAR-10, CIFAR-100 \cite{krizhevsky2009learning}, and  Mixed Digits dataset \cite{li2021fedbn}.
Specifically, for  label distribution skew, we consider two settings: (i) Same size of local dataset: following \cite{mcmahan2017communication}, we split data samples based on class to clients (e.g., $\#C=2$ denotes that each client holds two class samples);
(ii) Different sizes of local dataset:  following \cite{yurochkin2019bayesian}, we set  $\alpha$ of Dirichlet distribution $Dir(\alpha)$ as 0.1 and 0.5 to generate distribution $p_{i,c}$ by which   the $c$-th class samples are split to client $i$.
For feature distribution skew, we consider two settings:
(i) Real-world feature skew: we sample a subset with 10 classes of a real-world dataset  EMNIST  with natural feature skew;
(ii) Artificial feature skew:  we use a mixed-digit dataset from \cite{li2021fedbn}
consisting of  MNIST\cite{lecun1998gradient}, SVHN\cite{netzer2011reading},  USPS\cite{hull1994database}, SynthDigits and MNIST-M\cite{ganin2015unsupervised}.
 In Table \ref{feature skew} and Table \ref{ablation results}, we test the top-1 accuracy based on the global model, except for Mixed Digits where we report the average top-1 accuracy on five-benchmark digit datasets.

\begin{table*}[t]
\centering
\caption{The top-1 accuracy of FedFA and all the baselines under label distribution skew on the test datasets. We run three trials and report the mean and standard deviation. For FedAvg and FedFA, we also report their top-1 accuracy without label skew.}
\label{label skew}
\resizebox{\textwidth}{!}{%
\begin{tabular}{@{}c|ccccccccc@{}}
\toprule
\multirow{3}{*}{\begin{tabular}[c]{@{}c@{}}Method\\ (lr = 0.01)\end{tabular}} & \multicolumn{9}{c}{Label Distribution Skew}                                                                                                                                                                                    \\ \cmidrule(l){2-10} 
& \multicolumn{3}{c|}{FMNIST}                                                             & \multicolumn{3}{c|}{CIFAR-10}                                                           & \multicolumn{3}{c}{CIFAR-100}                                      \\
& $\#C=2$              & $\alpha = 0.1$       & \multicolumn{1}{c|}{$\alpha = 0.5$}       & $\#C=2$              & $\alpha = 0.1$       & \multicolumn{1}{c|}{$\alpha = 0.5$}       & $\#C=20$             & $\alpha = 0.1$       & $\alpha = 0.5$       \\ \midrule
FedAvg w/o skew                                                               & \multicolumn{3}{c|}{85.90(0.14)}                                                        & \multicolumn{3}{c|}{59.66(0.05)}                                                        & \multicolumn{3}{c}{25.37(0.28)}                                    \\
FedFA w/o skew                                                                & \multicolumn{3}{c|}{\textbf{89.67(0.16)}}                                               & \multicolumn{3}{c|}{\textbf{64.95(0.53)}}                                               & \multicolumn{3}{c}{\textbf{33.94(0.44)}}                           \\ \midrule
FedAvg                                                                        & 74.60(1.42)          & 69.81(3.00)          & \multicolumn{1}{c|}{82.80(0.65)}          & 36.07(3.02)          & 35.20(3.72)          & \multicolumn{1}{c|}{48.66(3.00)}          & 22.62(0.84)          & 21.79(0.79)          & 26.52(1.09)          \\
FedProx                                                                       & 74.63(1.30)          & 69.59(2.99)          & \multicolumn{1}{c|}{82.92(0.38)}          & 36.63(2.64)          & 35.21(3.78)          & \multicolumn{1}{c|}{48.43(2.27)}          & 22.27(0.90)          & 22.30(0.47)          & 26.03(0.73)          \\
FedDyn                                                                        & 74.77(1.76)          & 70.09(2.24)          & \multicolumn{1}{c|}{83.95(0.29)}          & 36.11(3.35)          & 36.00(3.78)          & \multicolumn{1}{c|}{50.46(2.33)}          & 13.28(2.19)          & 1.00(0.00)           & 1.00(0.00)           \\
MOON                                                                          & 74.25(1.59)          & 68.52(2.26)          & \multicolumn{1}{c|}{82.72(0.42)}          & 35.90(3.17)          & 34.89(3.18)          & \multicolumn{1}{c|}{48.74(2.45)}          & 22.03(1.00)          & 22.04(0.62)          & 26.69(1.03)          \\
FedProc                                                                       & 74.96(1.94)          & 69.80(3.26)          & \multicolumn{1}{c|}{82.94(0.34)}          & 36.57(3.61)          & 35.02(4.53)          & \multicolumn{1}{c|}{48.99(2.85)}          & 23.00(0.35)          & 22.32(0.63)          & 26.38(0.52)          \\
FedFA (Our)                                                                   & \textbf{84.08(1.22)} & \textbf{83.42(1.14)} & \multicolumn{1}{c|}{\textbf{88.40(0.12)}} & \textbf{52.64(1.46)} & \textbf{52.95(2.01)} & \multicolumn{1}{c|}{\textbf{60.40(0.38)}} & \textbf{26.68(1.18)} & \textbf{24.05(2.32)} & \textbf{29.16(1.03)} \\ \bottomrule
\end{tabular}%
 }
\end{table*}

\subsubsection{Baselines}
We consider two popular research branches in enhancing the performance of federated learning as our baselines in addition to the canonical method FedAvg.
One branch is to implement weight regularizers in clients' local objectives by controlling the distance between client models and the global model.
Our two baselines for this branch are FedProx \cite{li2020federated} and FedDyn \cite{acar2021federated}.
Different from this branch, our method FedFA controls the distance by aligning features and calibrating classifiers across clients.
A similar branch to FedFA is to align features extracted by client models across clients, where    MOON \cite{li2021model} and FedProc \cite{mu2021fedproc} are considered as our baselines.
However, these two methods do not calibrate classifiers across clients since they overlook the vicious circle between feature and classifier updates.
Note that our method exclusively modifies the client-side local updates.
Therefore, server-side-based methods, such as improving aggregation schemes  \cite{wang2020tackling}  and adding server-update momentum \cite{reddi2021adaptive}, are complementary to   FedFA and not considered as our baselines.

 Moreover, we carefully select the coefficient of local regularization from $\{1, 0.1, 0.01\}$ (i.e., $\mu/2=0.05$ for FedProx and FedDyn, $\mu=1$ for MOON except $\mu=5$ on CIFAR-10), set the temperature hyperparameter $\tau=0.5$ for MOON and FedProc, and report their best results in our experiments.
 
\subsubsection{Models} 
To ensure a fair comparison, our models adhere to the reported baselines.
 Following \cite{acar2021federated}, we use a CNN model with two convolution layers for EMNIST, FMNIST, and CIFAR-10.
We utilize ResNet-18 \cite{he2016deep} with a linear projector from \cite{li2021model}  for CIFAR-100  and a CNN model with three convolutional layers from  \cite{li2021fedbn} for  Mixed Digits.

\subsubsection{FedFA setup} 
We set the coefficient of exponential
moving average $\lambda=  0.5$  in  momentum accumulation for feature anchors in local training and local loss coefficient $\mu=0.1$ in (\ref{local optimization}).
For anchor initialization, we initiate the pairwise orthogonal feature anchors $\mathbf{a}_c$  by sampling column vector from an identity matrix whose dimension is the same as the size of the features.
Other settings of FedFA are the same as baselines in all experiments.
 
\subsubsection{Federated simulation setups}
In Tables \ref{label skew}  and  \ref{feature skew}, 100 clients attend federated training,  10 clients participate in each round, the local batch size is  64, the local epochs number is 5, and the targeted communication round is 200.
We use the SGD optimizer with a 0.01 learning rate and 0.001 weight decay for all experiments. 
Furthermore, in Table \ref{local SGD with momentum1} and Figures \ref{federated skew},  we follow the setups of \cite{li2021model}  to investigate the impact of different federated setups    with 200 rounds and a local SGD with a 0.01 learning rate and 0.9 momentum. 
  All experiments are performed based on PyTorch   and one node of the High-Performance Computing platform with 4 NVIDIA A30 Tensor Core GPUs with 24GB.
 
 
 \begin{table}[t]
\centering
\caption{The top-1 accuracy of all methods under label $\&$ feature distribution skews. Note that we report the average top-1 accuracy on five-digit datasets in Mixed Digits.  The first horizontal frame reports the results of w/o skew.}
\label{feature skew}
\resizebox{0.49\textwidth}{!}{%
\begin{tabular}{@{}c|cclccc@{}}
\toprule
\multirow{3}{*}{\begin{tabular}[c]{@{}c@{}}Method\\ \end{tabular}} & \multicolumn{3}{c|}{Feature Distribution Skew} & \multicolumn{3}{c}{Label $\&$ Feature Distribution Skew} \\ \cmidrule(l){2-7} 
 & \multirow{2}{*}{EMNIST} & \multicolumn{2}{c|}{\multirow{2}{*}{Mixed Digits}} & \multicolumn{3}{c}{Mixed Digits} \\
 &  & \multicolumn{2}{c|}{} & $\#C=2$ & $\alpha = 0.1$ & $\alpha = 0.5$ \\ \midrule
FedAvg  & \textbf{-} & \multicolumn{5}{c}{82.66(2.38)} \\
FedFA & \textbf{-} & \multicolumn{5}{c}{\textbf{88.10(0.39)}} \\ \midrule
FedAvg & 98.50(0.04) & \multicolumn{2}{c|}{82.66(2.83)} & 56.13(5.59) & 63.74(2.35) & 78.34(1.58) \\
FedProx & 98.44(0.06) & \multicolumn{2}{c|}{82.46(2.65)} & 54.86(5.80) & 62.57(2.22) & 78.08(1.84) \\
FedDyn & 97.63(0.19) & \multicolumn{2}{c|}{83.59(2.33)} & 51.66(7.33) & 63.55(2.02) & 79.40(1.76) \\
MOON & 98.51(0.06) & \multicolumn{2}{c|}{81.46(2.84)} & 55.40(5.69) & 62.18(1.93) & 78.05(1.82) \\
FedProc & 98.28(0.04) & \multicolumn{2}{c|}{82.06(2.68)} & 59.53(3.66) & 64.59(2.15) & 78.66(1.16) \\
FedFA & \textbf{99.28(0.33)} & \multicolumn{2}{c|}{\textbf{90.73(2.01)}} & \textbf{83.46(2.57)} & \textbf{85.71(0.71)} & \textbf{89.82(0.49)} \\ \bottomrule
\end{tabular}%
}
\end{table}

\subsection{Experiment Results}
\subsubsection{Performance under label distribution skew} 
Table \ref{label skew} shows that FedFA provides significant gains in different label-skew settings regardless of the dataset. 
Compared with $\alpha=0.5$,  both $\#C=2$ and $\alpha=0.1$ indicate  more severe  label distribution skew,  but clients under $\#C=2$ have the same sample number while the ones with $\alpha=0.1$ do not.
Firstly, we find that the performance of all methods degrades as the degree of data heterogeneity increases. 
Nevertheless, the decline of FedFA is much smaller than that of other methods. 
For example, when $\alpha$ changes from $0.5$ to $0.1$,  the top-1 accuracy of all the baselines goes down by about $13 \%$ on FMNIST and CIFAR-10, which is twice as large as FedFA.
Secondly, under the same label skew, FedFA achieves larger gains over other methods when label distribution skew becomes more severe, up to $18.06\%$ (i.e., MOON: $34.89\%$ and FedFA $52.95\%$ under $\alpha=0.1$ in CIFAR-10).
Thirdly, to explore more difficult tasks, we test on CIFAR-100 with ResNet18, and our method still achieves the best performance (i.e., about $3\%$ accuracy advance).
It is important to note that FedDyn exhibits unstable performance compared to other methods when considering the same hyperparameter setups. 
For instance, it shows significantly low accuracy in the CIFAR-100 task. 
This is due to the fact that FedDyn is much more sensitive to hyperparameters.

\subsubsection{Performance under feature distribution skew}  According to Table \ref{feature skew}, our method obtains higher accuracy than all baselines on EMNIST and Mixed Digits.
 Specifically, the accuracy of FedFA in EMNIST reaches $99.28\%$, which is $0.77\%$ higher than the best baseline (i.e., MOON $98.51\%$). 
Moreover, we split each digit dataset of Mixed Digits into 20 subsets, one for each client with the same sample number  (i.e., a skewed feature distribution exists between the clients with a subset of SVHN and the ones with a subset of MNIST).
Compared with the best baseline (Feddyn: $83.59\%$) on Mixed Digits, our method achieves performance gains by $7.14\%$.

 \begin{figure}[t]
\begin{center}
\includegraphics[width=0.49\textwidth]{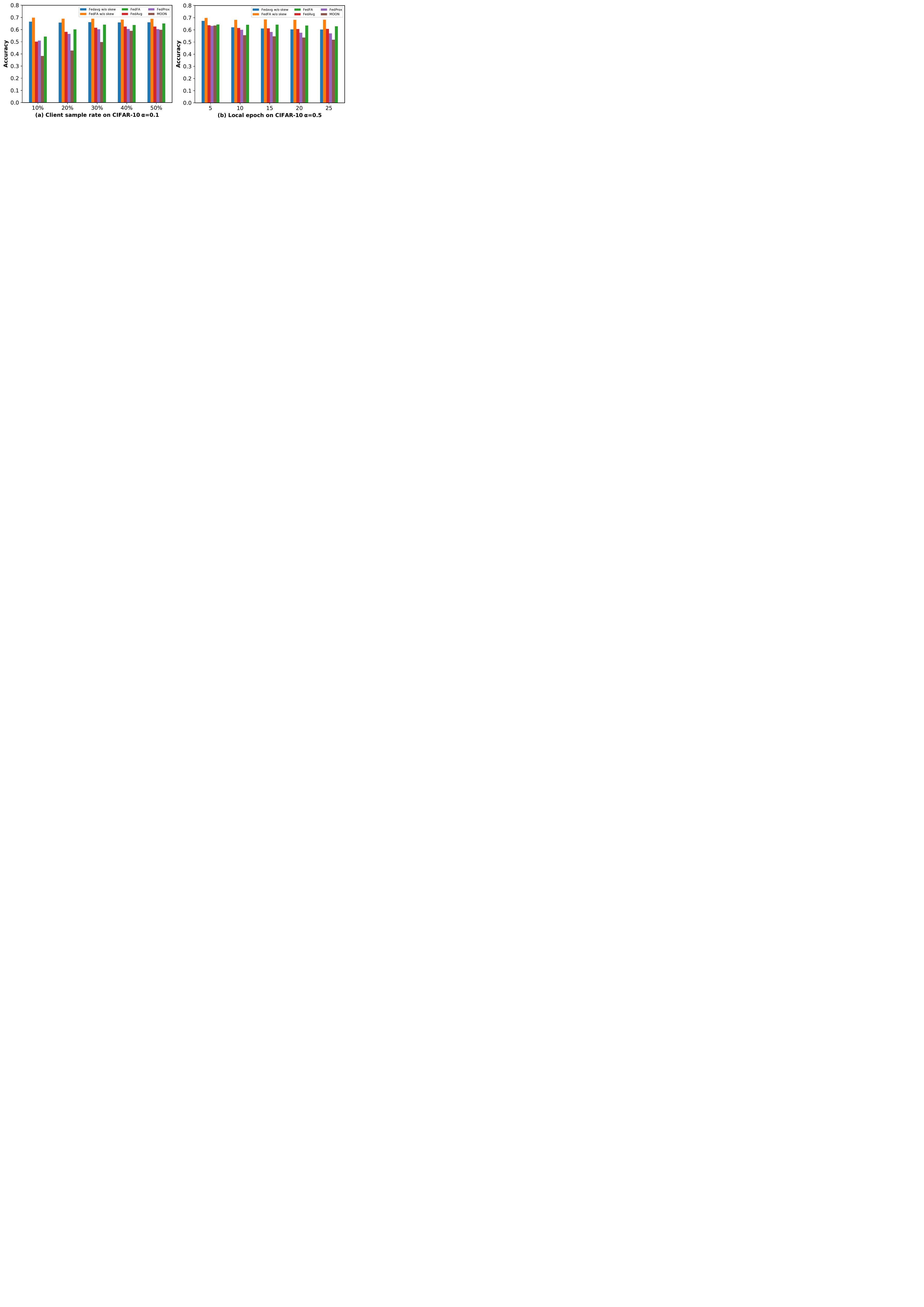}
\end{center}
\vspace{-15pt}
\caption{Test accuracy under different client sample rates and local epochs.}
\vspace{-10pt}
\label{federated skew}
\end{figure}

\subsubsection{Performance under both label and feature skews}
We combine label skew and feature skew to explore the impact of data heterogeneity further.
Namely, we not only split each dataset in Mixed Digits into 20 subsets, one for each client but also set  different label distributions for various clients (i.e., clients are subject to at least one of label distribution skew and feature distribution skew). 
The results in Table \ref{feature skew} show that all the methods are more susceptible under this setting than that of feature distribution skew. 
For example, the most significant performance drop reaches $31.93\%$  (i.e., FedDyn from $83.59\%$ to $51.66\%$ under $\#C = 2$).
Nevertheless, FedFA significantly mitigates this performance degradation with a mild decrease from $90.73\%$ to $83.46\%$. 
Meanwhile, FedFA maintains at least $10\%$ performance advantage over all baselines under this case,  with the largest gap reaching $31.8\%$ (i.e., FedFA from $83.59\%$ to FedDyn $51.66\%$ under $\#C = 2$).

\subsubsection{Performance without label and feature distribution skew} 
We compare our method with FedAvg under more homogeneous data and take the same learning rate of this case as that of data heterogeneity for comparison, where the results are reported in Table \ref{label skew} and Table \ref{feature skew}.
The results demonstrate that FedFA still brings a significant advance in the presence of data homogeneity.
For example, FedFA is  $8.64\%$ more accurate than FedAvg on  CIFAR-100.
Incredibly,  FedFA under mild data heterogeneity (e.g.,   $\alpha =0.5$ in FMNIST or Mixed Digits) even obtains higher accuracy than FedAvg without any label or feature skew (e.g., FedFA: $88.40\%$ vs. FedAvg: $85.90\%$ in FMNIST).
This reveals that the effect of data heterogeneity on federated learning deserves to be further explored.

\begin{table}[t]
\centering
\caption{Test accuracy under local SGD with momentum.}
\resizebox{0.37\textwidth}{!}{%
\begin{tabular}{@{}c|ccc@{}}
\toprule
\multirow{2}{*}{\begin{tabular}[c]{@{}c@{}}Method\\ \end{tabular}} & \multicolumn{3}{c}{CIFAR-10} \\
 & $\#C=2$ & $\alpha=0.1$ & $\alpha=0.5$ \\ \midrule
FedAvg w/o skew & \multicolumn{3}{c}{67.58(0.23)} \\
FedFA w/o skew & \multicolumn{3}{c}{\textbf{69.32(0.36)}} \\ \midrule
FedAvg & 48.17(3.40) & 47.91(5.95) & 64.12(1.02) \\
FedProx & 48.14(2.89) & 49.87(6.82) & 63.71(1.15) \\
MOON & 48.13(2.12) & 47.11(6.96) & 64.08(1.20) \\
FedFA (Our) & \textbf{57.30(2.05)} & \textbf{54.21(5.55)} & \textbf{64.63(0.57)} \\ \bottomrule
\end{tabular}%
\label{local SGD with momentum1}
}
\end{table}

\begin{table}[t]
\centering
\caption{The top-1 accuracy of FedFA in different ablations on anchor updating (AU) and classifier calibration (CC).}
\label{ablation results}
\resizebox{0.48\textwidth}{!}{%
\begin{tabular}{@{}c|c|c|c@{}}
\toprule
\begin{tabular}[c]{@{}c@{}}Method\\ \end{tabular} & \begin{tabular}[c]{@{}c@{}}Label Skew\\ (FMNIST $\#C=2$)\end{tabular} & \begin{tabular}[c]{@{}c@{}}Feature Skew\\ (Mixed Digits)\end{tabular} & \begin{tabular}[c]{@{}c@{}}Label $\&$ Feature Skew\\ (Mixed Digits $\#C=2$)\end{tabular} \\ \midrule
FedFA w/o AU                                    & 81.89(1.87)                                                           & 88.69(0.75)                                                           & 76.81(1.78)                                                                              \\
FedFA w/o CC                            & 78.07(2.23)                                                           & 79.25(1.25)                                                           & 61.36(4.00)                                                                              \\
\midrule
FedFA  (Our)                                                      & \textbf{84.08(1.22)}                                                  & \textbf{90.86(1.92)}                                                   & \textbf{83.73(2.76)}                                                                     \\ \bottomrule
\end{tabular}%
}
\end{table}

\subsubsection{Performance on different client sample rate and local epoch}
Following the setup of Table \ref{local SGD with momentum1}, we further explore the impact of federated setups.
As shown in Figure \ref{federated skew}(a), a larger client sample rate achieves better test accuracy for all methods. Especially, the accuracy gains (about $10\%$) when increasing the sample rate from $0.1$ to $0.3$ is much larger than that from $0.3$ to $0.5$. 
As shown in Figure \ref{federated skew}(b), larger local epochs have a negative impact on performance, but FedProx and MOON have worse performance degradation than FedAvg and FedFA.
Besides, the performance advantage of FedFA under various batch sizes and client numbers is shown in Figures \ref{federated skew1} and \ref{Performance under Different client numbers}.
Overall, our method  FedFA consistently achieves better than all baselines under different setups.

\subsubsection{Performance of Lipschitzness of loss}
   In \cite{yu2019linear,lin2020dont},  it has been found that the local optimizer with momentum is more robust to different smoothness of loss to improve generalization.
We explore the effect of FedFA on the Lipschitzness of loss by comparing local SGD optimizers with or without momentum on CIFAR-10.
Note that  FedDyd and FedProc are not compatible with local SGD with momentum, and thus Table \ref{local SGD with momentum1} and Figure \ref{federated skew} do not include their results.
Comparing Table \ref{label skew} with Table \ref{local SGD with momentum1}, all methods with momentum work better than that without momentum, but FedFA without momentum has superior performance than baselines with momentum under $\#C=2$ and $\alpha=0.1$ (e.g., FedFA without momentum: 52.95$\%$ vs. FedProx with momentum: 49.87$\%$ under $\alpha=0.1$).
As expected by Theorem \ref{Lipschitzness improvement}, a smoother loss of FedFA achieves better generalization.

\subsection{Ablation Studies}\label{Ablation Studies}


  \begin{figure}[t]
     \centering
     \includegraphics[width=0.5\textwidth]{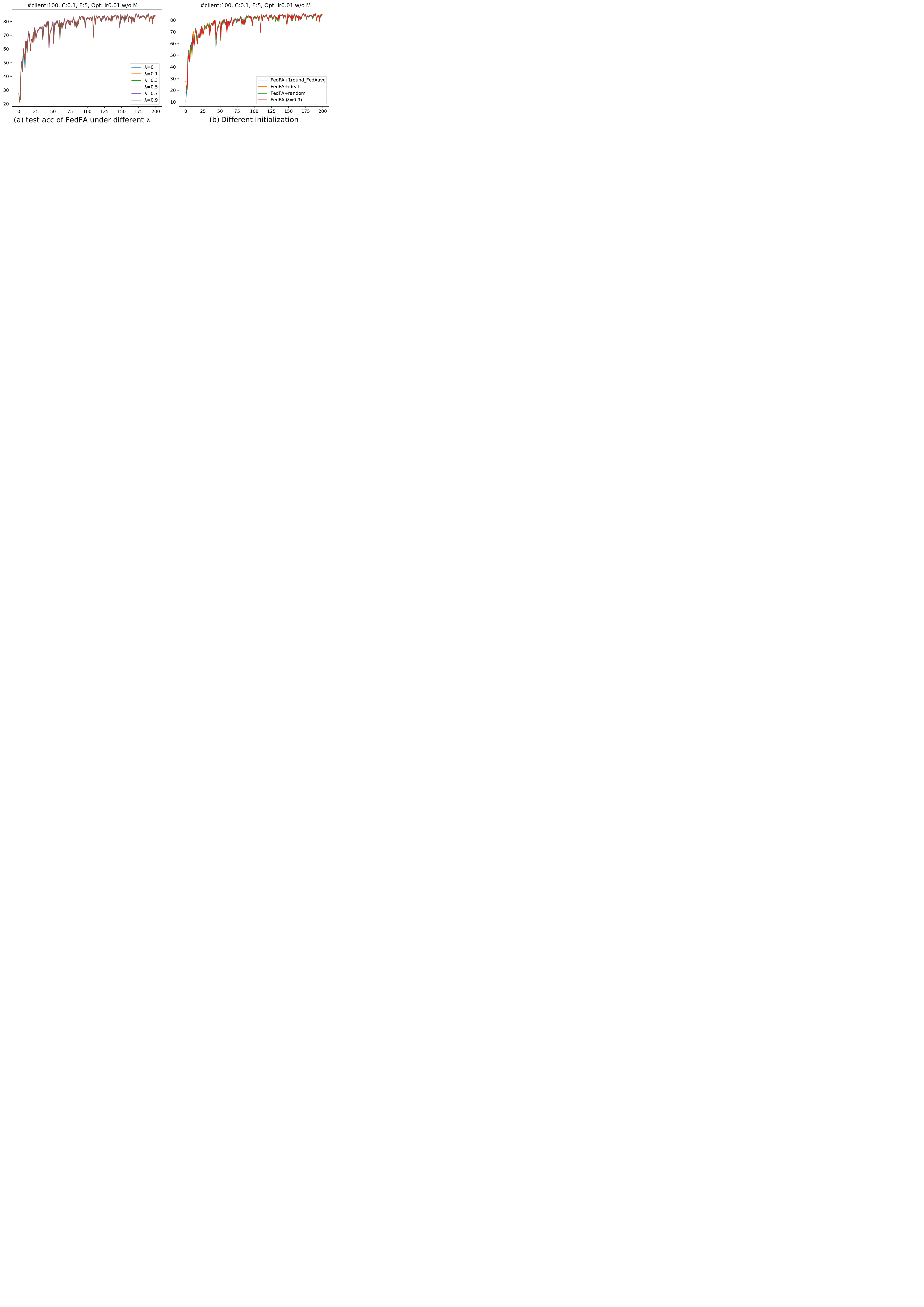}
          \centering
     \caption{Test accuracy (\textit{y-axis}) along the training communication round (\textit{x-axis}). Figure \ref{anchor lambda and init}(a) shows the performance of  FedFA with different $\lambda$ on momentum updates of feature anchors. 
     Figure \ref{anchor lambda and init}(b) shows the performance of  FedFA on feature anchors with different initialization.}
     \label{anchor lambda and init}
 \end{figure}

\subsubsection{Ablation on anchor updates and classifier calibration of FedFA} 
As shown in Table \ref{ablation results}, we conduct ablation studies on  FedFA  without anchor updating in (\ref{local optimization}) and FedFA without classifier calibration in (\ref{calibration loss}) to give an intuition of FedFA performance.
On the one side,  feature anchors can be fixed during federated training (i.e., \textit{the client would not aggregate any information into feature anchors, which would not bring potential privacy leakage}).
FedFA performs better than the best baseline under label and feature skew.
Meanwhile, the anchor updating brings consistent performance benefits (i.e., at least around $2\%$ boost) since the updated anchors keep more representative in the shared feature space across clients.
On the other hand, classifier calibration plays the most crucial role in FedFA because data heterogeneity induces a low classifier update similarity as observed in Figure \ref{Experimental Validation}.
For instance, classifier calibration boosts performance by $22.37\%$ in the case combined by label skew and feature skew.
Overall, both feature alignment and classifier calibration play an essential role in FedFA to overcome data heterogeneity.

\subsubsection{The momentum update of feature anchors}
We experiment on the FMNIST with different momentum coefficients ($\lambda$) of feature anchor updates with the label-skew case of  $\#C=2$.
This experiment setting is the same as Table \ref{label skew}.  
 Specifically,  when $\lambda=1$, feature anchors will not be updated; when $\lambda=0$, feature anchors will be set as the mean feature of the last epoch.
 Figure \ref{anchor lambda and init}(a) shows that the performance of FedFA with different $\lambda$ is similar. 
Meanwhile,  although  FedFA with $\lambda = 0$ introduces more oscillations during training, it performs similarly to other cases.
 This means that  FedFA is not sensitive to the momentum coefficient $\lambda$. 

\subsubsection{The initialization of feature anchors}
To explore the impact of the initialization of feature anchors in FedFA, we design three experiments, including random initialization, one-round-FedAvg initialization (i.e., performing FedAvg but accumulating anchors at first round),  and ``ideal” initialization (i.e., feature anchors are initiated by the trained feature anchors obtained from a finished training of FedFA with the same setting).
 Figure \ref{anchor lambda and init}  reveals that the initialization of the feature anchors does not affect the convergence speed   because the anchors are updated in each communication round so that the impact of initialization of feature anchors is quickly and drastically mitigated. For example, as shown in Figure \ref{anchor lambda and init}, FedFA with orthogonal initialization provides better accuracy in the first round but does not obtain the best accuracy finally.


\subsubsection{Timing to calibrate classifiers}
Compared with \cite{luo2021no} that calibrates classifier with virtual representation (CCVR) \textit{after training},  we perform it
 during different phases of training and the setting of Table \ref{Classifier calibration at different training phase} is the same as Table \ref{label skew}.
The result of each mini-batch calibration done by FedFA is the best in all cases.
This reveals that the classifier divergence induced by data heterogeneity should be corrected as early as possible.
Meanwhile, maintaining the virtuous cycle between feature and classifier updates during local training helps the final model converge at a point that generalizes better, since compared with FedFA without classifier calibration,  FedFA with classifier calibration after training only improves little.

 \begin{table}[t]
\centering
\caption{Classifier calibration (CC) at different phases.}
\label{Classifier calibration at different training phase}
\resizebox{0.3\textwidth}{!}{%
\begin{tabular}{@{}ll@{}}
\toprule
 Method (FMNIST $\#C=2$) & Accuracy \\ \midrule
FedAvg & 73.19 \\
FedAvg w/ CCVR \cite{luo2021no} & 75.95 \\
FedFA w/ CCVR & 84.94 \\ \midrule
FedFA w/o CC & 76.81 \\
FedFA w/ CC after training & 76.94 \\
FedFA w/ CC at the end of each epoch & 82.05 \\
\midrule
FedFA at the end of each mini batch & 84.90 \\ \bottomrule
\end{tabular}%
}
\end{table}  
\section{Conclusion and Future Works}
\label{Conclusion}
This work proposes FedFA, a framework that aims to alleviate performance degradation caused by label and feature distribution skews in federated learning. 
FedFA creates a shared feature space across clients, assisted by feature anchors, and keeps the classifier consistent in this space.
With the help of the shared feature space, FedFA brings a \textit{virtuous cycle} between feature and classifier updates and significantly outperforms baselines on various data-heterogeneity tasks.
The \textit{virtuous cycle} in FedFA provides a fundamental solution to the issue of data heterogeneity in federated learning.
This contrasts previous attempts, which often resulted in a \textit{vicious cycle} between feature inconsistency and classifier divergence. Such attempts only focus on addressing either feature inconsistency or classifier divergence and fail to consider the relationship between the two, leading to a decline in performance.
Overall, this work provides insights into how feature and classifier updates are related under data heterogeneity and proposes that FedFA exploit this relationship to improve federated learning effectively.

In future work, it is interesting to explore further the causes of feature inconsistency in tasks beyond classification.
For example, it is valuable to verify whether a \textit{vicious cycle}  between encoder and decoder updates exists in encoder-decoder-based tasks with heterogeneous data and to extend the observations of this work into more general tasks.
In addition,  a promising direction is to investigate potential improvements to FedFA, such as aligning the features of shallow layers instead of the last layer of the feature extractor, to improve the performance of federated learning training deep models, etc.

\bibliography{example_paper}

\begin{thebibliography}{10}
\providecommand{\url}[1]{#1}
\csname url@samestyle\endcsname
\providecommand{\newblock}{\relax}
\providecommand{\bibinfo}[2]{#2}
\providecommand{\BIBentrySTDinterwordspacing}{\spaceskip=0pt\relax}
\providecommand{\BIBentryALTinterwordstretchfactor}{4}
\providecommand{\BIBentryALTinterwordspacing}{\spaceskip=\fontdimen2\font plus
\BIBentryALTinterwordstretchfactor\fontdimen3\font minus \fontdimen4\font\relax}
\providecommand{\BIBforeignlanguage}[2]{{%
\expandafter\ifx\csname l@#1\endcsname\relax
\typeout{** WARNING: IEEEtran.bst: No hyphenation pattern has been}%
\typeout{** loaded for the language `#1'. Using the pattern for}%
\typeout{** the default language instead.}%
\else
\language=\csname l@#1\endcsname
\fi
#2}}
\providecommand{\BIBdecl}{\relax}
\BIBdecl

\bibitem{mcmahan2017communication}
B.~McMahan, E.~Moore, D.~Ramage, S.~Hampson, and B.~A. y~Arcas, ``Communication-efficient learning of deep networks from decentralized data,'' in \emph{Proc. Int. Conf. Artif. Intell. Statist. (AISTATS)}, Ft. Lauderdale, FL, USA, Apr. 2017, pp. 1273--1282.

\bibitem{li2020federated}
T.~Li, A.~K. Sahu, M.~Zaheer, M.~Sanjabi, A.~Talwalkar, and V.~Smith, ``Federated optimization in heterogeneous networks,'' in \emph{Proc. Mach. Learn. Syst. ({MLSys})}, Austin, TX, USA, Mar. 2020.

\bibitem{karimireddy2020scaffold}
S.~P. Karimireddy, S.~Kale, M.~Mohri, S.~Reddi, S.~Stich, and A.~T. Suresh, ``Scaffold: Stochastic controlled averaging for federated learning,'' in \emph{Proc. Int. Conf. Mach. Learn. (ICML)}, vol. 119, Virtual Event, 2020, pp. 5132--5143.

\bibitem{wang2022accelerating}
Z.~Wang, H.~Xu, J.~Liu, Y.~Xu, H.~Huang, and Y.~Zhao, ``Accelerating federated learning with cluster construction and hierarchical aggregation,'' \emph{IEEE Trans. Mobile Comput.}, vol.~22, no.~7, pp. 3805--3822, Jul. 2023.

\bibitem{sun2023accelerating}
W.~Sun, Y.~Zhao, W.~Ma, B.~Guo, L.~Xu, and T.~Q. Duong, ``Accelerating convergence of federated learning in mec with dynamic community,'' \emph{IEEE Trans. Mobile Comput.}, pp. 1--17, Feb. 2023.

\bibitem{zhao2018federated}
Y.~Zhao, M.~Li, L.~Lai, N.~Suda, D.~Civin, and V.~Chandra, ``Federated learning with non-iid data,'' [Online]. Available \url{https://arxiv.org/pdf/1806.00582.pdf}.

\bibitem{li2021federated}
Q.~Li, Y.~Diao, Q.~Chen, and B.~He, ``Federated learning on non-iid data silos: An experimental study,'' in \emph{Proc. Int. Conf. Data Eng., (ICDE)}, Kuala Lumpur, Malaysia, May 2022, pp. 965--978.

\bibitem{wei2021user}
K.~Wei, J.~Li, M.~Ding, C.~Ma, H.~Su, B.~Zhang, and H.~V. Poor, ``User-level privacy-preserving federated learning: Analysis and performance optimization,'' \emph{IEEE Trans. Mobile Comput.}, vol.~21, no.~9, pp. 3388--3401, Sep. 2021.

\bibitem{kairouz2021advances}
P.~Kairouz, H.~B. McMahan, B.~Avent, A.~Bellet, M.~Bennis, A.~N. Bhagoji, K.~Bonawitz, Z.~Charles, G.~Cormode, R.~Cummings \emph{et~al.}, ``Advances and open problems in federated learning,'' \emph{Found. Trends Mach. Learn.}, vol.~14, no. 1--2, pp. 1--210, 2021.

\bibitem{wang2020tackling}
J.~Wang, Q.~Liu, H.~Liang, G.~Joshi, and H.~V. Poor, ``Tackling the objective inconsistency problem in heterogeneous federated optimization,'' in \emph{Proc. Conf. Adv. Neural Inf. Process. Syst. (NeurIPS)}, Virtual Event, Dec. 2020, pp. 7611--7623.

\bibitem{chen2020joint}
M.~Chen, Z.~Yang, W.~Saad, C.~Yin, H.~V. Poor, and S.~Cui, ``A joint learning and communications framework for federated learning over wireless networks,'' \emph{IEEE Trans. Wirel. Commun.}, vol.~20, no.~1, pp. 269--283, Oct. 2020.

\bibitem{zhang2021optimizing}
W.~Zhang, D.~Yang, W.~Wu, H.~Peng, N.~Zhang, H.~Zhang, and X.~Shen, ``Optimizing federated learning in distributed industrial iot: A multi-agent approach,'' \emph{IEEE J. Sel. Areas Commun.}, vol.~39, no.~12, pp. 3688--3703, Oct. 2021.

\bibitem{xu2022adaptive}
Y.~Xu, Y.~Liao, H.~Xu, Z.~Ma, L.~Wang, and J.~Liu, ``Adaptive control of local updating and model compression for efficient federated learning,'' \emph{IEEE Trans. Mobile Comput.}, vol.~22, no.~10, pp. 5675--5689, Sep. 2023.

\bibitem{li2021model}
Q.~Li, B.~He, and D.~Song, ``Model-contrastive federated learning,'' in \emph{Proc. IEEE/CVF Conf. Comput. Vision Pattern Recognit. (CVPR)}, Virtual Event, Jun. 2021, pp. 10\,713--10\,722.

\bibitem{mu2021fedproc}
X.~Mu, Y.~Shen, K.~Cheng, X.~Geng, J.~Fu, T.~Zhang, and Z.~Zhang, ``{FedProc}: Prototypical contrastive federated learning on non-iid data,'' \emph{Future Gener. Comput. Syst.}, vol. 143, pp. 93--104, Mar 2023.

\bibitem{tan2022towards}
A.~Z. Tan, H.~Yu, L.~Cui, and Q.~Yang, ``Towards personalized federated learning,'' \emph{IEEE Trans. Neural Networks Learn. Syst.}, pp. 1--17, Mar. 2022.

\bibitem{shao2023survey}
J.~Shao, Z.~Li, W.~Sun, T.~Zhou, Y.~Sun, L.~Liu, Z.~Lin, and J.~Zhang, ``A survey of what to share in federated learning: Perspectives on model utility, privacy leakage, and communication efficiency,'' [Online]. Available \url{https://arxiv.org/pdf/2307.10655.pdf}.

\bibitem{acar2021federated}
D.~A.~E. Acar, Y.~Zhao, R.~Matas, M.~Mattina, P.~Whatmough, and V.~Saligrama, ``Federated learning based on dynamic regularization,'' in \emph{Proc. Int. Conf. Learn. Repr. (ICLR)}, Virtual Event, May 2021.

\bibitem{luo2021no}
M.~Luo, F.~Chen, D.~Hu, Y.~Zhang, J.~Liang, and J.~Feng, ``No fear of heterogeneity: Classifier calibration for federated learning with non-iid data,'' in \emph{Proc. Conf. Adv. Neural Inf. Process. Syst. (NeurIPS)}, vol.~34, Virtual Event, Dec. 2021, pp. 5972--5984.

\bibitem{zhang2022federated}
J.~Zhang, Z.~Li, B.~Li, J.~Xu, S.~Wu, S.~Ding, and C.~Wu, ``Federated learning with label distribution skew via logits calibration,'' in \emph{Proc. Int. Conf. Mach. Learn. (ICML)}.\hskip 1em plus 0.5em minus 0.4em\relax PMLR, Jul. 2022, pp. 26\,311--26\,329.

\bibitem{li2022federated}
Z.~Li, J.~Shao, Y.~Mao, J.~H. Wang, and J.~Zhang, ``Federated learning with gan-based data synthesis for non-iid clients,'' in \emph{FL Workshop in Proc. Int. Joint Conf Artif. Intell ({IJCAI})}, vol. 13448, Vienna, Austria, Jul. 2022, pp. 17--32.

\bibitem{tang2022virtual}
Z.~Tang, Y.~Zhang, S.~Shi, X.~He, B.~Han, and X.~Chu, ``Virtual homogeneity learning: Defending against data heterogeneity in federated learning,'' in \emph{Proc. Int. Conf. Mach. Learn. (ICML)}, vol. 162, Baltimore, Maryland, USA, Jul. 2022, pp. 21\,111--21\,132.

\bibitem{sun2022stochastic}
Y.~Sun, J.~Shao, S.~Li, Y.~Mao, and J.~Zhang, ``Stochastic coded federated learning with convergence and privacy guarantees,'' in \emph{IEEE Int. Symp. Inf. Theory (ISIT)}, Espoo, Finland, Aug. 2022, pp. 2028--2033.

\bibitem{shao2022DReS}
J.~Shao, Y.~Sun, S.~Li, and J.~Zhang, ``{DReS-FL}: Dropout-resilient secure federated learning for non-iid clients via secret data sharing,'' in \emph{Proc. Conf. Adv. Neural Inf. Process. Syst. (NeurIPS)}, LA, CA, USA, May 2022.

\bibitem{tan2022fedproto}
Y.~Tan, G.~Long, L.~Liu, T.~Zhou, Q.~Lu, J.~Jiang, and C.~Zhang, ``{Fedproto}: Federated prototype learning across heterogeneous clients,'' in \emph{Proc. AAAI Conf. Artif. Intell. (AAAI)}, vol.~1, Virtual Event, Feb. 2022, p.~3.

\bibitem{snell2017prototypical}
J.~Snell, K.~Swersky, and R.~Zemel, ``Prototypical networks for few-shot learning,'' in \emph{Proc. Conf. Adv. Neural Inf. Process. Syst. (NeurIPS)}, vol.~30, Long Beach, CA, USA, Dec. 2017.

\bibitem{itahara2021distillation}
S.~Itahara, T.~Nishio, Y.~Koda, M.~Morikura, and K.~Yamamoto, ``Distillation-based semi-supervised federated learning for communication-efficient collaborative training with non-iid private data,'' \emph{IEEE Trans. Mobile Comput.}, vol.~22, no.~1, pp. 191--205, Jul. 2023.

\bibitem{zhang2021federated}
L.~Zhang, Y.~Luo, Y.~Bai, B.~Du, and L.-Y. Duan, ``Federated learning for non-iid data via unified feature learning and optimization objective alignment,'' in \emph{Proc. IEEE/CVF Int. Conf. Comput. Vision (ICCV)}, Montreal, QC, Canada, Oct. 2021, pp. 4420--4428.

\bibitem{reddi2021adaptive}
S.~J. Reddi, Z.~Charles, M.~Zaheer, Z.~Garrett, K.~Rush, J.~Kone{\v{c}}n{\'y}, S.~Kumar, and H.~B. McMahan, ``Adaptive federated optimization,'' in \emph{Proc. Int. Conf. Learn. Repr. (ICLR)}, Virtual Event, May 2021.

\bibitem{Wang2020Federated}
H.~Wang, M.~Yurochkin, Y.~Sun, D.~Papailiopoulos, and Y.~Khazaeni, ``Federated learning with matched averaging,'' in \emph{Proc. Int. Conf. Learn. Repr. (ICLR)}, Addis Ababa, Ethiopia, Apr. 2020.

\bibitem{wang2022performance}
S.~Wang, M.~Chen, C.~G. Brinton, C.~Yin, W.~Saad, and S.~Cui, ``Performance optimization for variable bitwidth federated learning in wireless networks,'' \emph{IEEE Trans. Wirel. Commun.}, Mar. 2023.

\bibitem{zhang2023reliable}
W.~Zhang, H.~Liang, Y.~Xu, and C.~Zhang, ``Reliable and privacy-preserving federated learning with anomalous users,'' \emph{ZTE Communications}, vol.~21, no.~1, pp. 15--24, Feb 2023.

\bibitem{Yang2023DetFed}
D.~Yang, W.~Zhang, Q.~Ye, C.~Zhang, N.~Zhang, C.~Huang, H.~Zhang, and X.~Shen, ``{DetFed}: Dynamic resource scheduling for deterministic federated learning over time-sensitive networks,'' \emph{IEEE Trans. Mobile Comput.}, pp. 1--17, Aug. 2023.

\bibitem{he2021fedcv}
C.~He, A.~D. Shah, Z.~Tang, D.~F.~N. Sivashunmugam, K.~Bhogaraju, M.~Shimpi, L.~Shen, X.~Chu, M.~Soltanolkotabi, and S.~Avestimehr, ``{FedCV}: a federated learning framework for diverse computer vision tasks,'' [Online]. Available: \url{https://arxiv.org/pdf/2111.11066.pdf}.

\bibitem{xiao2017fashion}
H.~Xiao, K.~Rasul, and R.~Vollgraf, ``Fashion-mnist: a novel image dataset for benchmarking machine learning algorithms,'' [Online]. Available: \url{https://arxiv.org/pdf/1708.07747.pdf}.

\bibitem{li2021fedbn}
X.~Li, M.~Jiang, X.~Zhang, M.~Kamp, and Q.~Dou, ``{FedBN}: Federated learning on non-iid features via local batch normalization,'' in \emph{Proc. Int. Conf. Learn. Repr. (ICLR)}, Virtual Event, May 2021.

\bibitem{van2008visualizing}
L.~Van~der Maaten and G.~Hinton, ``Visualizing data using t-sne,'' \emph{J. Mach. Learn. Res.}, vol.~9, no.~11, 2008.

\bibitem{movshovitz2017no}
Y.~Movshovitz-Attias, A.~Toshev, T.~K. Leung, S.~Ioffe, and S.~Singh, ``No fuss distance metric learning using proxies,'' in \emph{Proc. IEEE/CVF Int. Conf. Comput. Vision (ICCV)}, Venice, Italy, Oct. 2017, pp. 360--368.

\bibitem{wang2021addressing}
L.~Wang, S.~Xu, X.~Wang, and Q.~Zhu, ``Addressing class imbalance in federated learning,'' in \emph{Proc. AAAI Conf. Artif. Intell. (AAAI)}, vol.~35, Virtual Event, Feb. 2021, pp. 10\,165--10\,173.

\bibitem{wen2016discriminative}
Y.~Wen, K.~Zhang, Z.~Li, and Y.~Qiao, ``A discriminative feature learning approach for deep face recognition,'' in \emph{Proc. Eur. Conf. Comp. Vision (ECCV)}, vol. 9911, Amsterdam, Netherlands, Oct. 2016, pp. 499--515.

\bibitem{molchanov2016pruning}
P.~Molchanov, S.~Tyree, T.~Karras, T.~Aila, and J.~Kautz, ``Pruning convolutional neural networks for resource efficient inference,'' in \emph{Proc. Int. Conf. Learn. Repr. (ICLR)}, Toulon, France, Apr. 2016.

\bibitem{yurochkin2019bayesian}
M.~Yurochkin, M.~Agarwal, S.~Ghosh, K.~Greenewald, N.~Hoang, and Y.~Khazaeni, ``Bayesian nonparametric federated learning of neural networks,'' in \emph{Proc. Int. Conf. Mach. Learn. (ICML)}, vol.~97, Long Beach, California, USA, Jun. 2019, pp. 7252--7261.

\bibitem{cohen2017emnist}
G.~Cohen, S.~Afshar, J.~Tapson, and A.~Van~Schaik, ``Emnist: Extending mnist to handwritten letters,'' in \emph{Proc. Int. Jt. Conf. Neural Networks (IJCNN)}, Anchorage, AK, USA, May 2017, pp. 2921--2926.

\bibitem{krizhevsky2009learning}
A.~Krizhevsky, G.~Hinton \emph{et~al.}, ``Learning multiple layers of features from tiny images,'' [Online]. Available: \url{https://www.cs.toronto.edu/~kriz/learning-features-2009-TR.pdf}.

\bibitem{lecun1998gradient}
Y.~LeCun, L.~Bottou, Y.~Bengio, and P.~Haffner, ``Gradient-based learning applied to document recognition,'' \emph{Proc. IEEE}, vol.~86, no.~11, pp. 2278--2324, Mar. 1998.

\bibitem{netzer2011reading}
Y.~Netzer, T.~Wang, A.~Coates, A.~Bissacco, B.~Wu, and A.~Y. Ng, ``Reading digits in natural images with unsupervised feature learning,'' in \emph{Deep Learning Workshop of Neural Inf. Process. Syst. (NeurIPS)}, 2011.

\bibitem{hull1994database}
J.~J. Hull, ``A database for handwritten text recognition research,'' \emph{IEEE Trans. Pattern Anal. Mach. Intell.}, vol.~16, no.~5, pp. 550--554, May 1994.

\bibitem{ganin2015unsupervised}
Y.~Ganin and V.~Lempitsky, ``Unsupervised domain adaptation by backpropagation,'' in \emph{Proc. Int. Conf. Mach. Learn. (ICML)}, vol.~37, Lille, France, Jul. 2015, pp. 1180--1189.

\bibitem{he2016deep}
K.~He, X.~Zhang, S.~Ren, and J.~Sun, ``Deep residual learning for image recognition,'' in \emph{Proc. IEEE/CVF Conf. Comput. Vision Pattern Recognit. (CVPR)}, Las Vegas, NV, USA, Jun. 2016, pp. 770--778.

\bibitem{yu2019linear}
H.~Yu, R.~Jin, and S.~Yang, ``On the linear speedup analysis of communication efficient momentum sgd for distributed non-convex optimization,'' in \emph{Proc. Int. Conf. Mach. Learn. (ICML)}, vol.~97, Long Beach, California, USA, Jun. 2019, pp. 7184--7193.

\bibitem{lin2020dont}
T.~Lin, S.~U. Stich, K.~K. Patel, and M.~Jaggi, ``Don't use large mini-batches, use local {SGD},'' in \emph{ICLR - Proc. Int. Conf. Learn. Repr. (ICLR)}, Addis Ababa, Ethiopia, Apr. 2020.

\end{thebibliography}
\bibliographystyle{IEEEtran}

\begin{IEEEbiography}[{\includegraphics[width=1in,height=1.25in,clip,keepaspectratio]{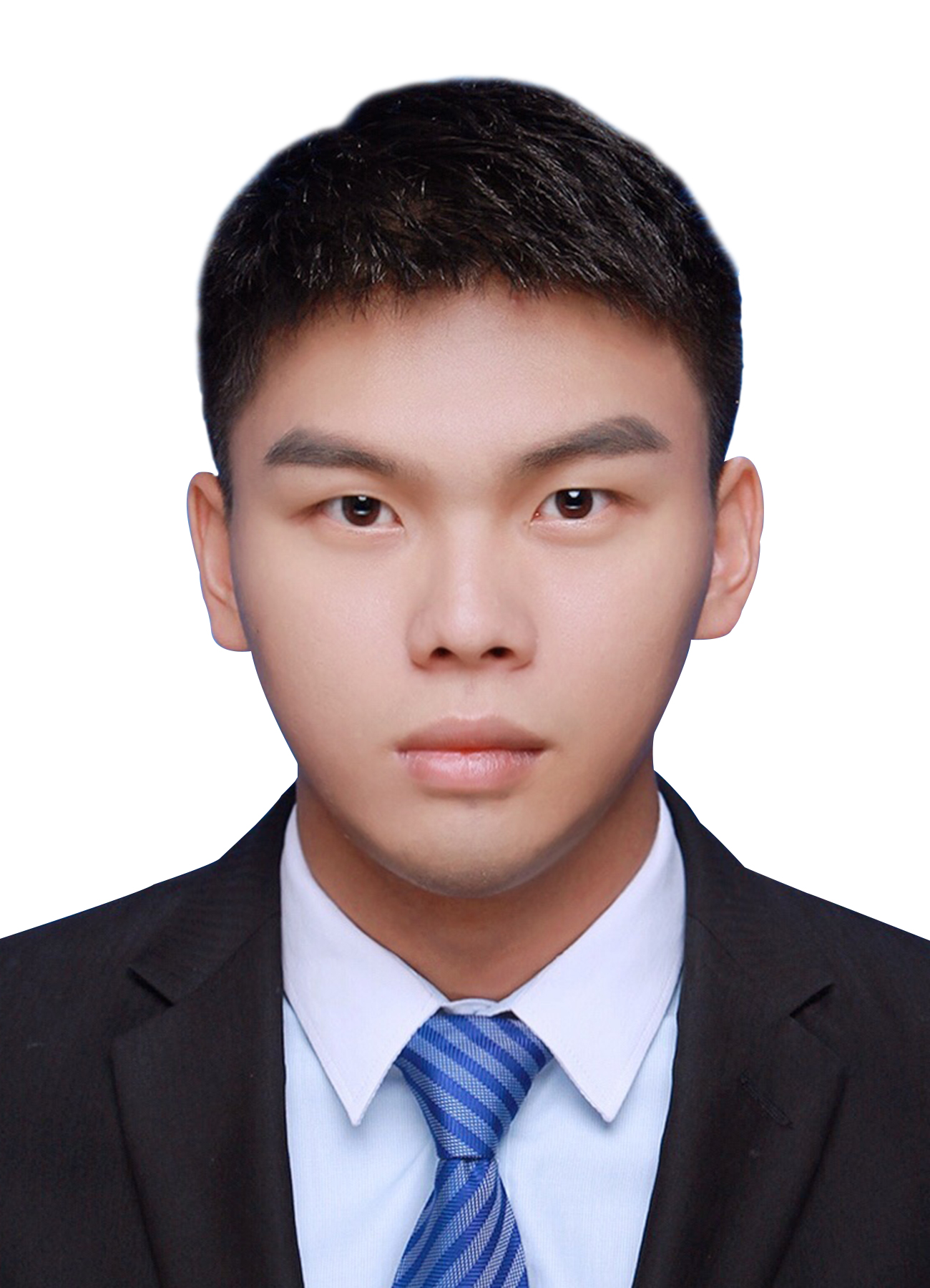}}]{Tailin Zhou}
(Graduate student member, IEEE) received his B.Eng. degree in Electrical Engineering and Automation Engineering from Sichuan University in 2018, and his Master's degree in Electrical Engineering from South China University of Technology in 2021.
He is pursuing a Ph.D. degree at the Hong Kong University of Science and Technology under the supervision of  Professor Jun Zhang and Professor Danny H.K. Tsang.
His research interests include federated/collaborative learning and its application in the Internet of Things and smart grids.
\end{IEEEbiography}
\vskip -2\baselineskip plus -1fil

\begin{IEEEbiography}
[{\includegraphics[width=1in,height=1.25in,clip,keepaspectratio]{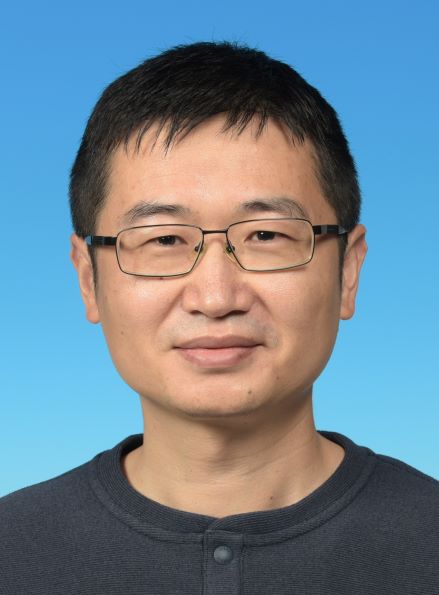}}]{Jun Zhang}
(Fellow, IEEE) received the B.Eng. degree in electronic engineering from the University of Science and Technology of China in 2004, the M.Phil. degree in information engineering from The Chinese University of Hong Kong in 2006, and the Ph.D. degree in Electrical and Computer Engineering from the University of Texas at Austin in 2009. 
He is an Associate Professor in the Department of Electronic and Computer Engineering at the Hong Kong University of Science and Technology. His research interests include privacy-preserving collaborative learning and cooperative AI. 
He is an IEEE Fellow.
\end{IEEEbiography}  
\vskip -2\baselineskip plus -1fil

\begin{IEEEbiography}
[{\includegraphics[width=1in,height=1.25in,clip,keepaspectratio]{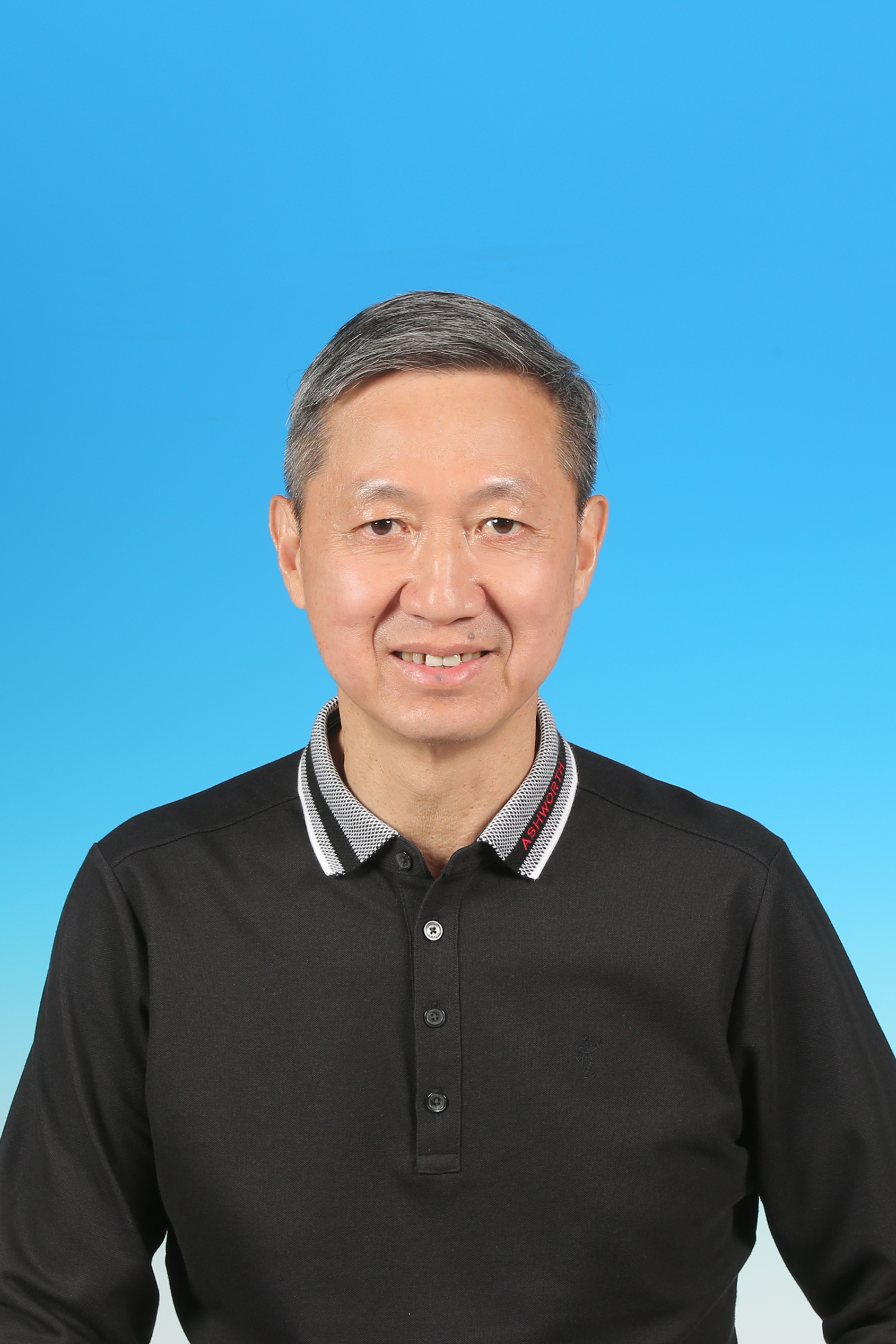}}]{Danny H.K. Tsang}
(Life Fellow, IEEE)  received the Ph.D. degree in electrical engineering from the Moore School of Electrical Engineering, University of Pennsylvania, Philadelphia, PA, USA, in 1989. After graduation, he joined the Department of Computer Science, Dalhousie University, Halifax, NS, Canada. He later joined the Department of Electronic and Computer Engineering, The Hong Kong University of Science and Technology (HKUST), Hong Kong, in 1992, where he is currently a Professor. He has also been serving as the Thrust Head of the Internet of Things Thrust, HKUST (Guangzhou), Guangzhou, China, since 2020. During his leave from HKUST from 2000 to 2001, he assumed the role of Principal Architect with Sycamore Networks, Chelmsford, MA, USA. His current research interests include cloud computing, edge computing, NOMA networks, and smart grids. He was a Guest Editor of the IEEE JOURNAL ON SELECTED AREAS IN COMMUNICATIONS’ special issue on Advances in P2P Streaming Systems, an Associate Editor of Journal of Optical Networking published by the Optical Society of America, and a Guest Editor of IEEE SYSTEMS JOURNAL. He currently serves as a member of the Special Editorial Cases Team of IEEE Communications Magazine. He was responsible for the network architecture design of Ethernet MAN/WAN over SONET/DWDM networks. He invented the 64B/65B encoding (U.S. Patent No.: U.S. 6 952 405 B2) and contributed it to the proposal for Transparent GFP in the T1X1.5 standard that was advanced to become the ITU G.GFP standard. The coding scheme has now been adopted by International Telecommunication Union (ITU)’s Generic Framing Procedure Recommendation GFP-T (ITUT G.7041/Y.1303) and Interfaces of the Optical Transport Network (ITU-T G.709). He was nominated to become an HKIE Fellow in 2013.
\end{IEEEbiography}

\clearpage
\renewcommand{\appendixname}{Appendix}
\appendix
\subsection{Proof of Non-convex Convergence rate of FedFA}\label{proof:FedFA_convergence}
\begin{proof}
According to the Lipschitz-smooth assumption, let  $\boldsymbol{g}_i^{(t)}=\sum_{k=0}^{\tau_K-1}  \mathbf{g}_i(\mathbf{w}_i^{(t, k) };\xi_i^{(t,\tau_k)})$  and $q_i=n_i/\sum_{i=1}^N n_i$, then we have:
\begin{equation}
  \begin{aligned}
  \mathbb{E} [\mathcal{L} (\mathbf{w}^{(t+\frac{1}{2},0)})]-\mathcal{L} (\mathbf{w}^{(t, 0)})  \leq  \frac{\tau_K^2 \eta^2 L_1}{2} \underbrace{\mathbb{E} [ \|\sum_{i=1}^N q_i \boldsymbol{g}_i^{(t)}\|^2]}_{T_2} \\
 \ - \tau_K \eta \underbrace{\mathbb{E} [ \langle\nabla \mathcal{L} (\mathbf{w}^{(t, 0)}), \sum_{i=1}^N q_i \boldsymbol{g}_i^{(t)}\rangle]}_{T_1} 
\end{aligned}  
\label{proofstep:T1+T2}
\end{equation}
 where $\tau_K$ is the total number of update iterations and the expectation is taken over mini-batches samples $\xi_i^{(t,\tau_k)}, \tau_k \in \{0,1,\dots,\tau_K-1\}$.

 For the first term $T_1$, let $\boldsymbol{d}_i^{(t)}=\sum_{k=0}^{\tau_K-1} \nabla \mathcal{L}_i(\mathbf{w}_i^{(t, k)})$ and we have:
\begin{equation}
    \begin{aligned}
T_1  = & 
\frac{1}{2} \|\nabla \mathcal{L} (\mathbf{w}^{(t,0)})\|^2+\frac{1}{2} \mathbb{E} [ \|\sum_{i=1}^N q_i \boldsymbol{d}_i^{(t)}\|^2] \\
 & -\frac{1}{2} \mathbb{E} [ \|\nabla \mathcal{L} (\mathbf{w}^{(t, 0)})-\sum_{i=1}^N q_i \boldsymbol{d}_i^{(t)} \|^2 ] 
\end{aligned}
\label{proofstep:T1}
\end{equation}
where the equation follows $2\langle a, b\rangle=\|a\|^2+\|b\|^2-\|a-b\|^2$.

 For the first term $T_2$, we have:
\begin{equation}
\begin{aligned}
T_2 & =\mathbb{E} [ \|\sum_{i=1}^N q_i (\boldsymbol{g}_i^{(t)}-\boldsymbol{d}_i^{(t)})+\sum_{i=1}^N q_i \boldsymbol{d}_i^{(t)}\|^2] \\
& \leq 2 \mathbb{E} [ \|\sum_{i=1}^N q_i (\boldsymbol{g}_i^{(t)}-\boldsymbol{d}_i^{(t)})\|^2]+2 \mathbb{E} [ \|\sum_{i=1}^N q_i \boldsymbol{d}_i^{(t)}\|^2] \\
& =2 \sum_{i=1}^N q_i^2 \mathbb{E} [ \|\boldsymbol{g}_i^{(t)}-\boldsymbol{d}_i^{(t)}\|^2]+2 \mathbb{E} [ \|\sum_{i=1}^N q_i \boldsymbol{d}_i^{(t)}\|^2]
\\
& \leq  2 \frac{\sigma^2}{\tau_K} +2 \mathbb{E} [ \|\sum_{i=1}^N q_i \boldsymbol{d}_i^{(t)}\|^2]
\end{aligned}
\label{proofstep:T2}
\end{equation}
where the first inequality follows $\|a+b\|^2 \leq 2\|a\|^2+2\|b\|^2$; the second equality uses the special property of $\boldsymbol{g}_i^{(t)}, \boldsymbol{d}_i^{(t)}$, that is, $\mathbb{E} \langle\boldsymbol{g}_i^{(t)}-\boldsymbol{d}_i^{(t)}, \boldsymbol{g}_j^{(t)}-\boldsymbol{d}_j^{(t)}\rangle=0, \forall i \neq j$;  the second inequality follows Assumption \ref{ass-2}.

 Combining (\ref{proofstep:T1}) and (\ref{proofstep:T2}) back into (\ref{proofstep:T1+T2}), when $\tau_K \eta L_1 \leq 1/2$, we have:
\begin{equation*}
  \begin{aligned}
  \mathbb{E} &  [\mathcal{L} (\mathbf{w}^{(t+\frac{1}{2},0)})]-\mathcal{L} (\mathbf{w}^{(t, 0)})  \\
   =   &  \frac{-\tau_K \eta}{2} \|\nabla \mathcal{L} (\mathbf{w}^{(t,0)})\|^2   \\
  &  -\frac{\tau_K \eta}{2}(1-2\tau_K \eta L_1) \mathbb{E} [ \|\sum_{i=1}^N q_i \boldsymbol{d}_i^{(t)}\|^2]  \\
 &  +  \tau_K L_1 \eta^2 \sigma^2  + \frac{ \tau_K \eta}{2} \mathbb{E} [ \| \nabla\mathcal{L} (\mathbf{w}^{(t, 0)})-\sum_{i=1}^N q_i \boldsymbol{d}_i^{(t)} \|^2 ]  \\
 \end{aligned}
\end{equation*}
\begin{equation}
  \begin{aligned}
  \leq &  \frac{-\tau_K \eta}{2} \|\nabla \mathcal{L} (\mathbf{w}^{(t,0)})\|^2 
  +  \tau_K L_1 \eta^2 \sigma^2  \\ 
&  + \frac{ \tau_K \eta}{2} \mathbb{E} [ \| \nabla\mathcal{L} (\mathbf{w}^{(t, 0)})-\sum_{i=1}^N q_i \boldsymbol{d}_i^{(t)} \|^2 ] \\
  \leq  & \frac{-\tau_K \eta}{2} \|\nabla \mathcal{L} (\mathbf{w}^{(t,0)})\|^2 \\ 
 & +  \tau_K L_1 \eta^2 \sigma^2  + \frac{ \tau_K \eta}{2} \sum_i^N q_i \mathbb{E} [ \| \nabla\mathcal{L}_i (\mathbf{w}^{(t, 0)})-\boldsymbol{d}_i^{(t)} \|^2 ] 
\end{aligned}
\label{proofstep:combine_T1_T2}
\end{equation}
where the last inequality uses $\mathcal{L} = \sum_i^N q_i\mathcal{L}_i$ and Jensen’s Inequality.

Next,  we   derive the term related to $\boldsymbol{d}_i^{(t)}$ in (\ref{proofstep:combine_T1_T2}) as:
\begin{equation}
  \begin{aligned}
\mathbb{E}&  [ \| \nabla\mathcal{L}_i (\mathbf{w}^{(t, 0)})-\boldsymbol{d}_i^{(t)} \|^2 ]  \\
  & = 
\mathbb{E} [ \|\sum_{k=0}^{\tau_K-1}  (\nabla\mathcal{L}_i (\mathbf{w}^{(t, 0)})- \nabla\mathcal{L}_i (\mathbf{w}_i^{(t, k)}) )\|^2 ] \\
& \leq  \sum_{k=0}^{\tau_K-1}  \mathbb{E} [ \|  (\nabla\mathcal{L}_i (\mathbf{w}^{(t, 0)})- \nabla\mathcal{L}_i (\mathbf{w}_i^{(t, k)}) )\|^2 ] \\
& \leq  L_1^2 \sum_{k=0}^{\tau_K-1}  \mathbb{E} [ \| \mathbf{w}^{(t, 0)} -  \mathbf{w}_i^{(t, k)}  \|^2 ] \\
\end{aligned}
\label{proofstep:distance_gradient}
\end{equation}
where the first inequality uses  Jensen’s Inequality and the second inequality follows Assumption \ref{ass-1}.
Further,  we   derive the right-hand-side term related to the distance between the global model and the client model in (\ref{proofstep:distance_gradient}) as:
\begin{equation}
  \begin{aligned}
  \sum_{k=0}^{\tau_K-1} & \mathbb{E} [ \| \mathbf{w}^{(t, 0)} -  \mathbf{w}_i^{(t, k)}  \|^2 ]    \\
 & = \eta^2 \sum_{k=0}^{\tau_K-1} \mathbb{E} [ \| \sum_{s=0}^{k-1}  \mathbf{g}_i(\mathbf{w}_i^{(t, s) };\xi_i^{(t, s)}) \|^2 ]  \\
 & \leq \eta^2 \sum_{k=0}^{\tau_K-1} \sum_{s=0}^{k-1} 
 \mathbb{E} [ \|   \mathbf{g}_i(\mathbf{w}_i^{(t, s) };\xi_i^{(t, s)}) \|^2 ]  \\
&  \leq   \eta^2 \sum_{k=0}^{\tau_K-1} \sum_{s=0}^{k-1} G^2 = \eta^2  (\tau_K-1) G^2/2 \\
    & \leq \eta^2\tau_K G^2/2
\end{aligned}
\label{proofstep:distance_model1}
\end{equation}
where the first inequality uses Jensen’s Inequality, the  second inequality follows Assumption \ref{ass-3} and the last inequality holds because of $G^2 \geq 0$.

Taking both (\ref{proofstep:distance_model1}) and (\ref{proofstep:distance_gradient})  into (\ref{proofstep:combine_T1_T2}), we have:
\begin{equation}
  \begin{aligned}
  \mathbb{E}& [\mathcal{L} (\mathbf{w}^{(t+\frac{1}{2},0)})]-\mathcal{L} (\mathbf{w}^{(t, 0)})  \\
  \leq &    \frac{-\tau_K \eta}{2} \|\nabla \mathcal{L} (\mathbf{w}^{(t,0)})\|^2 +  \tau_K L_1 \eta^2 \sigma^2  
   +  \frac{G^2 L_1^2 \tau_K^2 \eta^3}{4}. 
\end{aligned}
\label{proofstep:one_round_deviation_full}
\end{equation}
Herein, let the server would finish the aggregation of models and feature anchors at the time $(t+1,0)$,    we take $ (t+\frac{1}{2},0) $ to denote the time between   model  aggregation and anchor aggregation. That is, $\mathbf{w}^{(t+\frac{1}{2},0)} = \mathbf{w}^{(t+1,0)}$ = $\sum_{i=1}^N q_i\mathbf{w}_i^{(t,\tau_K)}$, $\bm{\theta}^{(t+\frac{1}{2},0)} = \bm{\theta}^{(t+1,0)}$ and $\mathbf{a}^{(t+\frac{1}{2} )}_c  = \mathbf{a}^{(t )}_c$, and we have:
\begin{equation}
  \begin{aligned}
 \mathcal{L} & (\mathbf{w}^{(t+1,0)})  \\
= & \mathcal{L} (\mathbf{w}^{(t+\frac{1}{2},0)}) +\mathcal{L} (\mathbf{w}^{(t+1,0)}) -\mathcal{L} (\mathbf{w}^{(t+\frac{1}{2},0)}) \\
= & \mathcal{L} (\mathbf{w}^{(t+\frac{1}{2},0)}) +\mu\sum_{i=1}^N q_i ( \|f_i (\bm{\theta}^{(t+1,0)})-\mathbf{a}^{(t+1)}_c\|^2 \\
&  -  \|f_i (\bm{\theta}^{(t+1,0)})-\mathbf{a}^{(t)}_c\|^2) \\
 \leq &   \mathcal{L} (\mathbf{w}^{(t+\frac{1}{2},0)})  +\mu\sum_{i=1}^N q_i \|\mathbf{a}^{(t+1)}_c-\mathbf{a}^{(t)}_c\|^2 \\
 = &\mathcal{L} (\mathbf{w}^{(t+\frac{1}{2},0)}) + \mu 
  \|\sum_{i=1}^N q_i (\mathbf{a}_{c,i}^{(t+1)}-\mathbf{a}_{c,i}^{(t)})\|^2 \\
   = & \mathcal{L} (\mathbf{w}^{(t+\frac{1}{2},0)}) \\
  & +\mu  \| \sum_{i=1}^N q_i \frac{1}{ |\mathcal{D}_i|} \sum_{k=1}^{ |\mathcal{D}_i|} (f_i (\bm{\theta}^{(t+1,0)} ; \xi_{i, k})-f_i (\bm{\theta}^{(t,0)}  ; \xi_{i, k}) \|^2. \\
   \leq &  \mathcal{L} (\mathbf{w}^{(t+\frac{1}{2},0)}) \\
& +\mu \sum_{i=1}^N \frac{q_i}{ |\mathcal{D}_i|} \sum_{k=1}^{ |\mathcal{D}_i|} \|f_i (\bm{\theta}^{(t+1,0)} ; \xi_{i, k})-f_i (\bm{\theta}^{(t,0)}  ; \xi_{i, k})\|^2 \\
 \leq &  \mathcal{L} (\mathbf{w}^{(t+\frac{1}{2},0)}) +\mu L_2^2 \sum_{i=1}^N q_i \|\bm{\theta}^{(t+1,0)}-\bm{\theta}^{(t,0)} \|^2 \\
  \leq & \mathcal{L} (\mathbf{w}^{(t+\frac{1}{2},0)}) +\mu L_2^2 \sum_{i=1}^N q_i \|\mathbf{w}^{(t+1,0)}-\mathbf{w}^{(t,0)}\|^2\\
  = &\mathcal{L} (\mathbf{w}^{(t+\frac{1}{2},0)}) +\mu L_2^2 \eta^2   \| \sum_{i=1}^N q_i \sum_{k=0}^{\tau_K-1} \mathbf{g}_i(\mathbf{w}_i^{(t, k) }) 
 \|^2 \\
  \leq &   \mathcal{L} (\mathbf{w}^{(t+\frac{1}{2},0)}) +\mu  L_2^2 \eta^2 \sum_{i=1}^N q_i \sum_{s=0}^{\tau_K-1}   \|  \mathbf{g}_i(\mathbf{w}_i^{(t, s) }) \|^2 \\
= &   \mathcal{L} (\mathbf{w}^{(t+\frac{1}{2},0)}) +\mu  L_2^2 \eta^2  \tau_K G^2
\end{aligned}
\label{proofstep:anchor_agg}
\end{equation}
where the first inequality follows $\|a-b\| -\|a-c\|\leq\|b-c\|$,   the second and last inequalities follow Jensen’s Inequality, the third inequality follows Assumption \ref{ass-4}, the fourth inequality is due to $\bm{\theta}$ is a subset of $\mathbf{w}$.

Take expectations of  on both sides and Jensen’s Inequality, then we have:
\begin{equation}
\begin{aligned}
\mathbb{E} [\mathcal{L} (\mathbf{w}^{(t+1,0)})] & \leq \mathbb{E}  [\mathcal{L} (\mathbf{w}^{(t+\frac{1}{2},0)})] + \mu  L_2^2 \eta^2  \tau_K G^2.
\end{aligned}
\label{proofstep:agg_1}
\end{equation}

Thus, we have the one-round loss deviation by adding (\ref{proofstep:agg_1}) into (\ref{proofstep:one_round_deviation_full}) as:
\begin{equation}
  \begin{aligned}
    \mathbb{E}    [\mathcal{L} (\mathbf{w}^{(t+1,0)})]-\mathcal{L} (\mathbf{w}^{(t, 0)})    \leq       \frac{-\tau_K \eta}{2} \|\nabla \mathcal{L} (\mathbf{w}^{(t,0)})\|^2  \\ 
    +     \tau_K L_1 \eta^2 \sigma^2  
      +  \frac{G^2 L_1^2 \tau_K^2 \eta^3}{4}   + \mu  L_2^2 \eta^2  \tau_K G^2.
\end{aligned}
\label{proofstep:one_round_deviation_all}
\end{equation}

Taking the expectation of both sides in (\ref{proofstep:one_round_deviation_all}) and the average across all rounds, we get:
\begin{equation}
  \begin{aligned}
   \frac{1}{T}\sum_{t=0}^{T-1}  \frac{  \tau_K \eta}{2} \|\nabla \mathcal{L} (\mathbf{w}^{(t,0)})\|^2  \leq        \frac{1}{T}\mathcal{L} (\mathbf{w}^{(0,0)})  +  \tau_K   L_1 \eta^2 \sigma^2    \\
   -    \frac{1}{T}\mathbb{E}  [\mathcal{L} (\mathbf{w}^{(T,0)})]     
   +    \frac{ G^2 L_1^2 \tau_K^2 \eta^3}{4}   +    \mu    L_2^2 \eta^2  \tau_K G^2 
\end{aligned}
\label{proofstep:all_round_deviation}
\end{equation}

Denote $ \Delta = \mathcal{L} (\mathbf{w}^{(0,0)})     -    \mathbb{E}  [\mathcal{L} (\mathbf{w}^{(T,0)})]$, given any $\epsilon >0$ and let  
\begin{equation}
  \begin{aligned}
   \frac{1}{T}\sum_{t=0}^{T-1}  \|\nabla \mathcal{L} (\mathbf{w}^{(t,0)})\|^2   \leq \epsilon,
\end{aligned}
\label{proofstep:epsilon}
\end{equation}
\begin{equation}
  \begin{aligned}
  T \geq \frac{8\Delta  }{4\tau_K \eta  \epsilon -  \tau_K \eta  (4  \eta L_1   \sigma^2    +    \tau_K     \eta^2 L_1^2   G^2    +    4\mu  \eta    L_2^2G^2 )       }
\end{aligned}
\label{proofstep:T}
\end{equation}
where the equality holds, i.e., the denominator of the right-hand side should be larger than 0 when $\eta < \min \{\frac{4 \epsilon }{4 L_1 \sigma^2 +  \tau_K  G^2 L_1^2  +   4\mu  L_2^2  G^2}, \frac{1}{2\tau_KL_1}\}$ and $\mu < \frac{ \epsilon -   L_1 (\sigma^2 + L_1 G^2)}{ L_2^2G^2}$.
    
\end{proof}

\subsection{Proof of Lemma \ref{gradient of classifier}} \label{proof:Def 1}
\begin{proof}
We derive the gradient of cross-entropy loss as:
 \begin{equation}
 \begin{aligned}
\Delta_{\bm{\phi}_c} = &  \Delta \bm{\phi}_{a,c} -  \Delta \bm{\phi}_{b,c}\\
 \approx  &  ( \frac{\eta n_{a,c}}{n_a} (1-\overline{p_{a, c}^{(c)}}) \overline{\mathbf{h}}_{a,c} - \frac{\eta n_{b,c}}{n_b} (1-\overline{p_{b, c}^{(c)}}) \overline{\mathbf{h}}_{b,c})\\
 & - (   \frac{\eta}{n_a}  \sum_{\bar{c}_a\neq c} n_{a,\bar{c}_a}\overline{p_{a,\bar{c}}^{(\bar{c}_a)}} \overline{\mathbf{h}}_{a,\bar{c}_a} -\frac{\eta}{n_b} \sum_{\bar{c}_b\neq c} n_{b,\bar{c}_b} \overline{p_{b,\bar{c}}^{(\bar{c}_b)}} \overline{\mathbf{h}}_{b,\bar{c}_b} )\\
 \xlongequal[n_a=n]{n_b=n} & \frac{\eta}{n}   [ \underbrace{ ( n_{a,c} (1-\overline{p_{a, c}^{(c)}}) \overline{\mathbf{h}}_{a,c} - n_{b,c} (1-\overline{p_{b, c}^{(c)}}) \overline{\mathbf{h}}_{b,c})}_{\text {mean deviation by \textit{positive features}}}\\
 &- \underbrace{  (   \sum_{\bar{c}_a\neq c} n_{a,\bar{c}_a} \overline{p_{a,\bar{c}}^{(\bar{c}_a)}} \overline{\mathbf{h}}_{a,\bar{c}_a} - \sum_{\bar{c}_b\neq c}   n_{b,\bar{c}_b} \overline{p_{b,\bar{c}}^{(\bar{c}_b)}} \overline{\mathbf{h}}_{b,\bar{c}_b} )}_{\text {deviation by \textit{mean negative features}}}]
  \end{aligned}
  \label{Proof:Def 1 step1}
\end{equation}
where the term of $``\approx"$ holds from Property 1 in \cite{zhang2022federated} based on the statistic results of Figures 4 and 5 in Appendix of \cite{wang2021addressing} (i.e., we assume  $\overline{p_{\cdot,{c}}^{({c})}} \overline{\mathbf{h}}_{\cdot,{c}} = \frac{1}{n_{\cdot,c}}\sum_{y_{j}=c}^{n_{\cdot,c}} p_{\cdot, c}^{(j)}  \mathbf{h}_{\cdot,y_j}$ where $\overline{p_{\cdot,{c}}^{({c})}}=\frac{1}{n_{\cdot,c}}\sum_{y_{j}=c}^{n_{\cdot,c}} p_{\cdot, c}^{(j)}$ and $ \overline{\mathbf{h}}_{\cdot,{c}} = \frac{1}{n_{\cdot,c}}\sum_{y_{j}=c}^{n_{\cdot,c}} \mathbf{h}_{\cdot,y_j}$). 
Note that we do not assume that the extracted features of the same class across clients are similar (i.e., $ \overline{\mathbf{h}}_{a,{c}} \neq \overline{\mathbf{h}}_{b,{c}}$) like \cite{wang2021addressing}.
\label{proof of classifier deviation}
\end{proof}

\vspace{-30pt}
\subsection{Proof of Theorem \ref{theorem classifier}}\label{proof:classifier}
\begin{proof}
When  $c \in [C_a] \cap [C_b]$, $\mathbf{\overline{h}}_{a,c}=\mathbf{\overline{h}}_{b,c}$, $\overline{p_{a, c}^{(c)}} = \overline{p_{b, c}^{(c)}}$ and the deviation by mean positive features $\Delta_{\bm{\phi}_c}^{(+)}=0$, and we have:
\begin{equation}
\begin{aligned}
 \|\Delta_{\bm{\phi}_c}\|^2 &=
 \frac{\eta^2}{n^2} \| \Delta_{\bm{\phi}_c}^{(-)} \|^2 \\
 =& \frac{\eta^2}{n^2} \|   \sum_{\bar{c}_a\neq c} n_{a,\bar{c}_a} \overline{p_{a,c}^{(\bar{c}_a)}} \overline{\mathbf{h}}_{a,\bar{c}_a}  
- \sum_{\bar{c}_b\neq c}   n_{b,\bar{c}_b} \overline{p_{b,c}^{(\bar{c}_b)}} \overline{\mathbf{h}}_{b,\bar{c}_b}\|^2 \\
=  &  \frac{\eta^2}{n^2}\|   
\sum_{\hat{c}_a } n_{a,\hat{c}_a} \overline{p_{a,c}^{(\hat{c}_a)}} \overline{\mathbf{h}}_{a,\hat{c}_a}-
 \sum_{\hat{c}_b}   n_{b,\hat{c}_b} \overline{p_{b,c}^{(\hat{c}_b)}} \overline{\mathbf{h}}_{b,\hat{c}_b} \|^2 \geq 0 \\
\end{aligned}
\label{proof:thrm2 1}
\end{equation}
where $\hat{c}_a \in [C_a] \setminus \{[C_a] \cap [C_b]\} $ and $\hat{c}_b \in [C_b] \setminus \{[C_a] \cap [C_b]\}$ (i.e., $\hat{c}_a$ denotes the classes for which client $a$ has samples in its dataset but client $b$ does not).
When the equality  of (\ref{proof:thrm2 1}) holds,  $\overline{\mathbf{h}}_{a,\hat{c}_a } = \alpha \overline{\mathbf{h}}_{b,\hat{c}_b}$ where $\alpha =  n_{a,\hat{c}_a} \overline{p_{a,c}^{(\hat{c}_a)}}/n_{b,\hat{c}_b} \overline{p_{b,c}^{(\hat{c}_b)}} $, but it requires a perfect feature extractor for all clients, which is  not a piratical condition in federated learning according to \cite{zhao2018federated}.
Thus,  we have $ \|\Delta_{\bm{\phi}_c}\|^2 > 0$ when $c \in [C_a] \cap [C_b]$.

When $c \in \{[C]\setminus \{[C_a]\cup [C_b]\} \}$, $n_{a,c}=n_{b,c}=0$ and $\Delta_{\bm{\phi}_c}^{(+)}=0$, and then we have:
\begin{equation}
\begin{aligned}  
   \|\Delta_{\bm{\phi}_c}\|^2 & =  \frac{\eta^2}{n^2}\| \Delta_{\bm{\phi}_c}^{(-)} \|^2  \\
  & =  \frac{\eta^2}{n^2}\|   
\sum_{\hat{c}_a } n_{a,\hat{c}_a} \overline{p_{a,c}^{(\hat{c}_a)}} \overline{\mathbf{h}}_{a,\hat{c}_a} -
\sum_{\hat{c}_b}   n_{b,\hat{c}_b} \overline{p_{b,c}^{(\hat{c}_b)}} \overline{\mathbf{h}}_{b,\hat{c}_b} \|^2  >  0
\end{aligned}
\label{proof:thrm2 2}
\end{equation}
where the inequality holds   because $[C_a] \neq [C_b]$ and the second row of term $``="$ is the same as (\ref{proof:thrm2 1}).
Thus, there exists  classifier updates divergence between client $a$ and client $b$ when $c \in \{[C]\setminus \{[C_a]\cup [C_b]\} \}$.

When  $c \in [C_a] \setminus \{[C_a] \cap [C_b]\}$, $n_{b,c}=0$, and then we have:
\begin{equation}
\begin{aligned}
 \|\Delta_{\bm{\phi}_c}\|^2
=  &\frac{\eta^2}{n^2}
\|n_{a,c} (1-\overline{p_{a, c}^{(c)}}) \overline{\mathbf{h}}_{a,c} - ( \sum_{\bar{c}_a\neq c} n_{a,\bar{c}_a} \overline{p_{a,c}^{(\bar{c}_a)}} \overline{\mathbf{h}}_{a,\bar{c}_a} \\ 
&  -\sum_{\bar{c}_b\neq c}   n_{b,\bar{c}_b} \overline{p_{b,c}^{(\bar{c}_b)}} \overline{\mathbf{h}}_{b,\bar{c}_b} )\|^2 \\
\approx & \frac{\eta^2}{n^2}
\|n_{a,c}\overline{\mathbf{h}}_{a,c} - \overline{p_{a, c}^{(c)}} \overline{\mathbf{h}}_{a,c} \| >   0
\end{aligned}
\label{proof:thrm2 3}
\end{equation}
where the equality holds if and only if $\Delta_{\bm{\phi}_c}^{(-)}=n_{a,c} (1-\overline{p_{a, c}^{(c)}}) \overline{\mathbf{h}}_{a,c} $.
During training, it is quite challenging to maintain this situation between any two clients, so the term of $``\approx"$ is because $\overline{p_{a, c}^{(\bar{c})}}  \ll \overline{p_{a, c}^{(c)}} < 1$, so we can say that $\|\Delta_{\bm{\phi}_c}\|^2$ is much more likely to be positive.
The case of  $c \in [C_b] \setminus \{[C_a] \cap [C_b]\}$ is the same at that of $c \in [C_a] \setminus \{[C_a] \cap [C_b]\}$. We would not discuss this case herein.
Thus, there exists  classifier updates divergence between client $a$ and client $b$ when $c \in \{[C]\setminus \{[C_a]\cup [C_b]\} \}$ or $c \in [C_b] \setminus \{[C_a] \cap [C_b]\}$.

For feature distribution skew, client $a$ and client $b$ share the feature extractor at the start of each round, and the skewed input features of samples induce the client models to map inconsistent features (i.e., $\mathbf{\overline{h}}_{a,c} \neq \mathbf{\overline{h}}_{b,c}$ ), and we have:
\begin{equation}
\begin{aligned}
  \|\Delta_{\bm{\phi}_c}\|^2 
  &\geq \frac{\eta^2}{n^2}  | \| \Delta_{\bm{\phi}_c}^{(+)}\|^2  - \|\Delta_{\bm{\phi}_c}^{(-)} \|^2 | \geq 0.
\end{aligned}
\end{equation}
where  the equality holds if and only if $ \| \Delta_{\bm{\phi}_c}^{(+)}\|^2 = \|\Delta_{\bm{\phi}_c}^{(-)} \|^2$, but due to $\mathbf{\overline{h}}_{a,c} \neq \mathbf{\overline{h}}_{b,c}$, it is quite challenging to maintain this situation between any two clients during training. Thus, we can say that $\|\Delta_{\bm{\phi}_c}\|^2 > 0$, i.e.,  both label and feature skews diverge the classifier updates across clients.
\label{proof:theorem1}
\end{proof}

\vspace{-10pt}
\subsection{Proof of Lemma \ref{gradient of feature} and Theorem \ref{vicious cycle}} \label{proof:Def 2}
\begin{proof}
Similar to Proof of Lemma \ref{gradient of classifier}, we have:
 \begin{equation}
 \begin{aligned}
  \Delta_{\mathbf{h}_c}
  = &\Delta \mathbf{\overline{h}}_{a,c} -  \Delta \mathbf{\overline{h}}_{b,c} \\
 \xlongequal[n_{b,c}=n]{n_{a,c}=n} &  
 \eta [ \underbrace{ ( (1-\overline{p_{a, c}^{(c)}}) {\bm{\phi}}_{a,c} - (1-\overline{p_{b, c}^{(c)}}) {\bm{\phi}}_{b,c})}_{\text{mean deviation by \textit{ positive classifier proxy}}: \Delta_{{\mathbf{h}_c}}^{(+)}} \\
&  - \underbrace{  ( \sum_{\bar{c}} \overline{p_{a,\bar{c}}^{(c)}} {\bm{\phi}}_{a,\bar{c}}-\sum_{\bar{c}}   \overline{p_{b,\bar{c}}^{(c)}}  {\bm{\phi}}_{b,\bar{c}}  )}_{\text{mean deviation by \textit{ negative classifier proxies}}: \Delta_{{\mathbf{h}_c}}^{(-)}}]\\
  \xlongequal[\overline{p_{a,{c}}^{(c)}} = \overline{p_{b,{c}}^{(c)}}]{ \overline{p_{a,\bar{c}}^{(c)}} = \overline{p_{b,\bar{c}}^{(c)}} } & 
   \sum_{\hat{c}\in [C]}   \overline{p_{b,\hat{c}}^{(c)}} \Delta_{\bm{\phi}_{\hat{c}}}  -\eta \Delta_{\bm{\phi}_c}.
  \end{aligned}
\end{equation}
where we assume their possibility outputs for class $c$ are the same for the convenience of formulation, i.e., $\overline{p_{a,\bar{c}}^{(c)}} = \overline{p_{b,\bar{c}}^{(c)}}$. That is,   if client $a$ and client $b$ are training their models locally instead of federated training, they classify the $c$-th  classes they have in a local dataset with the same possibility output.
\end{proof} 

\vspace{-10pt}
 \subsection{Proof of Theorem \ref{Lipschitzness improvement}}
\label{proof:Lipschitzness improvement}

\begin{proof} 
 With the property of feature anchor loss (\ref{fa loss}), let the feature polymerization 
have  $\overline{\mathbf{h}}_{\cdot,c}=\mathbf{a}_c$. 
 Following (\ref{Proof:Def 1 step1}),  we represent the mean prediction output of FedAvg and FedFA as $\overline{{p}_{\cdot,{c}}^{({c})}}$ and $\overline{\hat{p}_{\cdot,{c}}^{({c})}}$, respectively.

With local classifier calibration,  comparing FedFA with FedAvg, for the $c$-th class prediction output, we have  $\overline{\hat{p}_{i,{c}}^{({c})}} > \overline{{p}_{i,{c}}^{({c})}}$ and $\overline{\hat{p}_{i,\bar{c}}^{({c})}} < \overline{{p}_{i,\bar{c}}^{({c})}}$.

We derive  the deviation of gradient norms of the global classifier between FedFA and FedAvg as:
\begin{equation}
 \begin{aligned}
& \|\Delta \bm{\hat{\phi}}_{c}\|^2 - \|\Delta \bm{{\phi}}_{c}\|^2 =   \|\sum_{i}^{N}  \Delta \bm{\hat{\phi}}_{i,c}\|^2  -  \|\sum_{i}^{N}  \Delta \bm{\phi}_{i,c}\|^2  \\
 =  &  \|\frac{\eta}{n}  [\underbrace{\sum_{i}^{N}  ( n_{i,c}(1-\overline{\hat{p}_{i, c}^{(c)}}))}_{\hat{A}_c}  {\mathbf{a}}_{c} -  \sum_{\bar{c}\neq c} \underbrace{\sum_{i}^{N}   n_{i,\bar{c}} \overline{\hat{p}_{i,{c}}^{(\bar{c})}}}_{\hat{B}_{ \bar{c}} } {\mathbf{a}}_{\bar{c}}  ]\|^2 \\
  & - \|\frac{\eta}{n}   [\underbrace{\sum_{i}^{N}  ( n_{i,c}(1-\overline{{p}_{i, c}^{(c)}}))}_{A_c}  {\mathbf{a}}_{c} -  \sum_{\bar{c}\neq c} \underbrace{\sum_{i}^{N}  n_{i,\bar{c}} \overline{{p}_{i,{c}}^{(\bar{c})}}}_{B_{ \bar{c}}} {\mathbf{a}}_{\bar{c}} ]\|^2\\
  \end{aligned}
\end{equation}
where $\Delta \bm{\hat{\phi}}_{c}$ and $\Delta \bm{{\phi}}_{c}$ is the update of the global classifier of FedFA and FedAvg, respectively;
${A}_c=\sum_{i}^{N}  ( n_{i,c}(1-\overline{{p}_{i, c}^{(c)}}))$,
$\hat{A}_c=\sum_{i}^{N} ( n_{i,c}(1-\overline{\hat{p}_{i, c}^{(c)}}))$,
$B_{ \bar{c}} = \sum_{i}^{N}  n_{i,\bar{c}} \overline{{p}_{i,{c}}^{(\bar{c})}}$, 
$\hat{B}_{ \bar{c}}  = \sum_{i}^{N}   n_{i,\bar{c}} \overline{\hat{p}_{i,{c}}^{(\bar{c})}}$.
Thus, we have:
\begin{equation}
 \begin{aligned}
\| & \nabla_{\bm{{\phi}}_{c}}\hat{\mathcal{L}}\|^2 - \|\nabla_{\bm{{\phi}}_{c}}{\mathcal{L}}\|^2    \\
 = &  \frac{\eta^2}{n^2}[\|\hat{A}_c  {\mathbf{a}}_{c} - \sum_{\bar{c}\neq c}\hat{B}_{\bar{c}} {\mathbf{a}}_{\bar{c}} \|^2 -\|{A}_c  {\mathbf{a}}_{c} - \sum_{\bar{c}\neq c} {B}_{\bar{c}} {\mathbf{a}}_{\bar{c}}\|^2  ] \\
= &  \frac{\eta^2}{n^2}[ (\hat{A}_c^2 - {A}_c^2)\|{\mathbf{a}}_{c}\|^2
 + \sum_{\bar{c}\neq c}(\hat{B}_{\bar{c}}^2 - {B}_{\bar{c}}^2)\|{\mathbf{a}}_{\bar{c}}\|^2 \\
&  + \sum_{\bar{c}_1\neq \bar{c}_2}2(\hat{B}_{c_1}\hat{B}_{{c}_2} - {B}_{c_1}{B}_{c_2}) {\mathbf{a}}_{\bar{c}_1}\cdot {\mathbf{a}}_{\bar{c}_2}  \\
& -  \sum_{\bar{c}\neq c}2(\hat{A}_c\hat{B}_{\bar{c}}-{A}_c{B}_{\bar{c}}){\mathbf{a}}_{{c}}\cdot {\mathbf{a}}_{\bar{c}}]\\
 =  &  \frac{\eta^2}{n^2}[ \underbrace{(\hat{A}_c^2 - {A}_c^2)}_{\textit{less than 0 due to }\overline{\hat{p}_{i,{c}}^{({c})}} > \overline{{p}_{i,{c}}^{({c})}}}\|{\mathbf{a}}_{c}\|^2  \\
  &+ \sum_{\bar{c}\neq c}\underbrace{(\hat{B}_{\bar{c}}^2 - {B}_{\bar{c}}^2)}_{\textit{less than 0 due to } \overline{\hat{p}_{i,\bar{c}}^{({c})}} < \overline{{p}_{i,\bar{c}}^{({c})}}}\|{\mathbf{a}}_{\bar{c}}\|^2  <   0
  \end{aligned}
\end{equation} 
where $\bar{c}_1\in \{[C]\setminus c\}$,  $\bar{c}_2\in \{[C]\setminus c\}$ and  $\bar{c}_1\neq \bar{c}_2$, and the last equality follows $\mathbf{a}_{{c}}\cdot {\mathbf{a}}_{\bar{c}}=0$.
It should be denoted that the assumption $\mathbf{a}_{{c}}\cdot {\mathbf{a}}_{\bar{c}}=0$ provides an initialization  method for feature anchors in FedFA, as discussed  in Figure \ref{anchor lambda and init}. 
\end{proof}

\clearpage
\renewcommand{\appendixname}{Supplementary Materials}
\appendix
The supplementary materials  include  as follows:
\begin{itemize}
    \item  Section \ref{Terminology}: the description of terminologies in this paper.
    \item Section  \ref{Privacy issues introduced by feature anchors}:   privacy issues introduced by feature anchors.
    \item  Section \ref{Additional Validation Experiment Results}: additional results of feature-inconsistency visualization and feature-similarity histograms  across all the baselines.
     \item  Section \ref{Additional Test Experiment Results}: additional experiments.
    \item Section \ref{Experiment Setup}: details of experiment setup.
\end{itemize}

 \subsection{Terminologies}\label{Terminology}
\textbf{\textit{Global model} vs. \textit{local model}.}
Let us first clarify the concepts of “global” vs. “local” models: 
in each communication round, local models denote the ones updated by the clients after local training, and the global model denotes the model obtained by aggregating all local models at the server. Moreover, client models denote the models being trained during local training.

\textbf{\textit{Vicious cycle} vs. \textit{virtuous cycle}.}
As shown in Figure \ref{toy_example} (a), the \textit{vicious cycle} represents the phenomenon that inconsistent features of local models diverge the classifier updates, such that the diverged classifiers of different clients induce feature extractors to map to more inconsistent features across clients.
As shown in Figure \ref{toy_example} (b), the \textit{virtuous cycle} represents the phenomenon that consistent features of client local models make the classifier updates similar, such that the updated classifiers make feature extractors of clients map to more consistent features across clients.

\textbf{\textit{Positive pair} vs. \textit{negative pair}.}
A \textit{positive pair} denotes a pair of \textbf{samples} with the same label (i.e., the samples belong to the same class).
A \textit{negative pair} denotes a pair of \textbf{samples} with different labels (i.e., the samples do not belong to the same class).

\textbf{\textit{Positive feature} vs. \textit{negative feature}.} For the $c$-th proxy $\bm{\phi}_c$,
the \textit{positive features} denote the \textbf{features} of the $c$-th class, and
the \textit{negative features} denote the \textbf{features} of  other classes except for the $c$-th class.

\textbf{\textit{Positive proxy} vs. \textit{negative proxy}.}
For the  feature of the $c$-th class,
the \textit{positive proxy} denotes the $c$-th \textbf{proxy} $\bm{\phi}_c$, and
the \textit{negative proxies} denote other \textbf{proxies} except for the $c$-th proxy $\bm{\phi}_c$.

 \subsection{Discussion of privacy issues introduced by feature anchors}\label{Privacy issues introduced by feature anchors}
According to current attack technology, we believe FedFA provides basic privacy protection with promising performance for federated learning.
\begin{itemize}
    \item  Firstly, feature anchors can be fixed/without updates during federated training (i.e., the client would not aggregate any information into feature anchors, which would not bring potential privacy leakage) because of the powerful representation of neural networks. That is, the fixed anchors in FedFA specified a feature space and a classifier between clients before training. The third row of Table \ref{ablation results} shows better results of experiments of FedFA without feature-anchor updates than the best baseline under label and feature skew even though the performance goes down about $2\%$ than FedFA with anchor updates. Therefore, there is a trade-off between privacy and generalization performance for FedFA. 
 Therefore, there is a trade-off between privacy and generalization performance for FedFA. 
    \item  Secondly,  FedFA only shares the feature centroid (statistic mean of features) of the last layer of the feature encoder and the feature centroid is changing rapidly under training, rather than the fully trained feature map from raw data, so the information leaked to the attacker by feature anchors may be limited, which is verified by \cite{luo2021no}.
\item  Thirdly, since feature anchors are a trained model component, many approaches, such as secure aggregation and differential privacy, can protect data privacy against reconstruction attacks based on feature anchors.
\end{itemize}
 
\subsection{Additional visualization Experiment Results}\label{Additional Validation Experiment Results}
 \subsubsection{Feature Visualization and Similarity Histogram for All  Methods under Label Distribution Skew  }\label{baseline feature visualization}
\begin{figure*}[t]
    \centering
    \includegraphics[width=0.9\textwidth]{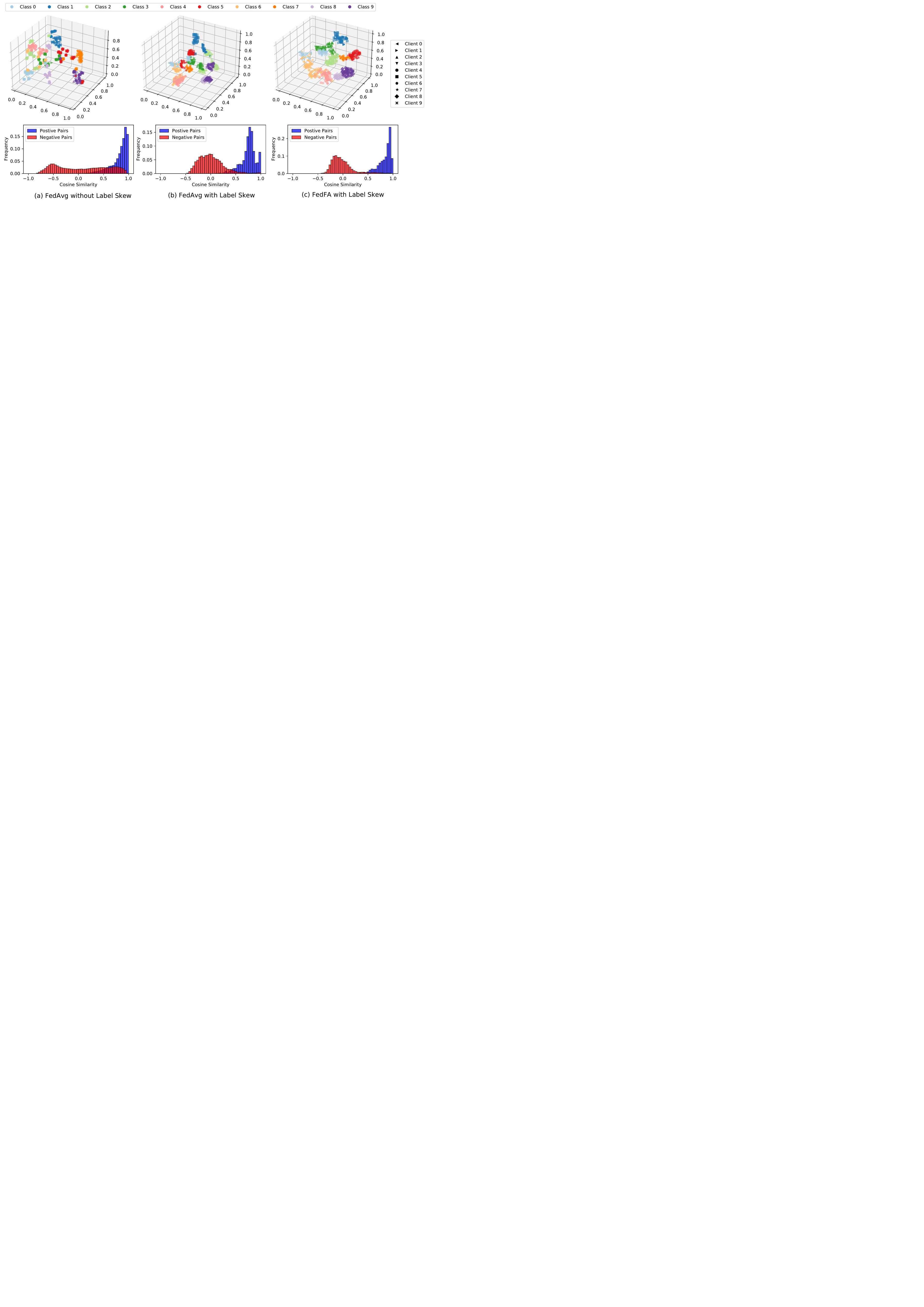}
    \caption{The t-SNE visualization and the histogram of  cosine similarity of features for FedAvg  under data homogeneity and for FedAvg and FedFA under label  distribution skew  with 10 clients. }
    \label{Experimental Validation1}
\end{figure*}

\begin{figure*}[t]
    \centering
    \includegraphics[width=\textwidth]{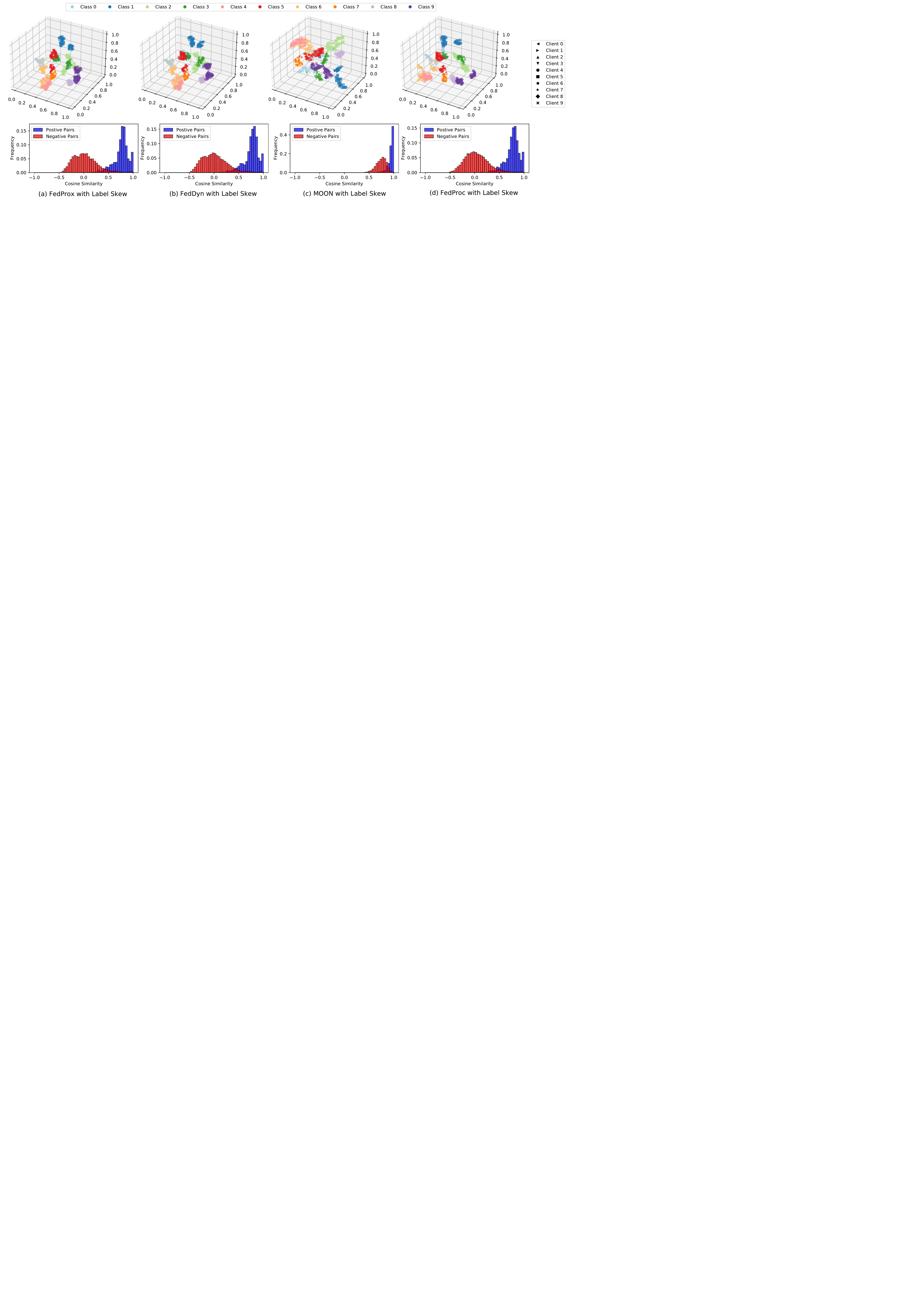}
    \caption{The t-SNE visualization and the histogram of  cosine similarity of features for all baselines under label  distribution skew  with 10 clients. }
    \label{Experimental Validation2}
\end{figure*}

 Figures \ref{Experimental Validation1} and \ref{Experimental Validation2} show the t-SNE visualization and the histogram of cosine similarity of features for label distribution skew for all methods.
We observe that all baselines under label skew exist feature mapping inconsistency across clients. Still, our method FedFA alleviates it significantly, such as class 1 (i.e., dark blue), class 5 (i.e., dark red) and class 9 (i.e., dark purple) in Figures \ref{Experimental Validation1} and \ref{Experimental Validation2}.
Besides, similar to the analysis of Figure \ref{Experimental Validation}, the histograms also show that label distribution skew could induce the lower similarity for \textit{positive pairs},  which means feature inconsistency. Moreover, there exists a low frequency of \textit{positive pairs} and a small gap between \textit{positive pairs} and \textit{negative pairs}, which indicates inconsistent polymerization and discrimination (i.e., sizeable intra-class feature distance and small inter-class feature distance) across clients in classification tasks.
These results of label distribution skew reveal that all client models are trained in inconsistent feature spaces by our baselines, which hurts their performance.

 \subsubsection{Feature Visualization and Similarity Histogram for All  Methods under Feature Distribution Skew  } 
\begin{figure*}[t]
    \centering
    \includegraphics[width=0.9\textwidth]{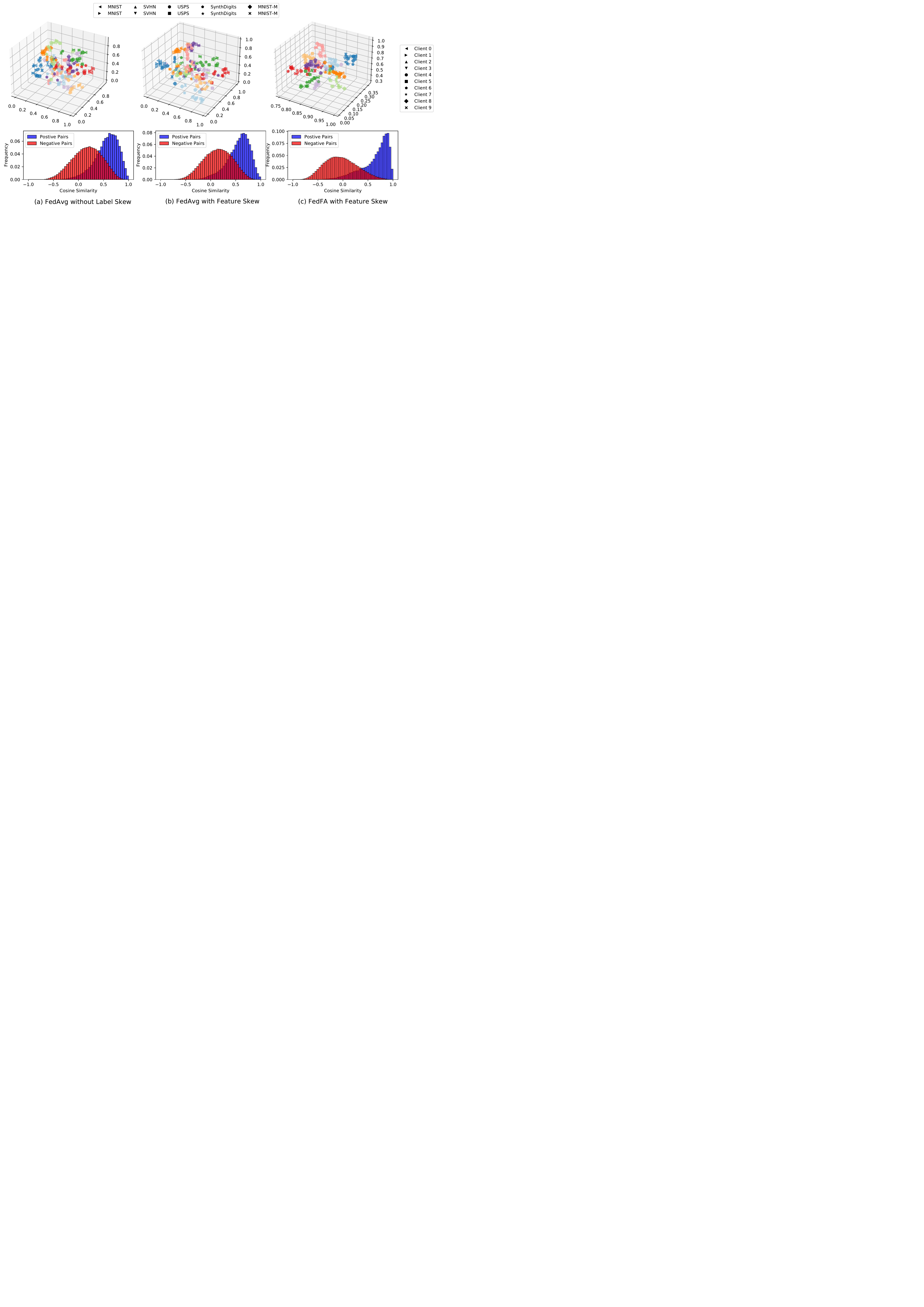}
    \caption{The t-SNE visualization and the histogram of  cosine similarity of features for FedAvg  under data homogeneity and for FedAvg and FedFA  under feature  distribution skew  with 10 clients. }
    \label{Experimental Validation3}
\end{figure*}

\begin{figure*}[t]
    \centering
    \includegraphics[width=\textwidth]{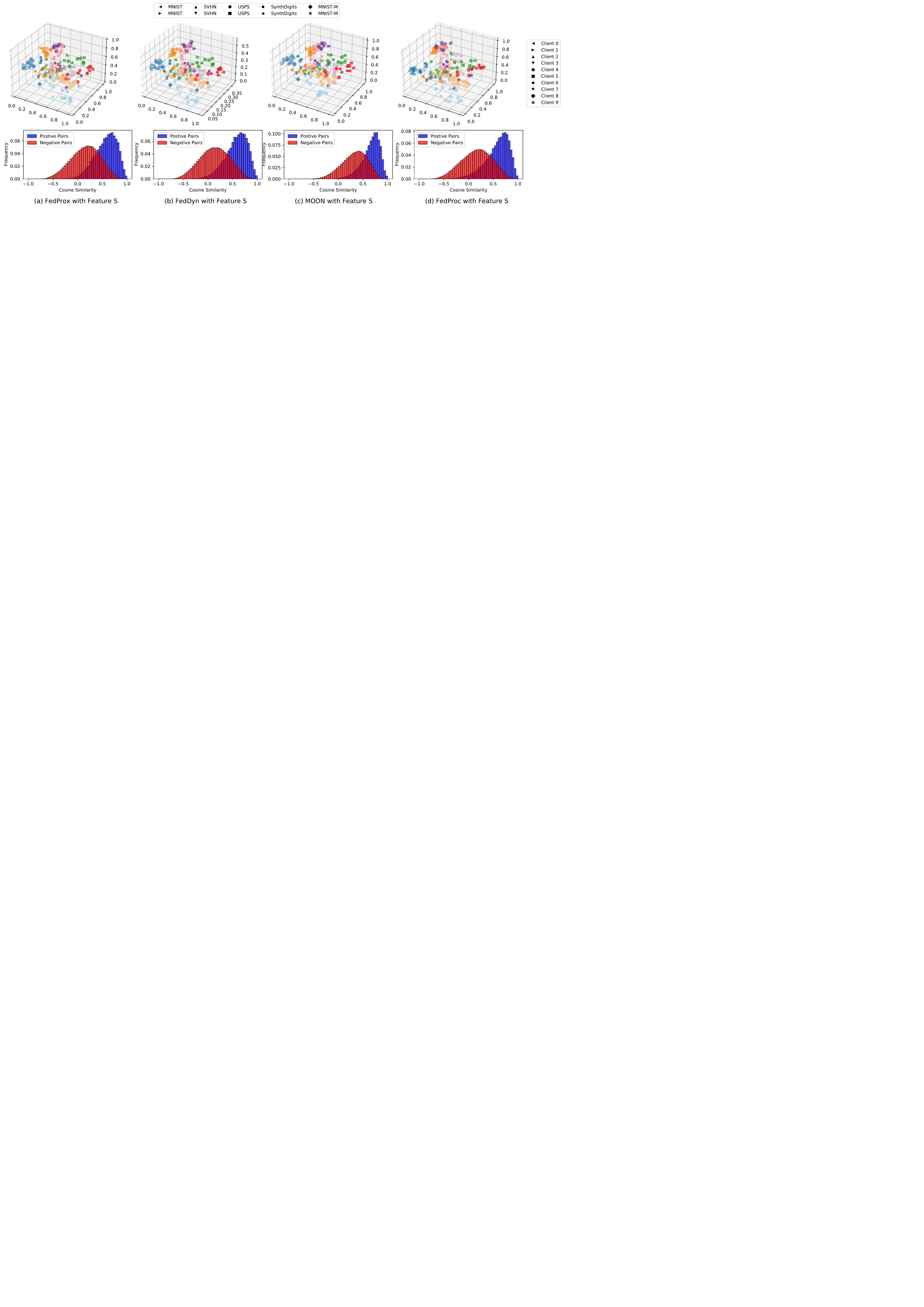}
    \caption{The t-SNE visualization and the histogram of  cosine similarity of features for all baselines under feature  distribution skew  with 10 clients. }
    \label{Experimental Validation4}
\end{figure*}

Similar to label distribution skew, Figures \ref{Experimental Validation3} and \ref{Experimental Validation4} show the t-SNE visualization and the histogram of cosine similarity of features for feature distribution skew for all methods.
We also observe that all baselines under feature skew still suffer from feature mapping inconsistency across clients, but our method does not.
Moreover,  without the feature alignment, all baselines present the weak feature polymerization and feature discrimination of clients' local models, which would make the classifier updates divergent as denoted in (\ref{gradient of classifier}).

 \subsection{Additional Test Experiment Results}\label{Additional Test Experiment Results}


 
   \subsubsection{Performance  under different client
sample rate with 400 rounds}
\begin{figure*}[t]
     \centering
     \includegraphics[width=\textwidth]{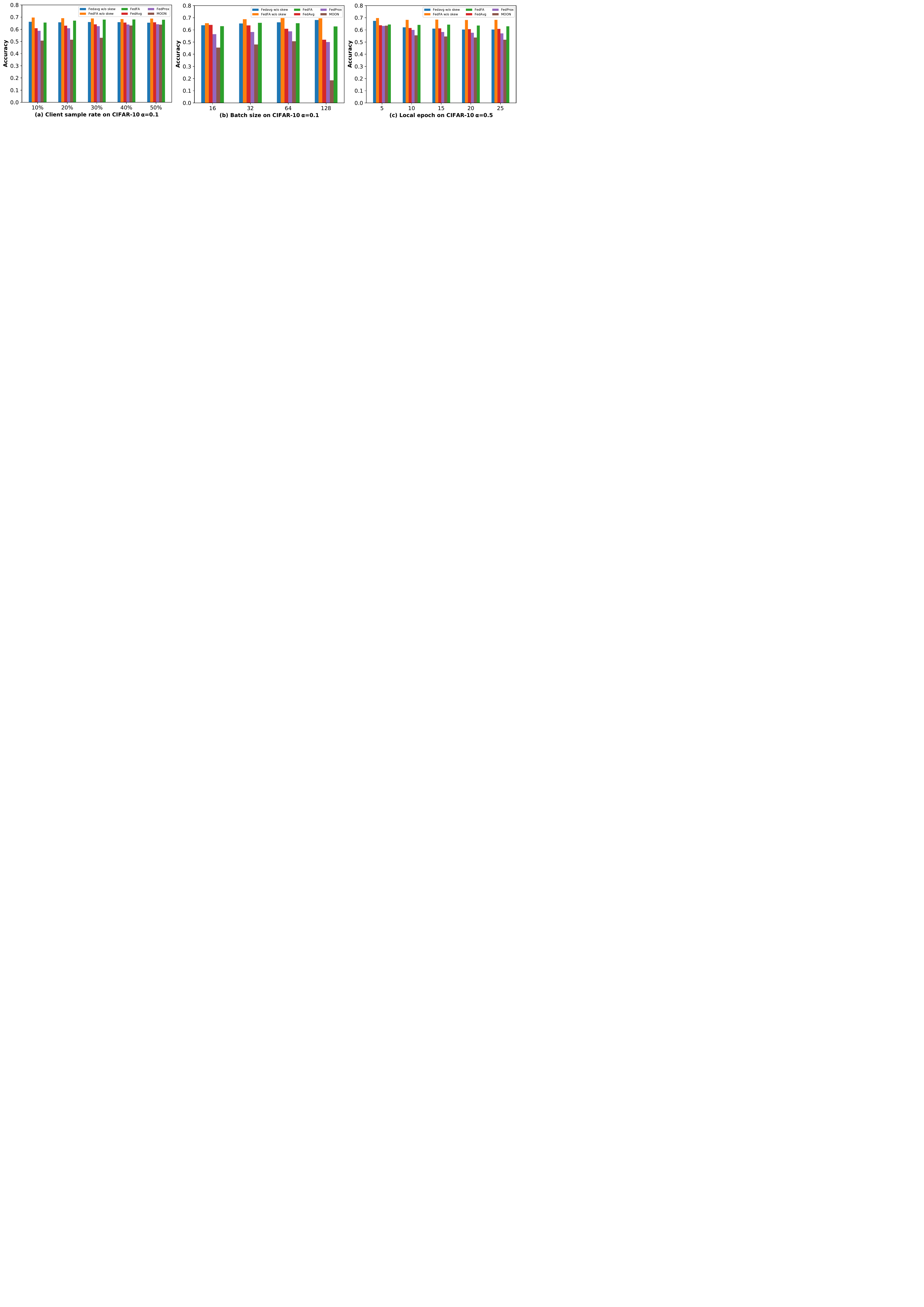}
          \vspace{-15pt}
     \caption{Performance under different (a)  client sample rate, (b) batch size and (c) local epoch on CIFAR-10. }
     \label{federated skew1}
 \end{figure*}

Figure \ref{federated skew1} (a) shows the results with 400 communication rounds to mitigate the impact of limited communication rounds.
Similar to  the results of 200 rounds in Figure \ref{federated skew}, FedFA still presents about $4\%$ accuracy advantage over other baselines.

 \subsubsection{Performance under different batch size}
 Following the setup form \cite{li2021model}, given 100 clients and 400 communication rounds, we investigate the impact of different federated settings on FedFA and  baselines with SGD optimizer with a 0.01 learning rate and  momentum 0.9  on CIFAR-10, including different local epoch, batch size and client sample rate, where the  results are shown in Figures \ref{federated skew}  and \ref{federated skew1}. 
For different local epochs, we observe that the bigger local epochs result in lower performance in all methods, FedProx and MOON suffer from worse performance degradation than FedAvg and FedFA.
For different batch sizes, Figure \ref{federated skew1} presents  FedAvg with a relatively small batch size can work better than that of larger batch sizes, and FedFA, FedProx and MOON performs best with batch size 64.
For different client sample rates, FedFA has a significant performance advantage over other methods, which demonstrates the unique advantage of FedFA in addressing data heterogeneity (i.e., it is robust to different heterogeneous settings).

\subsubsection{Performance under different client numbers}
 \begin{figure*}[htb]
     \centering
     \includegraphics[width=\textwidth]{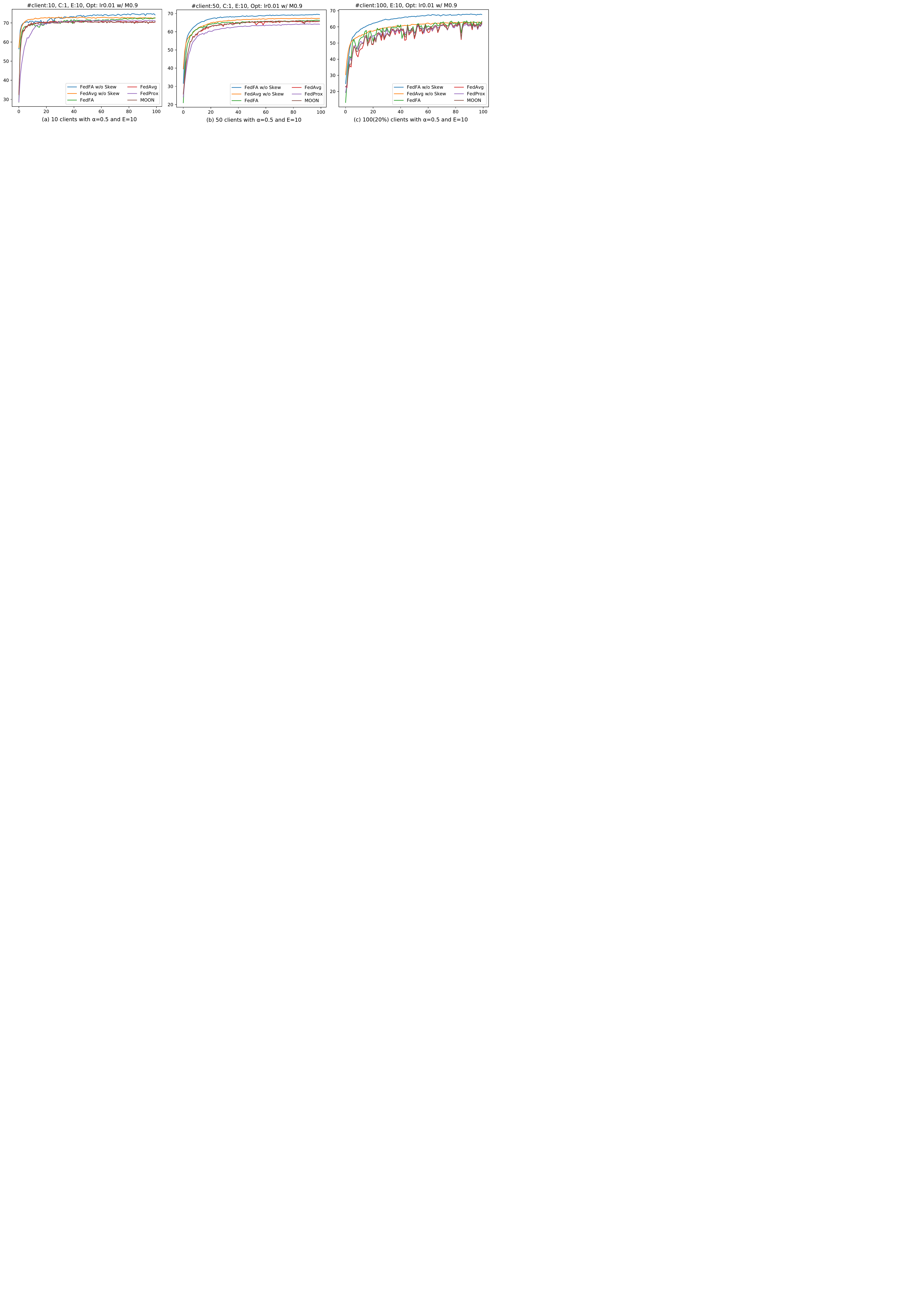}
          \vspace{-15pt}
     \caption{Performance under different client numbers}
     \label{Performance under Different client numbers}
 \end{figure*}
To explore the influence of client numbers (i.e., the flexibility of FedFA for cross-silo federated learning and cross-device federated learning), we follow the setting \cite{li2021model} to test the performance of all methods with 10 clients, 50 clients and 100 clients with 0.2 sample rate and report the top-1 accuracy across rounds in Figure \ref{Performance under Different client numbers}. 
We can see that FedFA keeps better performance than baselines in federated learning with different client numbers.

\subsection{Details of  Experiment Setup}\label{Experiment Setup}
\subsubsection{Specific Models} Our validation and test experiments, including label distribution skew, feature distribution skew, and label $\&$ feature distribution skews, use the models according to Table \ref{specific model}. 
Herein, to ablate the effect of BN layers, we 
replace the BN layer with the GroupNorm layer in all experiments. 
For fair comparison, our models follow those reported in the baselines' works. 
 Specifically, following \cite{acar2021federated}, we use a CNN model for EMNIST, FMNIST, and CIFAR-10, consisting of 
 two 5x5 convolution layers followed by 2x2 max pooling and two fully-connected layers with ReLU activation.
Following \cite{li2021model} and \cite{li2021fedbn}, we utilize the ResNet-18 \cite{he2016deep} with a linear projector for CIFAR-100 and a CNN model with three 5x5 convolution layers followed by five GroupNorm layers for the Mixed Digits dataset.

\begin{table*}[ht]
\caption{The specific parameters settings for all the models used in our experiments.}
\resizebox{\textwidth}{!}{%
\begin{tabular}{@{}c|cc|cccc@{}}
\toprule
                       & \multicolumn{2}{c|}{Validation Experiment}                                                                                                                         & \multicolumn{4}{c}{Test Experiment}                                                                                                                                                                                                                                                                                                                                                                            \\ \midrule
\multirow{2}{*}{Layer} & Label Skew                                                                      & Feature Skew                                                                     & \multicolumn{3}{c|}{Label Skew}                                                                                                                                                                                                                                                                                 & Feature Skew                                                                                 \\ \cmidrule(l){2-7} 
                       & FMNIST                                                                  & Mixed-digit dataset                                                              & FMNIST/EMNIST                                                                                  & CIFAR-10                                                                                       & \multicolumn{1}{c|}{CIFAR-100}                                                                                & Mixed-digit dataset                                                                          \\ \midrule
1                      & \begin{tabular}[c]{@{}c@{}}Conv2d(1, 32, 5)\\ ReLU,MaxPool2D(2,2)\end{tabular}  & \begin{tabular}[c]{@{}c@{}}Conv2d(3, 64, 5)\\ ReLU,MaxPool2D(2,2)\end{tabular}   & \begin{tabular}[c]{@{}c@{}}Conv2d(1, 32, 5)\\ ReLU,MaxPool2D(2,2)\end{tabular}                 & \begin{tabular}[c]{@{}c@{}}Conv2d(3, 64, 5)\\ ReLU,MaxPool2D(2,2)\end{tabular}                 & \multicolumn{1}{c|}{\begin{tabular}[c]{@{}c@{}}Basicbone of Resnet18 \\ with GroupNorm\end{tabular}}          & \begin{tabular}[c]{@{}c@{}}Conv2d(3, 64, 5, 1, 2)\\ ReLU,MaxPool2D(2,2)\end{tabular}  \\ \midrule
2                      & \begin{tabular}[c]{@{}c@{}}Conv2d(32, 32, 5)\\ ReLU,MaxPool2D(2,2)\end{tabular} & \begin{tabular}[c]{@{}c@{}}Conv2d(64, 64, 5)\\ ReLU, MaxPool2D(2,2)\end{tabular} & \begin{tabular}[c]{@{}c@{}}Conv2d(32, 32, 5)\\ ReLU,MaxPool2D(2,2)\end{tabular}                & \begin{tabular}[c]{@{}c@{}}Conv2d(64, 64, 5)\\ ReLU, MaxPool2D(2,2)\end{tabular}               & \multicolumn{1}{c|}{\begin{tabular}[c]{@{}c@{}}FC(512,512)\\ ReLU\end{tabular}}                               & \begin{tabular}[c]{@{}c@{}}Conv2d(64, 64, 5, 1, 2)\\ ReLU,MaxPool2D(2,2)\end{tabular} \\ \midrule
3                      & \begin{tabular}[c]{@{}c@{}}FC(992,384)\\ ReLU\end{tabular}                      & \begin{tabular}[c]{@{}c@{}}FC(1024,384)\\ ReLU\end{tabular}                      & \begin{tabular}[c]{@{}c@{}}FC(992,384)\\ ReLU\end{tabular}                                     & \begin{tabular}[c]{@{}c@{}}FC(1600,384)\\ ReLU\end{tabular}                                    & \multicolumn{1}{c|}{FC(512,256)}                                                                              & \begin{tabular}[c]{@{}c@{}}Conv2d(3, 128, 5, 1, 2)\\ ReLU\end{tabular}               \\ \midrule
4                      & FC(384,100)                                                                     & FC(384,100)                                                                      & \begin{tabular}[c]{@{}c@{}}FC(384,192)\\ ReLU\end{tabular}                                     & \begin{tabular}[c]{@{}c@{}}FC(384,192)\\ ReLU\end{tabular}                                     & \multicolumn{1}{c|}{FC(256,100)}                                                                              & \begin{tabular}[c]{@{}c@{}}FC(6272, 2048)\\ ReLU\end{tabular}                       \\ \midrule
5                      & FC(100,10)                                                                      & FC(100,10)                                                                       & FC(192,10)                                                                                     & FC(192,10)                                                                                     & \multicolumn{1}{c|}{}                                                                                         & \begin{tabular}[c]{@{}c@{}}FC(2048,512)\\ ReLU\end{tabular}                          \\ \midrule
6                      &                                                                                 &                                                                                  &                                                                                                &                                                                                                & \multicolumn{1}{c|}{}                                                                                         & FC(512,10)                                                                                   \\ \midrule
Source                 &                                                                                 &                                                                                  & \begin{tabular}[c]{@{}c@{}}model from\\ \cite{acar2021federated}\end{tabular} & \begin{tabular}[c]{@{}c@{}}model from\\ \cite{acar2021federated}\end{tabular} & \multicolumn{1}{c|}{\begin{tabular}[c]{@{}c@{}}model from\\ \cite{li2021model}\end{tabular}} & \begin{tabular}[c]{@{}c@{}}model from\\ \cite{li2021fedbn}\end{tabular}     \\ \bottomrule
\end{tabular}%
}
\label{specific model}
\end{table*}

\subsubsection{Validation Experiment Setup}\label{Validation Experiment Setup}

We separately sample a subset from test sets of FMNIST and Mixed Digit to visualize the normalized features of the local models based on t-SNE visualization \cite{van2008visualizing}.
In Figure \ref{Experimental Validation}, although we input the same Validation samples into all clients' local modes,  we only show their features mappings for which clients have the corresponding class (i.e., if client 1 only holds class 1 and class 2 samples, we only offer the feature maps of the client 1 model for these two classes, as it would be unfair to ask the local model of client 1 to map the feature of classes on which it did not learn.).
We visualize the features of client models according to the labels (digit dataset) owned by the corresponding client for label (feature) distribution skew. 
The specific setup is described as follows:
 \begin{itemize}
\item \textbf{Label Distribution Skew:} The experiment has 10 clients where each client has 2 classes with 500 samples per class from FMNIST  and utilizes the SGD optimizer with a 0.01 learning rate and without momentum. The federated setting involves 10 local epoch numbers, 15  communication rounds, and a 100$\%$ client sample rate.
     \item \textbf{Feature Distribution Skew:} The experiment has 10 clients where each client has 10 classes with 100 samples per class from one of the digit datasets in Mixed Digit (i.e., MNIST, SVHN, USPS, SynthDigits, and MNIST-M), and utilizes the SGD optimizer with a 0.01 learning rate and without momentum. The federated setting involves 10 local epoch numbers, 15  communication rounds, and a 100$\%$ client sample rate. 
 \end{itemize}

\subsubsection{Test Experiment Setup}\label{Test Experiment Setup}
 
\textbf{Baselines.}
Federated learning \cite{mcmahan2017communication} aims to train a global model parameterized by $\mathbf{w}$ by collaborating a total of $N$ clients  with a central server  to  solve the following optimization problem:
\begin{equation*}
       \min_{\mathbf{w} \in \mathbb{R}^d}  \mathcal{L}(\mathbf{w} ) :=\mathbb{E}_i[\mathcal{L}_i(\mathbf{w} ) ]  =  \sum_i^N \frac{n_i}{n} \mathcal{L}_i(\mathbf{w} )
\end{equation*}
where   $n = \sum_i n_i$ represents the total sample size with $n_i$ being the sample size of the $i$-th client, and $\mathcal{L}_i(\mathbf{w}) := \mathbb{E}_{\xi \in \mathcal{D}_i}[  l_i (\mathbf{w};\xi)]$ is the local objective function in local dataset $\mathcal{D}_i$ of the $i$-th client. 
 
Many methods have been proposed to solve this optimization problem and alleviate the negative impact of data heterogeneity across clients. 
Herein, from the view of local-optimization methods, we compare FedFA with the common federated learning algorithms, including FedAvg\cite{mcmahan2017communication}, FedProx\cite{li2020federated} and the state-of-the-art
methods based on well-designed local regularization including FedDyn \cite{acar2021federated},  MOON \cite{li2021model} and FedProc \cite{mu2021fedproc}.   
The specific description of these methods can be denoted as:
\begin{itemize}
    \item \textbf{FedAvg}: As a canonical method to solve (\ref{fl}) proposed by \cite{mcmahan2017communication},  in each communication round, FedAvg firstly selects a subset of clients and initiates client models as $\mathbf{w}$  and then  updates the local models $\mathbf{w}_i$ by minimizing $ \mathcal{L}_i(\mathbf{w})$, and finally aggregates the local models $\mathbf{w}_i$ as the new global model $\mathbf{w}$ until $\mathcal{L}(\mathbf{w} )$ arrives at a stationary point.
    \item \textbf{FedProx}: FedProx \cite{li2020federated} adds the Euclidean regularization loss between local models and the global model in the local optimization problem, which can be described as:
    \begin{equation}
    \mathcal{L}_i(\mathbf{w} ) = \min_{\mathbf{w}_i}  \mathbb{E}_{(\textbf{x},y)  \in \mathcal{D}_i}[  l_i (\mathbf{w}_i;\mathbf{w}^{(t-1)}) + \frac{\mu }{2} \| \mathbf{w}_i - \mathbf{w}^{(t-1)}\|^2].
\end{equation}
    \item \textbf{FedDyn}: FedDyn \cite{acar2021federated}  modifies the local objective with a dynamic regularization consisting of a linear term based on the first order condition and an above Euclidean-distance term, such that the local minima are consistent with the global stationary point, which can be described as:
\begin{equation}
        \begin{aligned}
                \mathcal{L}_i(\mathbf{w} ) &  = \min_{\mathbf{w}_i}  \mathbb{E}_{(\textbf{x},y)  \in \mathcal{D}_i}[  l_i (\mathbf{w}_i;\mathbf{w}^{(t-1)}) \\ & -  <\nabla\mathcal{L}_i(\mathbf{w}^{(t-1)}), \mathbf{w}_i> + \frac{\mu }{2} \| \mathbf{w}_i - \mathbf{w}^{(t-1)}\|^2].
        \end{aligned}
\end{equation}
    \item \textbf{MOON}: MOON \cite{li2021model}  utilizes the feature similarity of the client model with previous-round local models and with the global model as model-contrastive  regularization to correct the local training of each client, which can be described as:
            \begin{equation}
                    \begin{aligned}
    \mathcal{L}_i&(\mathbf{w} )    = \min_{\mathbf{w}_i}  \mathbb{E}_{(\textbf{x},y)  \in \mathcal{D}_i}[  l_i (\mathbf{w}_i;\mathbf{w}^{(t-1)}) \\  -& \mu  \log \frac{\exp(sim(\textbf{h}_i,\textbf{h}_{\rm global})/\tau)}{\exp(sim(\textbf{h}_i,\textbf{h}_{\rm global})/\tau) + \exp(sim(\textbf{h}_i,\textbf{h}_{\rm pre})/\tau)} ]
            \end{aligned}
\end{equation}
where $\textbf{h}_i,\textbf{h}_{\rm global},\textbf{h}_{\rm pre}$ denote the features of the local model $\mathbf{w}_i$, the global model $\mathbf{w}$, and the local model at previous round $\mathbf{w}_i^{t-1}$  given the same input $\textbf{x}$, respectively; $\tau$  is the hyperparameter to control the effect of cosine similarity  in the model-contrastive loss.
    \item \textbf{FedProc}: Instead of the model-contrastive term in MOON, FedProc \cite{mu2021fedproc} introduces a prototype-contrastive term to regularize the features within each class with class prototypes  \cite{snell2017prototypical}, which can be described as:        
    \begin{equation}
    \begin{aligned}
            \mathcal{L}_i(\mathbf{w} ) = \min_{\mathbf{w}_i}  \mathbb{E}_{(\textbf{x},y)  \in \mathcal{D}_i}[ \frac{t}{T} l_i (\mathbf{w}_i;\mathbf{w}^{(t-1)}) \\ +  (1-\frac{t}{T}) \log \frac{\exp(sim(\textbf{h}_i,\textbf{p}_{\rm c})/\tau)}{\sum_{c=1}^{c= C} \exp(sim(\textbf{h}_i,\textbf{p}_{\rm c})/\tau)} ]
    \end{aligned}
\end{equation}
where $T$ is the targeted communication round, and $\textbf{p}_{\rm c}$ is the prototype of class $c$. In FedProc, $\textbf{p}_{\rm c}$ is updated by the whole local dataset at the end of one communication round (i.e., $\mathbf{{p}}_{c,i}^{(t,k)}= \frac{1}{|\mathcal{D}_{i,c}|}\sum_{(\mathbf{x},c)\in \mathcal{D}_{i,c}} \textbf{h}_{i,c}$).
However, we need to denote that if $\textbf{p}_{\rm c}$ is updated like this, rather than the momentum update as ours and we found that FedProc would suffer from the divergence because the update of $\textbf{p}_{\rm c}$ is too drastic in our experiments.
Therefore, we improve FedProc with our momentum update.
\end{itemize}

\begin{figure}
    \centering
    \includegraphics[width=0.5\textwidth]{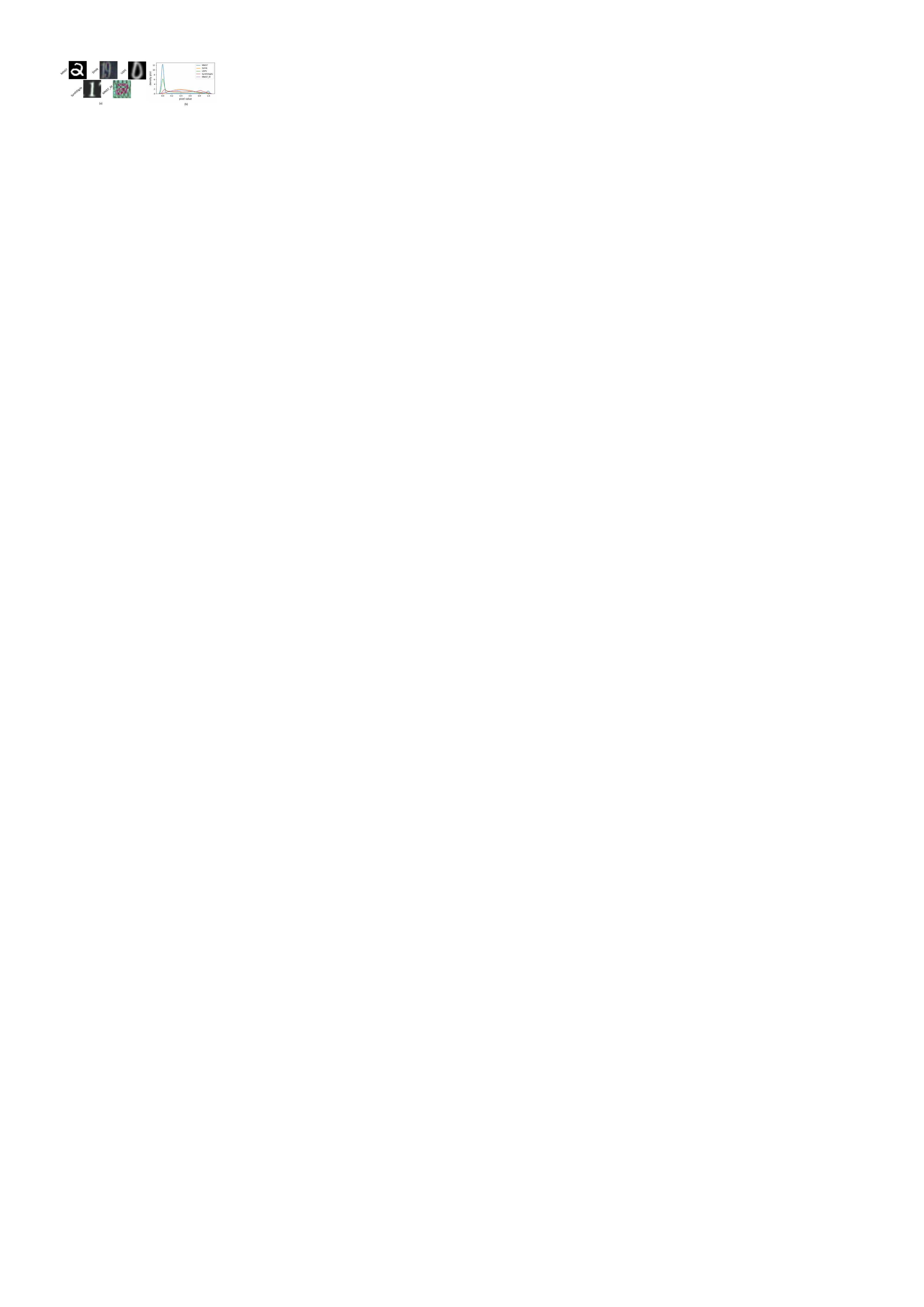}
    \caption{Data visualization. (a) Examples from each dataset (client) in Mixed Digit. (b) feature distributions skew across the datasets (over 100 random samples for each dataset).}
    \label{mixed_digit}
\end{figure}

\textbf{Datasets.} This work aims at image classification tasks under label distribution skew, feature distribution skew, and label $\&$ feature distribution skew, and uses benchmark datasets with the same data  heterogeneity setting as  \cite{mcmahan2017communication,yurochkin2019bayesian,li2021federated}, including  EMNIST\cite{cohen2017emnist}, FMNIST\cite{xiao2017fashion}, CIFAR-10, CIFAR-100 \cite{krizhevsky2009learning}, and  Mixed Digits dataset \cite{li2021fedbn}.
Specifically, for  label distribution skew, we consider two settings:
\begin{itemize}
    \item \textbf{Same size of local dataset}: Following \cite{mcmahan2017communication}, we split data samples based on classes to clients (e.g., $\#C=2$ denotes each client holds two class samples), where each client holds 250 samples per class;
     \item \textbf{Different sizes of local dataset}:  Following \cite{yurochkin2019bayesian}, we first sample $p_i$ from  Dirichlet distribution $Dir(\alpha)$   and then assign $p_{i,c}$ proportion of the samples of class $c$ to client $i$, where we set $\alpha$ as 0.1 and 0.5 to measure the level of data heterogeneity in our experiments. Moreover, when $\alpha=0.1$, the label distributions across clients are so skewed that the quantity of clients' local dataset is also skewed. 
     That is, the experiment cases related to $\alpha=0.1$ would involve label distribution skew and quantity distribution skew, which denotes the unbalanced data size of the local dataset across clients.
\end{itemize} 
For feature distribution skew, we consider two settings:
\begin{itemize}
    \item \textbf{Real-world feature imbalance}: We use a subset of the real-world dataset with natural feature imbalance, EMNIST\cite{cohen2017emnist}, including 10 classes  and 341873 samples (about 34000 samples per class) totally;
     \item \textbf{Artificial feature imbalance}: We use a mixed-digit dataset from \cite{li2021fedbn} consisting of five benchmark digit datasets: 
     MNIST, SVHN,  USPS, SynthDigits and MNIST-M, 
     including 7430 samples for one digit dataset and 743 samples per class. The data visualization is shown as Figure \ref{mixed_digit}.
\end{itemize}


\vfill

\end{document}